\documentclass[10pt,twocolumn,letterpaper]{article}

\usepackage{iccv}
\usepackage{times}
\usepackage{epsfig}
\usepackage{graphicx}
\usepackage{amsmath}
\usepackage{amssymb}
\usepackage{verbatim}
\usepackage{caption}
\usepackage{subcaption}
\usepackage{placeins}
\usepackage{capt-of}
\usepackage{array}
\usepackage{algorithm}      
\usepackage{algpseudocode}  
\usepackage{algorithmicx}   
\usepackage{multirow}

\usepackage{amsmath,amsfonts,bm}

\def\eqref#1{equation~\ref{#1}}

\def\1{\bm{1}}

\DeclareMathAlphabet{\mathsfit}{\encodingdefault}{\sfdefault}{m}{sl}
\SetMathAlphabet{\mathsfit}{bold}{\encodingdefault}{\sfdefault}{bx}{n}

\makeatletter

\def\env@sqcases{%
  \let\@ifnextchar\new@ifnextchar
  \left\lbrack
  \def\arraystretch{1.2}%
  \array{@{}l@{\quad}l@{}}%
}
\makeatother

\usepackage[toc,page]{appendix}

\usepackage[dvipsnames]{xcolor}

\newcommand{\DV}[1]{\textbf{\textcolor{orange}{DV: #1}}}

\usepackage{booktabs}
\newcommand{\ra}[1]{\renewcommand{\arraystretch}{#1}}
\usepackage[
pagebackref=true,
breaklinks=true,
letterpaper=true,
bookmarks=false,
colorlinks=true,
linkcolor=blue,
urlcolor=blue]{hyperref}

\iccvfinalcopy 

\def\iccvPaperID{6343} 

\ificcvfinal\fi
\begin{document}

\title{User-Controllable Multi-Texture Synthesis with Generative Adversarial Networks
\vspace{-0.5cm}
}

\author{
Aibek Alanov$^{1,2,3,}$$^{*}$ \quad
Max Kochurov$^{1,2,}$$^{*}$ \quad 
Denis Volkhonskiy$^{2}$ \quad 
Daniil Yashkov$^{5}$ \\ 
Evgeny Burnaev$^{2}$ \quad 
Dmitry Vetrov$^{1,4}$\\
\small{
$^1$Samsung AI Center Moscow \;
$^2$Skolkovo Institute of Science and Technology \;
$^3$National Research University Higher School of Economics} \\
\small{$^4$Samsung-HSE Laboratory, National Research University Higher School of Economics} \\
\small{$^5$Federal Research Center "Computer Science and Control" of the Russian Academy of Sciences}}

\twocolumn[{%
\renewcommand\twocolumn[1][]{#1}%
\maketitle
\vspace*{-3em}
\begin{center}
\centering
\includegraphics[width=.263\linewidth]{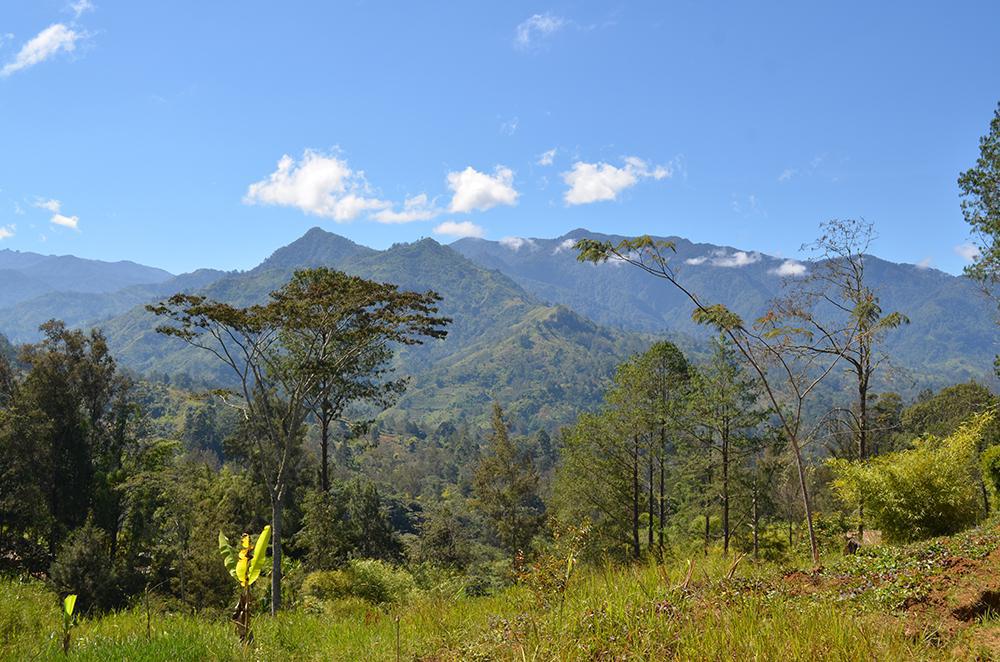}~%
\includegraphics[width=.174\linewidth]{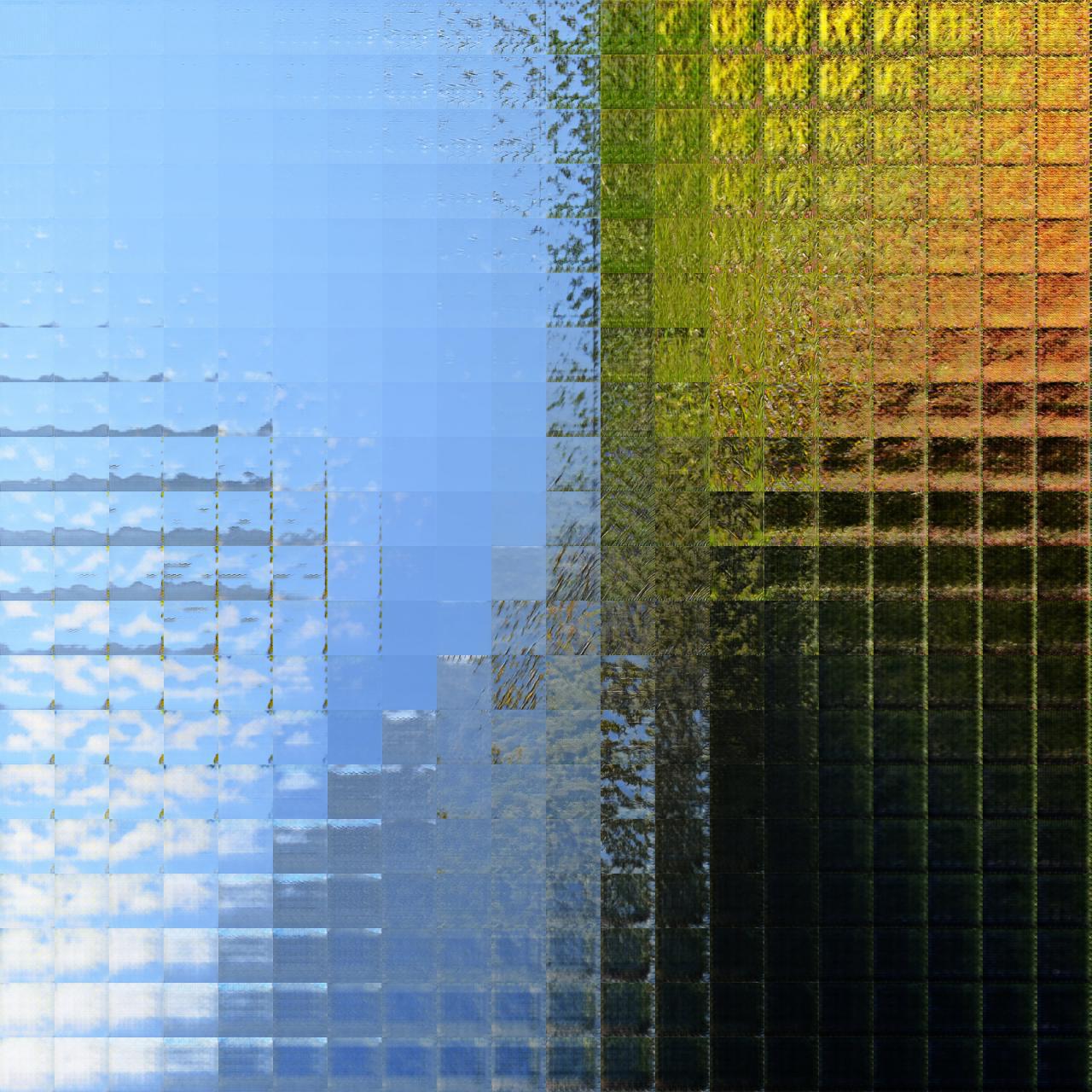}~%
\includegraphics[width=.263\linewidth]{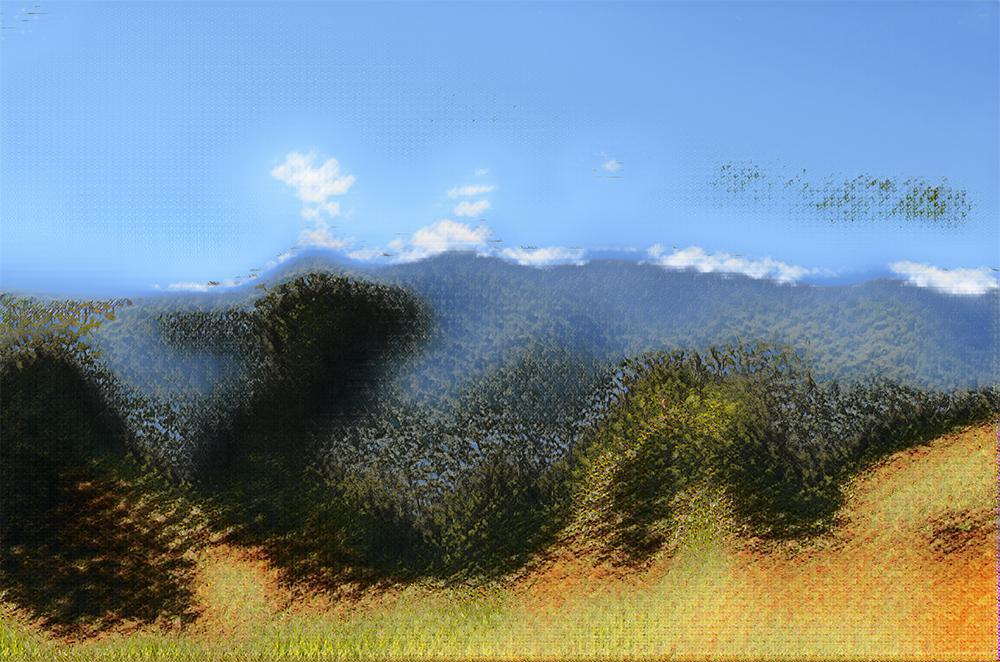}~%
\includegraphics[width=.263\linewidth]{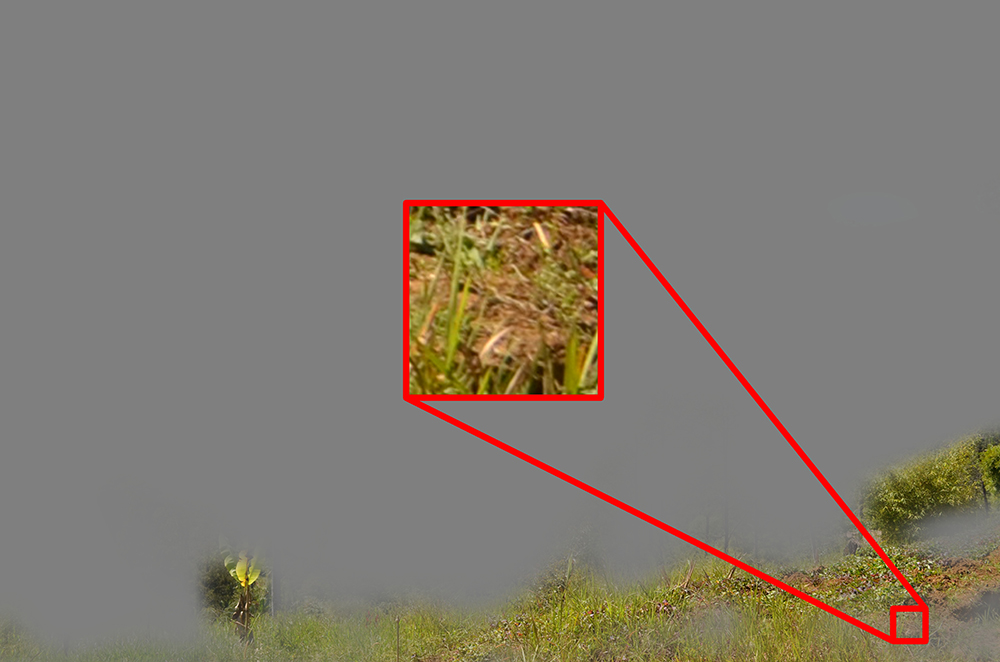}
\captionof{figure}{\small One can take 1)~New Guinea $3264\times4928$ landscape photo, learn 2)~a manifold of 2D texture embeddings for this photo, visualize 3)~texture map for the image and perform 4)~texture detection for a patch using distances between learned embeddings.}
\label{fig:teaser}
\end{center}
}]

\begin{abstract}
    We propose a novel multi-texture synthesis model based on generative adversarial networks (GANs) with a user-controllable mechanism. The user control ability allows to explicitly specify the texture which should be generated by the model. This property follows from using an encoder part which learns a latent representation for each texture
    from the dataset. To ensure a dataset coverage, we use an adversarial loss function that penalizes for incorrect reproductions of a given texture. In experiments, we show that our model can learn descriptive texture manifolds for large datasets and from raw data such as a collection of high-resolution photos. Moreover, we apply our method to produce 3D textures and show that it outperforms existing baselines.
    \vspace{-0.3cm}
\end{abstract}

\renewcommand{\thefootnote}{*}
\footnotetext{equal contribution}

\section{Introduction}

Textures are essential and crucial perceptual elements in computer graphics. They can be defined as images with repetitive or periodic local patterns. 
Texture synthesis models based on deep neural networks have recently drawn a great interest to a computer vision community. 
Gatys \etal \cite{gatys2015texture, gatys2016image} proposed to use a convolutional neural network as an effective texture feature extractor.  They proposed to use a Gram matrix of hidden layers of a pre-trained VGG network as a descriptor of a texture. Follow-up papers \cite{johnson2016perceptual, ulyanov2016texture, li2016precomputed} significantly speed up a synthesis of texture by substituting an expensive optimization process in \cite{gatys2015texture, gatys2016image} to a fast forward pass of a feed-forward convolutional network. However, these methods suffer from many problems such as generality inefficiency (i.e., train one network per texture) and poor diversity (i.e., synthesize visually indistinguishable textures). 

\begin{figure*}[t]
    \centering
    \includegraphics[width=\linewidth]{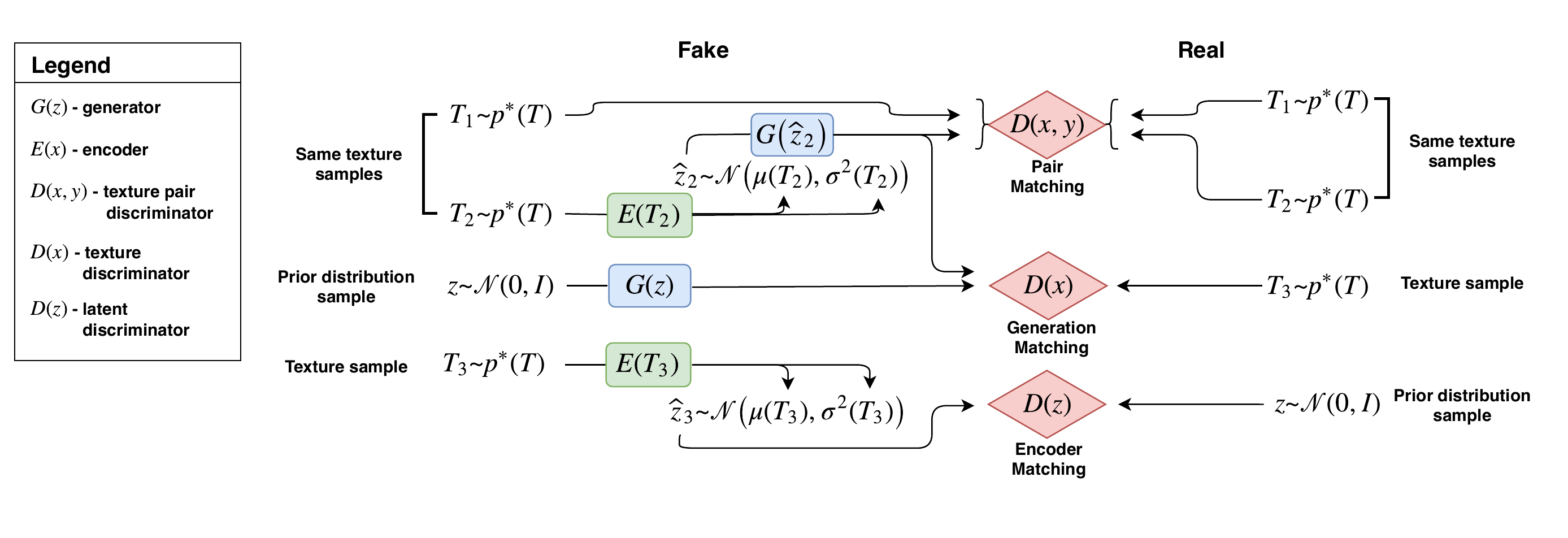}
    \vspace{-1.2cm}
    \caption{Training pipeline of the proposed method}
    \label{fig:pipeline}
    \vspace{-0.6cm}
\end{figure*}

\begin{table}[!b]
\vspace{-0.5cm}
    \centering
    \caption{Comparison of multi-texture synthesis methods}
    \ra{1.}
    \begin{tabular}{@{}>{\raggedright}p{3.3cm}lll@{}}\toprule
        & PSGAN \cite{bergmann2017learning} & DTS \cite{li2017diversified} & Ours \\
        \midrule
        multi-texture & \checkmark  & \checkmark & \checkmark \\
        user control &  & \checkmark & \checkmark \\
        dataset coverage & & \checkmark & \checkmark \\ 
        scalability with respect to dataset size & \checkmark & & \checkmark \\
        ability to learn textures from raw data & \checkmark & & \checkmark \\
        unsupervised texture detection & & & \checkmark \\
        applicability to 3D & \checkmark & & \checkmark \\
        \bottomrule
    \end{tabular}
    \label{table:comparison}
\end{table}

Recently, Periodic Spatial GAN (PSGAN) \cite{bergmann2017learning} and Diversified Texture Synthesis (DTS) \cite{li2017diversified} models were proposed as an attempt to partly solve these issues. PSGAN and DTS are multi-texture synthesis models, i.e., they train one network for generating many textures. However, each model has its own limitations (see Table ~\ref{table:comparison}). PSGAN has incomplete dataset coverage, and a user control mechanism is absent. Lack of dataset coverage means that it can miss some textures from the training dataset. The absence of a user control does not allow to explicitly specify the texture which should be generated by the model in PSGAN. DTS is not scalable with respect to dataset size, cannot be applied to learn textures from raw data and to synthesize 3D textures. It is not scalable because the number of parameters of the DTS model linearly depends on the number of textures in the dataset. The learning from raw data means that the input for the model is a high-resolution image as in Figure~\ref{fig:teaser} and the method should extract textures in an unsupervised way. DTS does not support such training mode (which we call \emph{fully unsupervised}) because for this model input textures should be specified explicitly. 
The shortage of generality to 3D textures in DTS model comes from inapplicability of VGG network to 3D images.

We propose a novel multi-texture synthesis model which does not have limitations of PSGAN and DTS methods. Our model allows for generating a user-specified texture from the training dataset. This is achieved by using an encoder network which learns a latent representation for each texture from the dataset. To ensure the complete dataset coverage of our method we use a loss function that penalizes for incorrect reproductions of a given texture. Thus, the generator is forced to learn the ability to synthesize each texture seen during the training phase.
Our method is more scalable with respect to dataset size compared to DTS and is able to learn textures in a fully unsupervised way from raw data as a collection of high-resolution photos. We show that our model learns a descriptive texture manifold in latent space. Such low dimensional representations can be applied as useful texture descriptors, for example, for an unsupervised texture detection (see Figure~\ref{fig:teaser}). Also, we can apply our approach to 3D texture synthesis because we use fully adversarial losses and do not utilize VGG network descriptors. 

We experimentally show that our model can learn large datasets of textures. We check that our generator learns all textures from the training dataset by conditionally synthesizing each of them. We demonstrate that our model can learn meaningful texture manifolds as opposed to PSGAN (see Figure~\ref{fig:2dmanifolds}). We compare the efficiency of our approach and DTS in terms of memory consumption and show that our model is much more scalable than DTS for large datasets. 

We apply our method to 3D texture-like porous media structures which is a real-world problem from Digital Rock Physics. Synthesis of porous structures plays an important role \cite{volkhonskiy2019reconstruction} because an assessment of the variability in the inherent material properties is often experimentally not feasible. Moreover, usually it is necessary to acquire a number of representative samples of the void-solid structure. We show that our method outperforms a baseline \cite{mosser2017reconstruction} in the porous media synthesis which trains one network per texture.

Briefly summarize, we can highlight the following key advantages of our model:
\begin{itemize}
    \vspace{-0.25cm}
    \item user control (conditional generation),
    \vspace{-0.25cm}
    \item full dataset coverage, 
    \vspace{-0.25cm}
    \item scalability with respect to dataset size,
    \vspace{-0.25cm}
    \item ability to learn descriptive texture manifolds from raw data in a fully unsupervised way,
    \vspace{-0.25cm}
    \item applicability to 3D texture synthesis. 
    \vspace{-0.25cm}
\end{itemize}

\begin{figure*}[t]
    \centering
    \includegraphics[width=\linewidth]{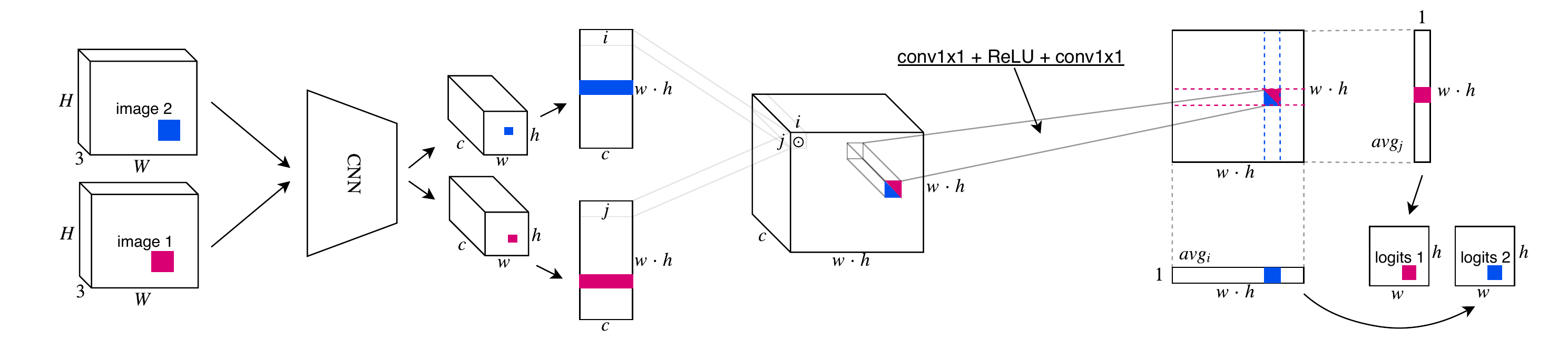}
    \vspace{-.8cm}
    \caption{The architecture of the discriminator on pairs $D_{\tau}(x, y)$.}
    \label{fig:arch.disc}
    \vspace{-0.6cm}
\end{figure*}

\section{Proposed Method}
We look for a multi-texture synthesis pipeline that can generate textures in a user-controllable manner, ensure full dataset coverage and be scalable with respect to dataset size. We use an encoder network $E_{\varphi}(x)$ which allows to map textures to a latent space and gives low dimensional representations. We use a similar generator network $G_{\theta}(z)$ as in PSGAN. 

The generator $G_{\theta}(z)$ takes as an input a noise tensor $z \in \mathbb{R}^{d\times h_z\times w_z}$ which has three parts $z = [z^g, z^l, z^p]$. These parts are the same as in PSGAN:
\begin{itemize}
    \vspace{-0.25cm}
    \item $z^g \in \mathbb{R}^{d^g\times h_z\times w_z}$ is a global part which determines the type of texture. It consists of only one vector $\bar{z}^g$ of size $d^g$ which is repeated through spatial dimensions. 
    \vspace{-0.25cm}
    \item $z^l \in \mathbb{R}^{d^l\times h_z\times w_z}$ is a local part and each element $z^l_{kij}$ is sampled from a standard normal distribution $\mathcal{N}(0, 1)$ independently. This part encourages the diversity within one texture. 
    \vspace{-0.25cm}
    \item $z^p \in \mathbb{R}^{d^p\times h_z\times w_z}$ is a periodic part and $z^p_{kij} = \sin(a_k(z^g)\cdot i + b_k(z^g)\cdot j + \xi_{k})$ where $a_k, b_k$ are trainable functions and $\xi_{k}$ is sampled from $U[0, 2\pi]$ independently. This part helps generating periodic patterns. 
    \vspace{-0.1cm}
\end{itemize}

We see that for generating a texture it is sufficient to put the vector $\bar{z}^g$ as an input to the generator $G_{\theta}$ because $z^l$ is obtained independently from $\mathcal{N}(0, 1)$ and $z^p$ is computed from $z^g$. It means that we can consider $\bar{z}^g$ as a latent representation of a corresponding texture and we will train our encoder $E_{\varphi}(x)$ to recover this latent vector $\bar{z}^g$ for an input texture $x$. Further, we will assume that the generator $G_{\theta}(z)$ takes only the vector $\bar{z}^g$ as input and builds other parts of the noise tensor from it. For simplicity, we denote $\bar{z}^g$ as $z$. 

The encoder $E_{\varphi}(x)$ takes an input texture $x$ and returns the distribution $q_{\varphi}(z|x) = \mathcal{N}(\mu_{\varphi}(x), \sigma^2_{\varphi}(x))$ of the global vector $z$ (the same as $\bar{z}^g$) of the texture $x$. 

Then we can formulate properties of the generator $G_{\theta}(z)$ and the encoder $E_{\varphi}(x)$ we expect in our model:
\begin{itemize}
    \vspace{-0.21cm}
    \item samples $G_{\theta}(z)$ are real textures if we sample $z$ from a prior distribution $p(z)$ (in our case it is $\mathcal{N}(0, I)$). 
    \vspace{-0.21cm}
    \item if $z_{\varphi}(x) \sim q_{\varphi}(z|x)$ then $G_{\theta}(z_{\varphi}(x))$ has the same texture type as $x$.
    \vspace{-0.21cm}
    \item an aggregated distribution of the encoder $E_{\varphi}(x)$ should be close to the prior distribution $p(z)$, i.e. $q_{\varphi}(z) = \int q_{\varphi}(z|x)p^*(x)dx \approx p(z)$ where $p^*(x)$ is a true distribution of textures. 
    \vspace{-0.21cm}
    \item samples $G_{\theta}(z_{\varphi})$ are real textures if $z_{\varphi}$ is sampled from aggregated $q_{\varphi}(z)$.
    \vspace{-0.21cm}
\end{itemize}

To ensure these properties we use three types of adversarial losses:
\begin{itemize}
    \vspace{-.2cm}
    \item \emph{generator matching:} $\mathcal{L}_x$ for matching the distribution of both samples $G_{\theta}(z)$ and reproductions $G_{\theta}(z_{\varphi})$ to the distribution of real textures $p^*(x)$. 
    \vspace{-.2cm}
    \item \emph{pair matching:} $\mathcal{L}_{xx}$ for matching the distribution of pairs $(x, x')$ to the distribution of pairs $(x, G_{\theta}(z_{\varphi}(x)))$ where $x$ and $x'$ are samples of the same texture. It will ensure that $G_{\theta}(z_{\varphi}(x))$ has the same texture type as $x$. 
    \vspace{-.2cm}
    \item \emph{encoder matching:} $\mathcal{L}_z$ for matching the aggregated distribution $q_{\varphi}(z)$ to the prior distribution $p(z)$. 
    \vspace{-.2cm}
\end{itemize}

We consider exact definitions of these adversarial losses in Section ~\ref{sec:objs}. We demonstrate the whole pipeline of the training procedure in Figure~\ref{fig:pipeline} and in Appendix~\ref{sec:alg_desc}. 

\subsection{Generator \& Encoder Objectives}
\label{sec:objs}

\textbf{Generator Matching.}
For matching both samples $G_{\theta}(z)$ and reproductions $G_{\theta}(z_{\varphi})$ to real textures we use a discriminator $D_{\psi}(x)$ as in PSGAN which maps an input image $x$ to a two-dimensional tensor of spatial size $s\times t$. Each element $D_{\psi}^{ij}(x)$ of the discriminator's output corresponds to a local part $x$ and estimates probability that such receptive field is real versus synthesized by $G_{\theta}$. 
Then a value function $V_x(\theta, \psi)$ of such adversarial game $\displaystyle\min_{\theta}\max_{\psi} V_x(\theta, \psi)$ will be the following:
\vspace{-0.15cm}
\begin{gather}
    V_x(\theta, \psi)
    = \dfrac{1}{st}\sum\limits_{i, j}^{s, t}\left[\mathbb{E}_{p^*(x)}\log D_{\psi}^{ij}(x) + \right. \\
    \left. + \mathbb{E}_{p(z)}\log(1 - D^{ij}_{\psi}(G_{\theta}(z))) + \mathbb{E}_{q_{\varphi}(z)}\log(1 - D^{ij}_{\psi}(G_{\theta}(z_{\varphi})))\right] \nonumber 
\end{gather}

As in \cite{goodfellow2014generative} we modify the value function $V_{x}(\theta, \psi)$ for the generator $G_{\theta}$ by substituting the term $\log(1 - D^{ij}_{\psi}(\cdot))$ to $-\log D^{ij}_{\psi}(\cdot)$. So, the adversarial loss $\mathcal{L}_x$ is 
\begin{gather}
    \mathcal{L}_x(\theta) = -\dfrac{1}{st}\sum\limits_{i, j}^{s, t}\left[\mathbb{E}_{p(z)}\log D^{ij}_{\psi}(G_{\theta}(z)) + \right. \\
    \left. + \mathbb{E}_{q_{\varphi}(z)}\log D^{ij}_{\psi}(G_{\theta}(z_{\varphi}))\right] \rightarrow \min_{\theta} \nonumber
\end{gather}

\textbf{Pair Matching.}
The goal is to match \emph{fake} pairs $(x, G_{\theta}(z_{\varphi}(x)))$ to \emph{real} ones $(x, x')$ where $x$ and $x'$ are samples of the same texture (in practice, we can obtain real pairs by taking two different random patches from one texture). For this purpose we use a discriminator $D_{\tau}(x, y)$ of special architecture which is provided in Figure~\ref{fig:arch.disc}. 
The discriminator $D_{\tau}(x, y)$ takes two input images and convolves them separately with the same convolutional layers. After obtaining embedded tensors with dimensions $c\times h\times w$ for each input image, we reshape each tensor to a matrix with size $c\times h\cdot w$. Each row in these matrices represents an embedding for the corresponding receptive field in the initial images. Then we calculate pairwise element products of these two matrices which gives us a tensor with dimension $c \times h\cdot w\times h\cdot w$. We convolve it with two convolutional layers using $1\times 1$ kernels and obtain a two-dimensional matrix of size $h\cdot w\times h\cdot w$. The element in the $i$-th row and $j$-th column of this matrix represents the mutual similarity between the corresponding receptive field in the first image and the one in the second image. Then we average this matrix row-wisely (for $x$) and column-wisely (for $y$). We obtain two vectors of sizes $h\cdot w$ and reshape them into matrices with dimensions $h\times w$. To simplify the notation, we concatenate these matrices into a matrix $h\times 2w$, then we take element-wise sigmoid and output it as a matrix of discriminator's probabilities like in PSGAN. 

We consider the following distributions:
\begin{itemize}
    \vspace{-0.2cm}
    \item $p_{xx}^*(x, y)$ over real pairs $(x, y)$ where $x$ and $y$ are examples of the same texture;
    \vspace{-0.2cm}
    \item $p_{\theta, \varphi}(x, y)$ over fake pairs $(x, y)$ where $x$ is a real texture and $y$ is its reproduction, i.e.,  $y = G_{\theta}(z_{\varphi}(x))$.
    \vspace{-0.2cm}
\end{itemize}

We denote the dimension of the discriminator's output matrix as $p\times q$ and $D_{\tau}^{ij}(x, y)$ as the $ij$-th element of this matrix.
The value function $V_{xx}(\theta, \varphi, \tau)$ for this adversarial game $\displaystyle\min_{\theta,\varphi}\max_{\tau} V_{xx}(\theta, \varphi, \tau)$ is
\begin{gather}
    V_{xx}(\theta, \varphi, \tau) = \dfrac{1}{pq}\sum\limits_{i, j}^{p,q}\left[\mathbb{E}_{p_{xx}^*(x, y)}\log D_{\tau}^{ij}(x, y) \: + \right. \nonumber \\ 
    \left. + \: \mathbb{E}_{p_{\theta, \varphi}(x, y)}\log(1 - D^{ij}_{\tau}(x, y))\right]  
    \label{eq:VxxObj}
\end{gather}
The discriminator $D_{\tau}$ tries to maximize the value function $V_{xx}(\theta, \varphi, \tau)$ while the generator $G_{\theta}$ and the encoder $E_{\varphi}$ minimize it. 

Then the adversarial loss $\mathcal{L}_{xx}$ is
\begin{gather}
    \mathcal{L}_{xx}(\theta, \varphi) = -\dfrac{1}{pq}\sum\limits_{i, j}^{p,q}\mathbb{E}_{p_{\theta, \varphi}(x, y)}\log D^{ij}_{\tau}(x, y) \rightarrow \min_{\theta, \varphi}
\end{gather}
To compute gradients $\mathcal{L}_{xx}(\theta, \varphi)$ with respect to $\varphi$ parameters we use a reparameterization trick \cite{kingma2013auto, rezende2014stochastic, titsias2014doubly}.

\textbf{Encoder Matching.}
We need to use encoder matching because otherwise if we use only one objective $\mathcal{L}_{xx}(\theta, \varphi)$ for training the encoder $E_{\varphi}(x)$ then embeddings for textures can be very far from samples $z$ that come from the prior distribution $p(z)$. It will lead to unstable training of the generator $G_{\theta}(z)$ because it should generate good images both for samples from the prior $p(z)$ and for embeddings which come from the encoder $E_{\varphi}$.

Therefore, to regularize the encoder $E_{\varphi}$ we match the prior distribution $p(z)$ and the aggregated encoder distribution $q_{\varphi}(z) = \int q_{\varphi}(z|x)p^*(x)dx$ using the discriminator $D_{\zeta}(z)$. It classifies samples $z$ from $p(z)$ versus ones from $q_{\varphi}(z)$. The minimax game of $E_{\varphi}(x)$ and $D_{\zeta}$ is defined as $\displaystyle \min_{\varphi}\max_{\zeta} V_z(\varphi, \zeta)$, where $V_z(\varphi, \zeta)$ is
\vspace{-0.7cm}
\begin{gather}
    V_z(\varphi, \zeta) = \mathbb{E}_{p(z)}\log D_{\zeta}(z)\\
    + \mathbb{E}_{q_{\varphi}(z)}\log(1 - D_{\zeta}(z)) \nonumber 
\end{gather}
To sample from $q_{\varphi}(z)$ we should at first sample some texture $x$ then sample $z$ from the encoder distribution by $z = \mu_{\varphi}(x) + \sigma_{\varphi}(x) * \varepsilon$, where $\varepsilon \sim \mathcal{N}(0, I)$. The adversarial loss $\mathcal{L}_z(\varphi)$ is
\vspace{-0.2cm}
\begin{gather}
    \mathcal{L}_z(\varphi) = -\mathbb{E}_{q_{\varphi}(z)}\log D_{\zeta}(z) \rightarrow \min_{\varphi}
\end{gather}
As for the loss $\mathcal{L}_{xx}(\theta, \varphi)$, we compute gradients of $\mathcal{L}_z(\varphi)$ with respect to $\varphi$ using the reparameterization trick. 

\textbf{Final Objectives.}
Thus, for both the generator $G_{\theta}$ and the encoder $E_{\varphi}$ we optimize the following objectives:
\begin{itemize}
    \item the generator $G_{\theta}$ loss
    \vspace{-0.2cm}
    \begin{gather}
        \mathcal{L}(\theta) = \alpha_1\mathcal{L}_x(\theta) + \alpha_2\mathcal{L}_{xx}(\theta, \varphi) \rightarrow \min_{\theta}
    \end{gather}\vspace{-0.8cm}
    \item the encoder $E_{\varphi}$ loss
    \vspace{-0.2cm}
    \begin{gather}
        \mathcal{L}(\varphi) = \beta_1\mathcal{L}_z(\varphi) + \beta_2\mathcal{L}_{xx}(\theta, \varphi) \rightarrow \min_{\varphi}
    \end{gather}\vspace{-0.8cm}
\end{itemize}
In experiments, we use $\alpha_1 = \alpha_2 = \beta_1 = \beta_2 = 1$. 

\section{Related Work}
Traditional texture synthesis models can broadly be divided into two categories: non-parametric and parametric. Non-parametric methods \cite{efros2001image, efros1999texture, kwatra2003graphcut, wei2000fast} synthesize new texture by repeating and resampling local patches from the given example.
Such techniques allow obtaining large textures. However, these approaches require heavy computations and can be slow. Parametric approaches \cite{heeger1995pyramid, portilla2000parametric} consider an explicit model of textures by introducing statistical measures. To generate new texture, we should run an optimization process which matches the statistics of the synthesized image and a given texture. The method \cite{portilla2000parametric} shows good results in generating different textures. The main limitations of this approach are its high time complexity and the need to define handcrafted statistics for matching textures.  

Deep learning methods were shown to be an efficient parametric model for texture synthesis. Papers of Gatys \etal \cite{gatys2015texture, gatys2016image} are a milestone: they proposed to use Gram matrices of VGG intermediate layer activations as texture descriptors. This approach allows for generating high-quality images of textures \cite{gatys2015texture} by running an expensive optimization process. Subsequent works \cite{ulyanov2016texture, johnson2016perceptual, li2016precomputed} significantly accelerate a texture synthesis by approximating this optimization procedure by fast feed-forward convolutional networks. Further works improve this approach either by using optimization techniques \cite{frigo2016split, gatys2016preserving, li2016combining}, introducing an instance normalization \cite{ulyanov2017improved, ulyanovinstance} or applying GANs-based models for non-stationary texture synthesis \cite{zhou2018non}. These methods have significant limitations such as the requirement to train one network per texture and poor diversity of samples. 

\textbf{Multi-texture synthesis methods.} DTS \cite{li2017diversified} was introduced by Li \etal as a multi-texture synthesis model. It consists of one feed-forward convolutional network which takes one-hot vector corresponding to a specific texture and a noise vector, passes them through convolutional layers and generates an image. Such architecture makes DTS non-scalable for large datasets because the number of model parameters depends linearly on the dataset size. It cannot learn from raw data in a fully unsupervised way because input textures for this model should be specified explicitly by one-hot vectors. Also, this method is not applicable for 3D textures because it utilizes VGG Gram matrix descriptors which are suitable only for 2D images. 

Spatial GAN (SGAN) model \cite{jetchev2016texture} was introduced by Jetchev \etal as the first method where GANs \cite{goodfellow2014generative} are applied to texture synthesis. It showed good results on certain textures, surpassing the results of \cite{gatys2015texture}. Bergmann \etal \cite{bergmann2017learning} improved SGAN by introducing Periodic Spatial GAN (PSGAN) model. It allows learning multiple textures due to an input noise in this method has a hierarchical structure. Since PSGAN optimizes only vanilla GAN loss it does not ensure the full dataset coverage. It is also known as mode collapse and it is considered as a common problem in GAN models \cite{arjovsky17a, radford2015unsupervised, thanh2019improving}. Also this method does not allow conditional generating of textures, i.e. we cannot explicitly  specify  the  texture  which  should  be  generated  by the model. 

Our model is based on GANs with an encoder network which allows mapping an input texture to a latent embedding. There are many different ways to train an autoencoding GANs \cite{rosca2017variational, UlyanovVL18, brock2016neural, dumoulin2017learned, donahue2016adversarial, li2017alice, CycleGAN2017}. The main part in such models is the objective which is responsible for accurate reproduction of a given image by the model. Standard choices are $L_1$ and $L_2$ norms \cite{rosca2017variational, UlyanovVL18, CycleGAN2017} or perceptual distances \cite{brock2016neural}. For textures, the VGG Gram matrix-based loss is more common \cite{gatys2015texture, ulyanov2016texture, johnson2016perceptual}. We use the adversarial loss for this purpose inspired by \cite{xian2018texturegan} where it is used for image synthesis guided by sketch, color, and texture. The benefit of such loss is that it can be easily applied to 3D textures. Previous works \cite{mosser2017reconstruction, volkhonskiy2019reconstruction} on synthesizing 3D porous material used GANs-based methods with 3D convolutional layers inside a generator and a discriminator. However, they trained separate models for each texture. We show that our model allows to learn multiple 3D textures with a conditional generation ability. 

\section{Experiments}
\begin{figure}
    \centering
    \begin{subfigure}{.45\linewidth}
    \includegraphics[width=\linewidth]{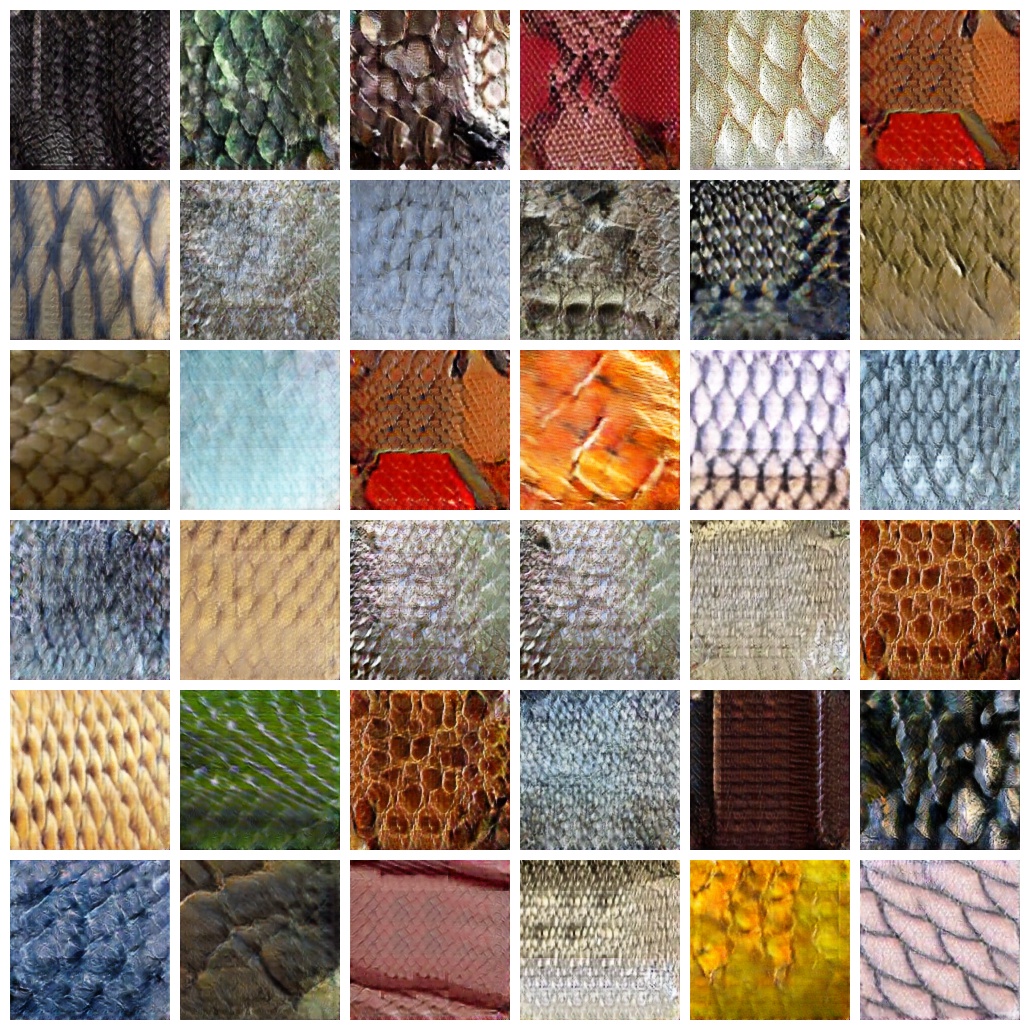}
    \caption{PSGAN-5D samples}
    \end{subfigure}~%
    \begin{subfigure}{.45\linewidth}
    \includegraphics[width=\linewidth]{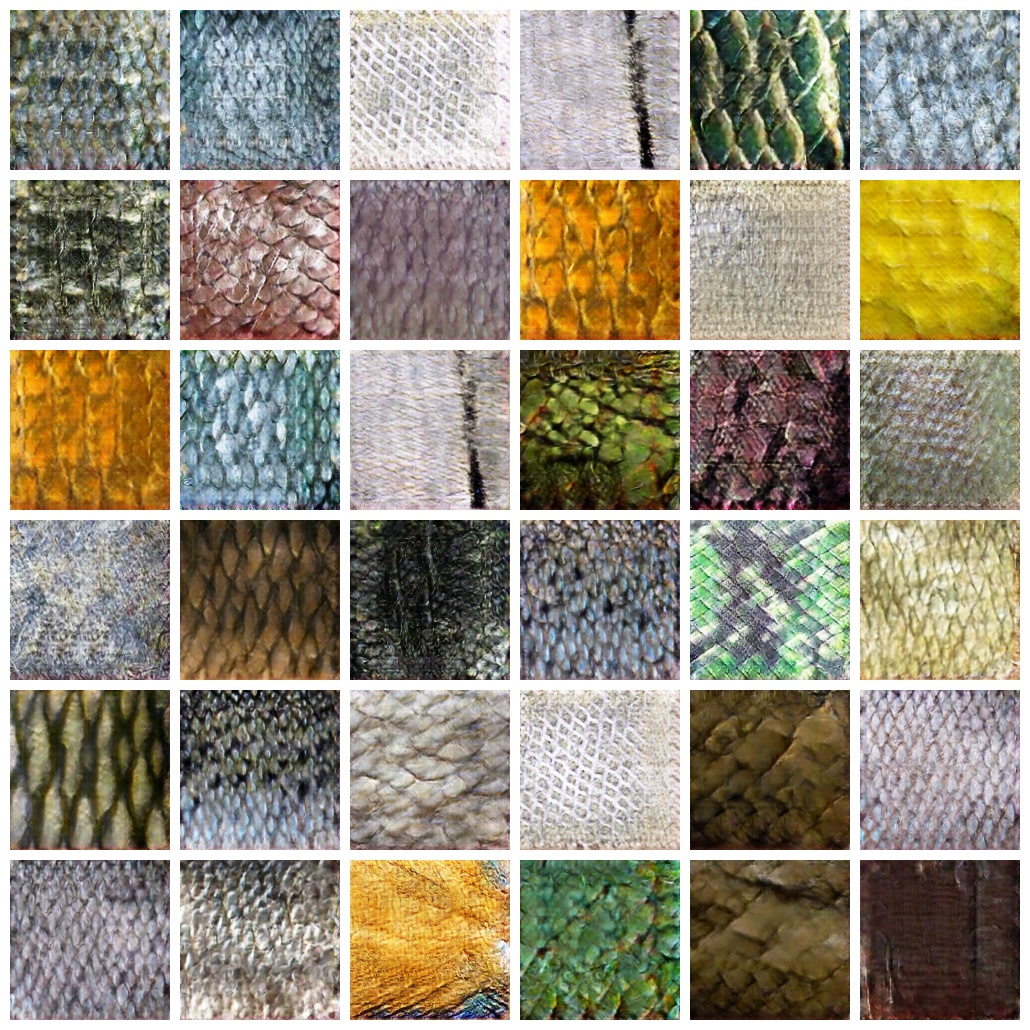}
    \caption{Our-2D model samples}
    \end{subfigure}
    \begin{subfigure}{\linewidth}
    \includegraphics[width=\linewidth]{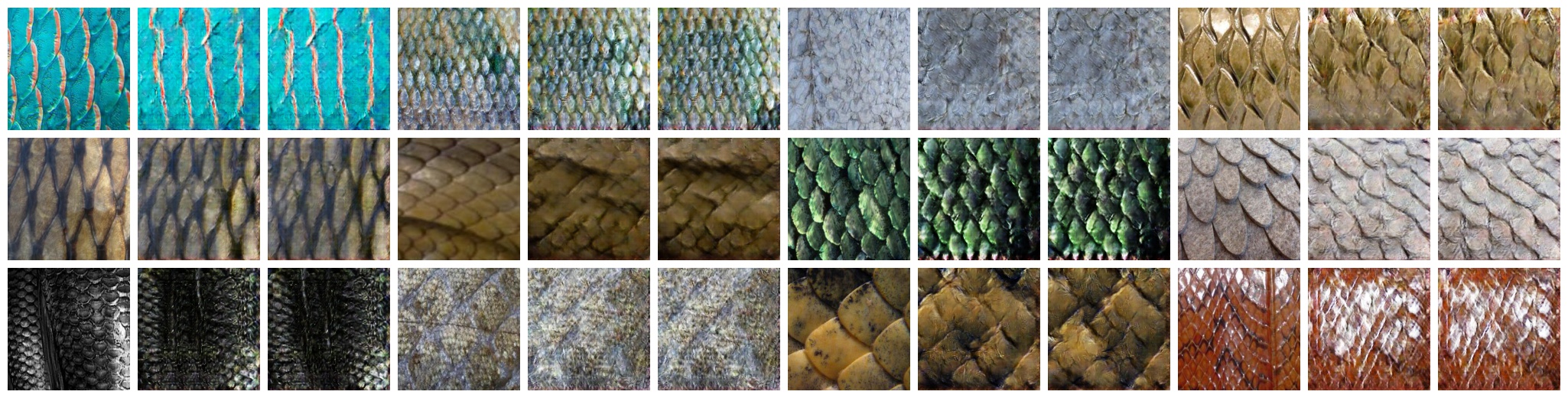}
    \caption{Our-2D model reproductions. Columns 1,4,7,10 are real textures, others are reproductions}
    \end{subfigure}
    \vspace{-0.3cm}
    \caption{Examples of generated/reproduced textures from PSGAN and our model}
    \label{fig:sample-examples}
    \vspace{-0.5cm}
\end{figure}
In experiments, we train our model on scaly, braided, honeycomb and striped categories from Oxford Describable Textures Dataset \cite{cimpoi2014dtw}. These are datasets with natural textures in the wild. We use the same fully-convolutional architecture for $D_{\psi}(x)$, $G_\theta(z)$ as in PSGAN \cite{bergmann2017learning}. We used a spectral normalization \cite{miyato2018spectral} for discriminators that significantly improved training stability.
For $E_{\varphi}(x)$ we used similar architecture as for $D_{\psi}(x)$. Global dimension $d^g$ was found to be a sensitive parameter and we choose it separately for different models. The encoder $E_{\varphi}$ network outputs a tensor with $2d^g$ channels followed by global average pooling to get parameters $\mu_g$, $\log \sigma_g$ for encoding distribution $q(z\;|\; x)=\mathcal{N}(\mu_g(x), \sigma^2_g(x))$. As in PSGAN model we fix $d^l=20$ and $d^p=4$. For the discriminator $D_{\tau}(x, y)$ we used the architecture described in Figure~\ref{fig:arch.disc}. A complete reference for network architectures can be found in Appendix~\ref{app:network-architectures}.

\vspace{-0.07cm}
\subsection{Inception Score for Textures}
\label{sec:inception_score}
It is a common practice in natural image generation to evaluate a model that approximates data distribution $p^*(x)$ using Inception Score \cite{szegedy2016inception}. For this purpose Inception network is used to get label distribution $p(t|x)$. Then one calculates 
\begin{equation}
    IS=\exp \left\{\mathbb{E}_{x\sim p^*(x)}\operatorname{KL}\bigl(p(t|x)\|p(t)\bigr)\right\},
\end{equation}
where $p(t)=\mathbb{E}_{x\sim p^*(x)}p(t|x)$ is aggregated label distribution. The straightforward application of Inception network does not make sense for textures. Therefore, we train a classifier with an  architecture similar\footnote{The only difference is the number of output logits} to $D_\psi(x)$ to predict texture types for a given texture dataset. To do that properly, we manually clean our data from duplicates so that every texture example has a distinct label and use random cropping as data augmentation. Our trained classifier achieves 100\% accuracy on a scaly dataset. We use this classifier to evaluate Inception Score for models trained on the same texture dataset.

\begin{table}[!b]
\vspace{-0.5cm}
    \centering
    \caption{Inception Scores for conditional and unconditional generation from PSGAN and our model. Classifier used to compute IS achieved perfect accuracy on train data.}
    \label{tab:inception-scores-with-psgan}
    \ra{1.}
    \begin{tabular}{@{}lll@{}}\toprule
        Model & Uncond. IS & Cond. IS \\
        \midrule
        PSGAN-5D & 73.68$\pm$0.6 & NA\\
        Our-2D& 73.74$\pm$0.3&\textbf{103.96$\pm$0.1}\\
        \bottomrule
    \end{tabular}
    \label{table:comparison}
\end{table}

\subsection{Unconditional and Conditional Generation}

\label{sec:comparision-to-psgan}
\begin{figure}
    \centering
    \begin{subfigure}{\linewidth}
    \includegraphics[width=\linewidth]{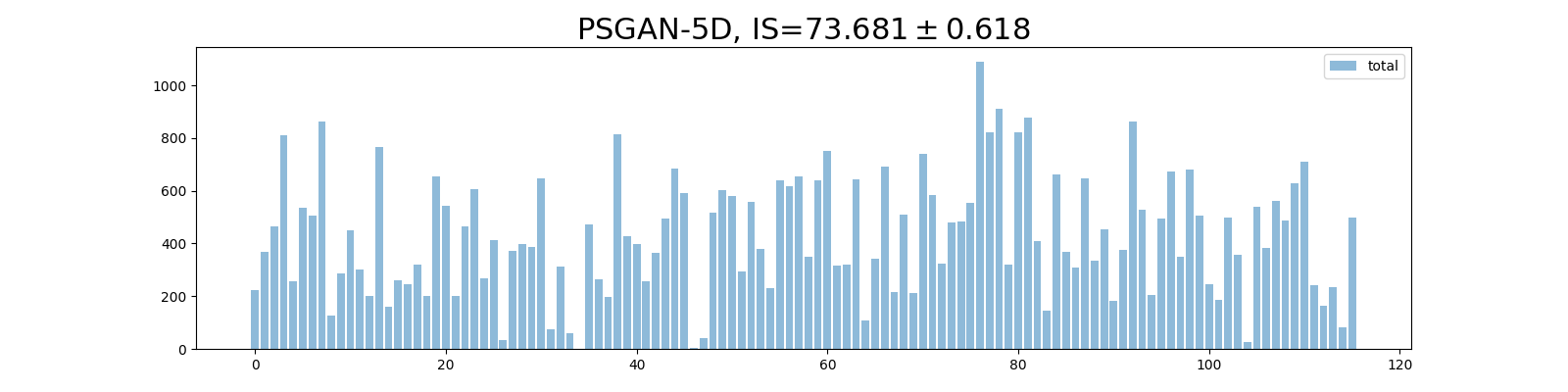}
    \caption{PSGAN-5D samples\label{fig:bins-unconditional:psgan}}
    \end{subfigure}
    \begin{subfigure}{\linewidth}
    \includegraphics[width=\linewidth]{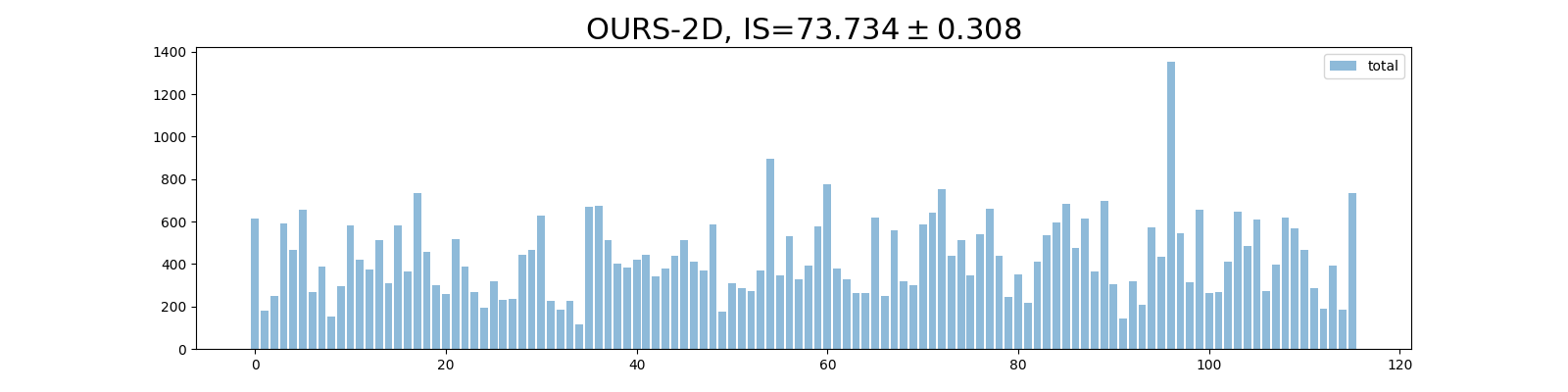}
    \caption{Our-2D model samples}
    \end{subfigure}
    \begin{subfigure}{\linewidth}
    \includegraphics[width=\linewidth]{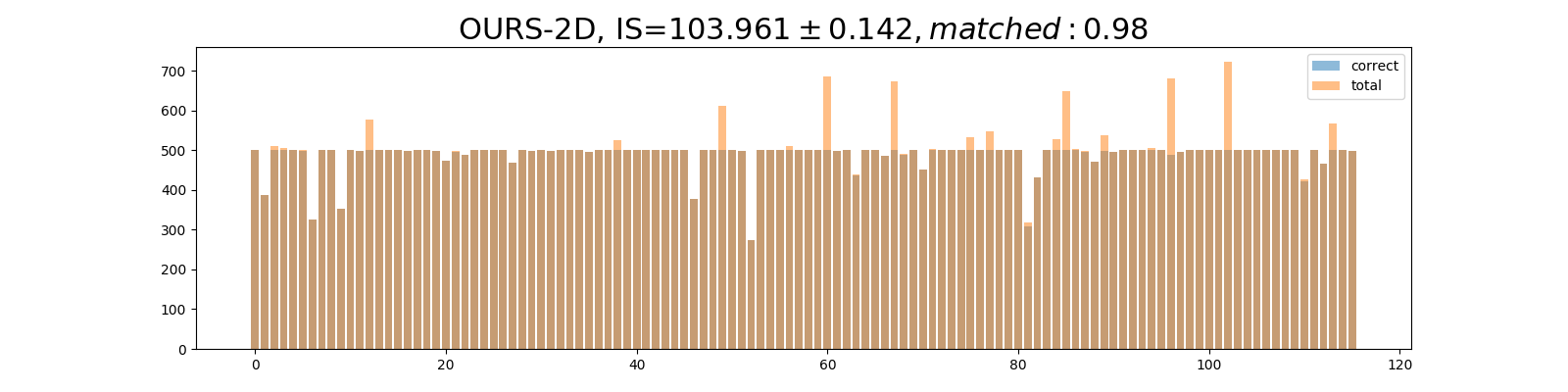}
    \caption{Our-2D model reproductions}
    \end{subfigure}
    \caption{Histogram of classifier predictions on 50000 generated samples from PSGAN (a) and Our model (b) and for 500 reproductions per class for our model (c). Each bin represents a separate texture class.}
    \label{fig:bins-unconditional}
    \vspace{-0.5cm}
\end{figure}
For models like PSGAN we are not able to obtain reproductions, we only have access to texture generation process. One would ask for the guarantees that a model is able to generate every texture in the dataset from only the prior distribution. We evaluate PSGAN and our model on a scaly dataset with 116 unique textures. After models are trained, we estimate the Inception Score. We observed that Inception Score differs with $d^g$ and thus picked the best $d^g$ separately for both PSGAN and our model obtaining $d^g=5$ and $d^g=2$ respectively. Both models were trained with Adam \cite{kingma2015adam} (betas=0.5,0.99)  with batch size 64 on a single GPU. Their best performance was achieved in around 80k iterations. For both models, we used spectral normalization to improve training stability \cite{miyato2018spectral}. 

Both models can generate high-quality textures from low dimensional space. Our model additionally can successfully generate reproductions for every texture in the dataset. Figure~\ref{fig:bins-unconditional} and Table~\ref{tab:inception-scores-with-psgan} summarise the results for conditional (reproductions) and unconditional texture generation. Figure~\label{fig:bins-unconditional:psgan} indicates PSGAN may have missing textures, our model does not suffer from this issue. Inception Score suggests that conditional generation is a far better way to sample from the model. In Figure~\ref{fig:sample-examples} we provide samples and texture reproductions for trained models. A larger set of samples and reproductions for every texture can be found in Appendix~\ref{app:attachments} along with evaluations on braided, honeycomb and striped categories from Oxford Describable Textures Dataset.

\subsection{Texture Manifold}
\label{sec:texture-manifold}
\begin{figure}
    \centering
    \begin{subfigure}{0.45\linewidth}
    \includegraphics[width=\linewidth]{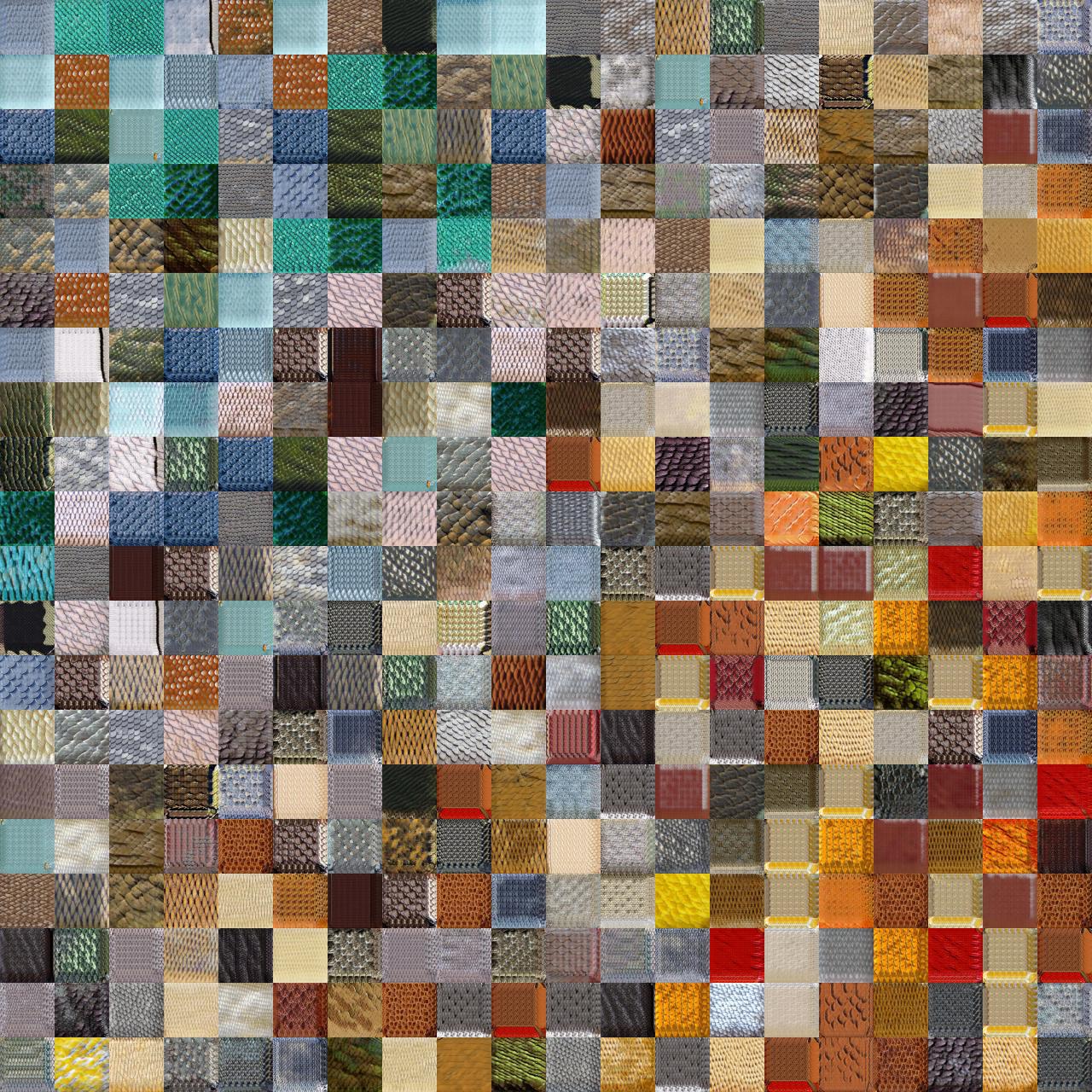}
    \caption{PSGAN 2D manifold}
    \end{subfigure}~%
    \begin{subfigure}{0.45\linewidth}
    \includegraphics[width=\linewidth]{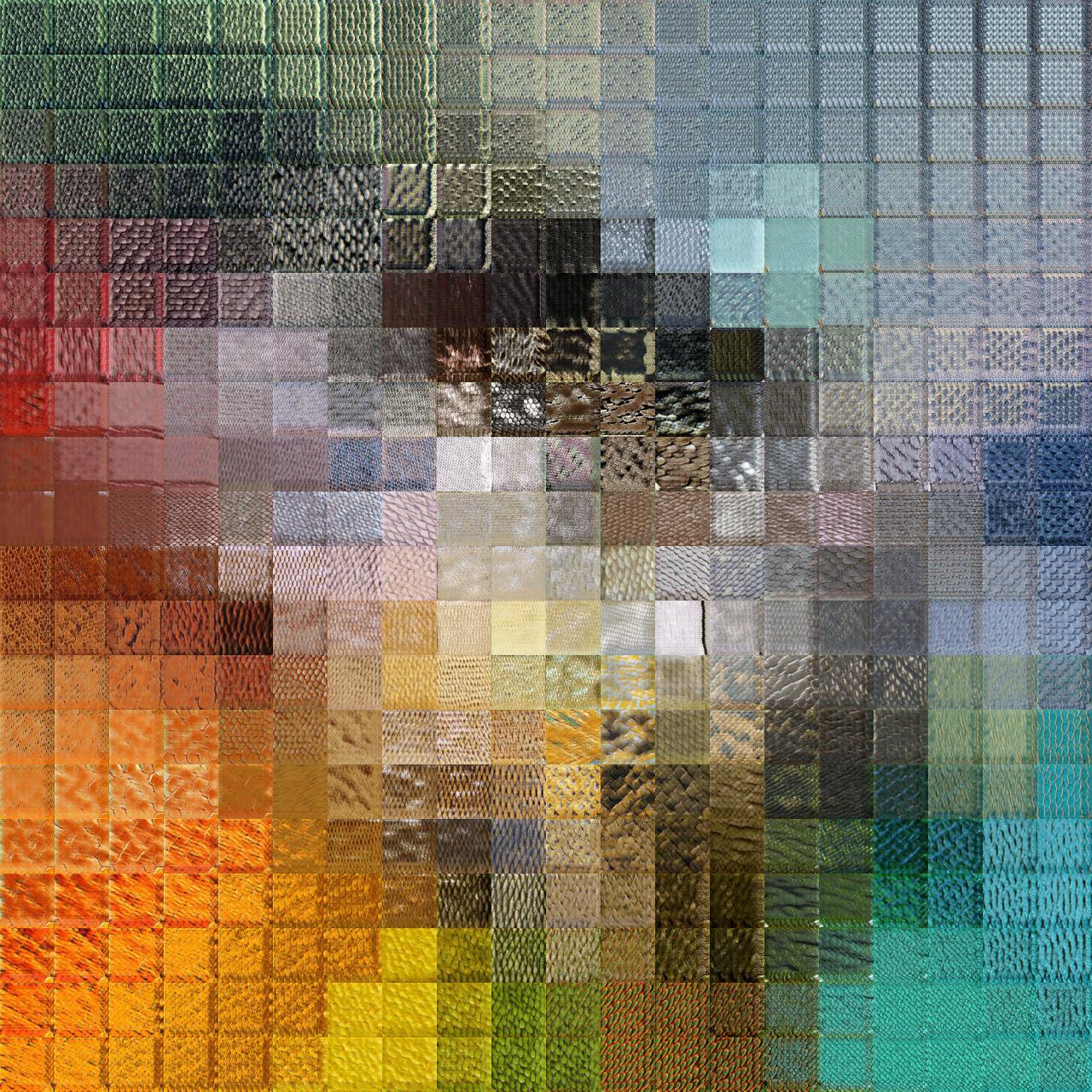}
    \caption{Our model 2D manifold}
    \end{subfigure}
    \caption{2D manifold for 116 textures from scaly dataset. Our model gives paces one texture to a distinct location. Grid is taken over $[-2.25, 2.25] \times [-2.25, 2.25]$ with step $0.225$}
    \label{fig:2dmanifolds}
    \vspace{-0.7cm}
\end{figure}
Autoencoding property is a nice to have feature for generative models. One can treat embeddings as low dimensional data representations. As shown in section~\ref{sec:comparision-to-psgan} our model can reconstruct every texture in the dataset. Moreover, we are able to visualize the manifold of textures since we trained this model with $d^g=2$. To compare this manifold to PSGAN, we train a separate PSGAN model with $d^g=2$. 2D manifolds near prior distribution for both models can be found in Figure~\ref{fig:2dmanifolds}. Our model learns visually better 2D manifold and allocates similar textures nearby.
Visualizations for manifolds while training (for different epochs) can be found in Appendix~\ref{app:attachments}.

\subsection{Learning Texture Manifolds from Raw Data}
\label{sec:learning-texture-manifolds-from-raw-data}
\begin{figure}[!t]
    \centering
    \begin{subfigure}{.6\linewidth}
    \includegraphics[width=\linewidth]{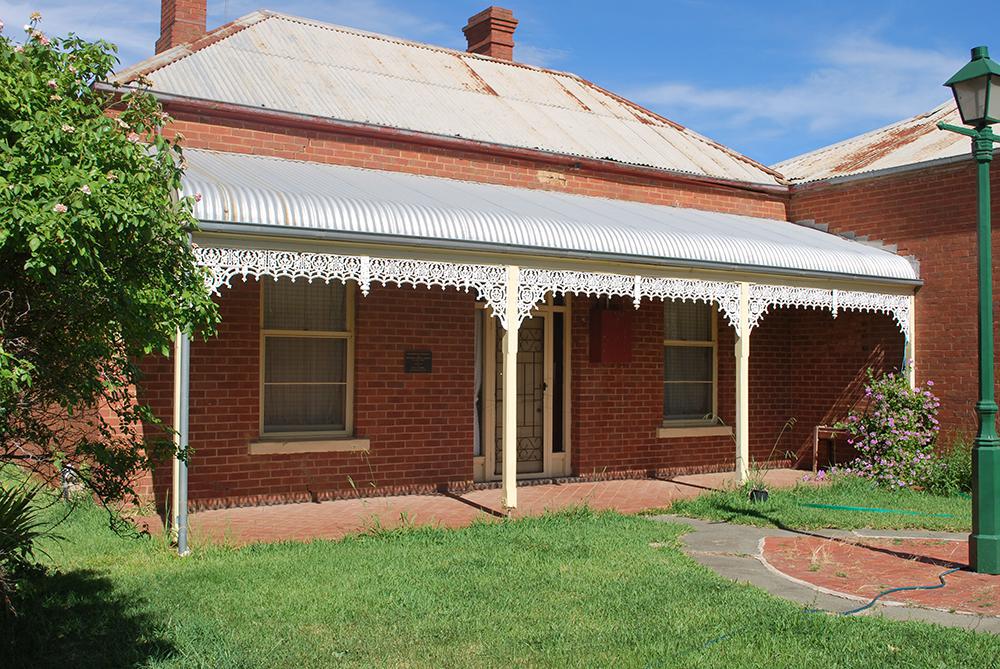}
    \end{subfigure}~%
    \begin{subfigure}{.4\linewidth}
    \includegraphics[width=\linewidth]{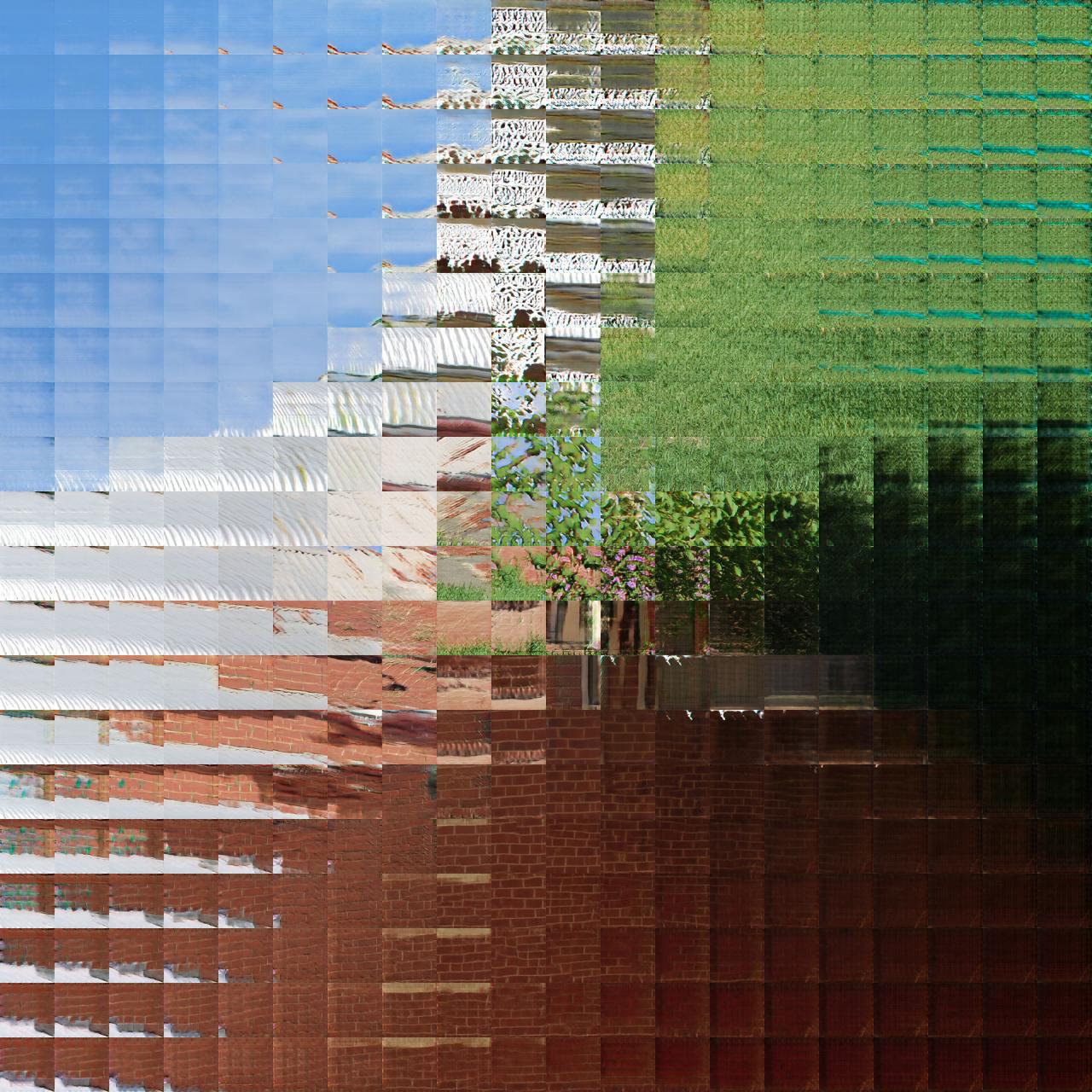}
    \end{subfigure}
    \caption{Merrigum House $3872\times2592$ photo and its learned 2D texture manifold using our model}
    \label{fig:house-2d-manifold}
    \vspace{-0.5cm}
\end{figure}

The learned manifold in section~\ref{sec:texture-manifold} was obtained from well prepared data. Real cases usually do not have clean data and require either expensive data preparation or unsupervised methods.
With minor corrections in data preparation pipeline, our model can learn texture manifolds from raw data such as a collection of high-resolution photos.
To cope with training texture manifolds on raw data, we suggest to construct $p^*(x, x')$ in \eqref{eq:VxxObj} with two crops from almost the same location with the stochastic procedure described in Algorithm~\ref{alg:unsupervized}. In Figure~\ref{fig:house-2d-manifold} we provide a manifold learned from House photo.

\begin{algorithm}[!t]
\begin{algorithmic}
\State Given: random image $I$, crop size $s$, window size $w>s$ (default $w=s\cdot1.1$)
\State $\bar x = \operatorname{RandomCrop}(I, \texttt{size=}(w, w))$
\State $x = \operatorname{RandomCrop}(\bar x, \texttt{size=}(s, s))$
\State $x' = \operatorname{RandomCrop}(\bar x, \texttt{size=}(s, s))$
\State Return: $(x, x')$
\end{algorithmic}
\caption{\label{alg:unsupervized} Obtaining $p^*(x, x')$ for the training on raw data}
\end{algorithm}
\subsection{Spatial Embeddings and Texture Detection\label{sec:spatial-embeddings-and-texture-detection}}

As described in sections~\ref{sec:learning-texture-manifolds-from-raw-data} and \ref{sec:texture-manifold}, our method can learn descriptive texture manifold from a collection of raw data in an unsupervised way. The obtained texture embeddings may be useful.
Consider a large input image $X$, \eg as the first in Figure~\ref{fig:teaser}, and the trained $G(z)$ and $E(x)$ on this image. Note that at the training stage encoder $E(x)$ is a fully convolutional network, followed by \textit{global} average pooling. Applied to $X$ as-is, the encoder's output would be "average" texture embedding for the whole image $X$. Replacing \textit{global} average pooling by \textit{spatial} average pooling with small kernel allows $E(X)$ to output texture embeddings for each receptive field in the input image $X$. We denote such modified encoder as $\widetilde E(x)$. 

$Z=\widetilde E(X)$ is a \textit{tensor} with spatial texture embeddings for X. They smoothly change along spatial dimensions as visualized by reconstructing them with generator $\tilde G(Z)$ (described in Appendix~\ref{app:network-architectures}) on the third picture in Figure~\ref{fig:teaser}.

One can take a reference patch $P$ with a texture (e.g., grass) and find similar textures in image $X$. This is illustrated in the last picture in Figure~\ref{fig:teaser}. We picked a patch $P$ with grass on it and constructed a heatmap $M$
\begin{equation}
    M_{ij}=\exp(-\alpha d(\widetilde E(X)_{ij}, E(P))^2),
\end{equation}
where $d(\cdot, \cdot)$ is Euclidean distance and $\alpha=3$ in our example. We then interpolated $M$ to the original size of $X$.

This example shows that $\widetilde E(x)$ allows using learned embeddings for other tasks that have no relation to texture generation. We believe supervised methods would benefit from adding additional features obtained in an unsupervised way.

\subsection{Memory Complexity}
In this section, we compare the scalability of DTS and our model with respect to dataset size. 
We denote the number of parameters as $M$, the dataset size as $N$. The number of parameters $M_{dts}$ of DTS model is\footnote{we use the official implementation from this \href{https://github.com/Yijunmaverick/MultiTextureSynthesis}{github page} in file "Model\_multi\_texture\_synthesis.lua"}
\begin{gather}
    M_{dts} \approx 34816\cdot(N + 71).
\end{gather}

We should note that DTS depends on $N$ which is the size of the whole dataset while the number of unique textures $n$ in the dataset can be much more smaller than $N$. Therefore, the method is not scalable to large datasets with duplications. 
To reduce memory complexity, DTS requires labeling. It will allow the method to find unique textures and set the size of the one-hot vector to the number of different texture types. 
Our model learns textures in an unsupervised way and instead of one-hot vector uses a low dimensional representation of textures. In Section~\ref{sec:learning-texture-manifolds-from-raw-data} we show that our method can detect different textures from high-resolution image without labeling. It means that our model complexity depends mostly on the number of unique textures in the dataset. The number of parameters of our model (the generator and the encoder) is
\begin{gather}
    M_{ours} \approx 25600(d + 336),
\end{gather}
where $d$ is the size of latent vector $z$ in our model, which consists of three parts $d = d^g + d^l + d^p$. 
In experiments, we show that the dimension $d = 26$ ($d^g = 2$, $d^l = 20$, $d^p = 4$) is sufficient to learn 116 unique textures. 

For example, let us consider a dataset of size 5000 which contains 100 unique textures with 50 variations per each one. Then for our model $d$ will be 26 and the number of parameters will be $9,27\cdot10^6$. Meanwhile, DTS will require $N = 5000$ and  $176,55\cdot10^6$ parameters. We see that in this case, our model memory consumption is less by approximately 20 times than DTS. 

\begin{table*}[!t]
\centering
\caption{KL divergence between real, our and the baseline distributions of statistics (permeability, Euler characteristic, and surface area) for size $160^3$. The standard deviation was computed using the bootstrap method with $1000$ resamples}
\ra{1.3}
\resizebox{\linewidth}{!}{%
\begin{tabular}{|c|c|c|c|c|c|c|}
\hline
\multirow{2}{*}{} & \multicolumn{2}{c|}{Permeability}                       & \multicolumn{2}{c|}{Euler characteristic}               & \multicolumn{2}{c|}{Surface area}                       \\ \cline{2-7} 
                  & $KL(p_{real}, p_{ours})$ & $KL(p_{real}, p_{baseline})$ & $KL(p_{real}, p_{ours})$ & $KL(p_{real}, p_{baseline})$ & $KL(p_{real}, p_{ours})$ & $KL(p_{real}, p_{baseline})$ \\ \hline
Ketton            & 5.06 $\pm$ 0.35          & 4.68 $\pm$ 0.56              & 3.66 $\pm$ 0.73          & \textbf{1.86 $\pm$ 0.42}     & \textbf{1.85 $\pm$ 0.62} & 7.73 $\pm$ 0.18              \\ \hline
Berea             & 0.49 $\pm$ 0.07          & 0.50 $\pm$ 0.12              & \textbf{0.34 $\pm$ 0.08} & 1.36 $\pm$ 0.25              & \textbf{0.33 $\pm$ 0.11} & 5.91 $\pm$ 0.54              \\ \hline
Doddington        & \textbf{0.42 $\pm$ 0.10} & 3.41 $\pm$ 1.68              & 2.65 $\pm$ 2.29          & 3.35 $\pm$ 1.13              & \textbf{4.83 $\pm$ 2.06} & 7.92 $\pm$ 0.27              \\ \hline
Estaillades       & \textbf{0.80 $\pm$ 0.24} & 3.41 $\pm$ 0.46              & 1.85 $\pm$ 0.29          & 2.05 $\pm$ 1.05              & \textbf{4.62 $\pm$ 0.66} & 6.93 $\pm$ 0.39              \\ \hline
Bentheimer        & \textbf{0.47 $\pm$ 0.08} & 1.38 $\pm$ 0.49              & \textbf{1.24 $\pm$ 0.41} & 3.44 $\pm$ 1.91              & 1.20 $\pm$ 0.73          & 1.25 $\pm$ 0.12              \\ \hline
\end{tabular}
}
\label{tab:permeability_KL_160}
\vspace{-0.5cm}
\end{table*}

\subsection{Application to 3D Porous Media Synthesis}
\label{sec:3d}

\begin{figure}[!h]
    \centering
    \begin{subfigure}[b]{0.09\textwidth}
        \includegraphics[width=\textwidth]{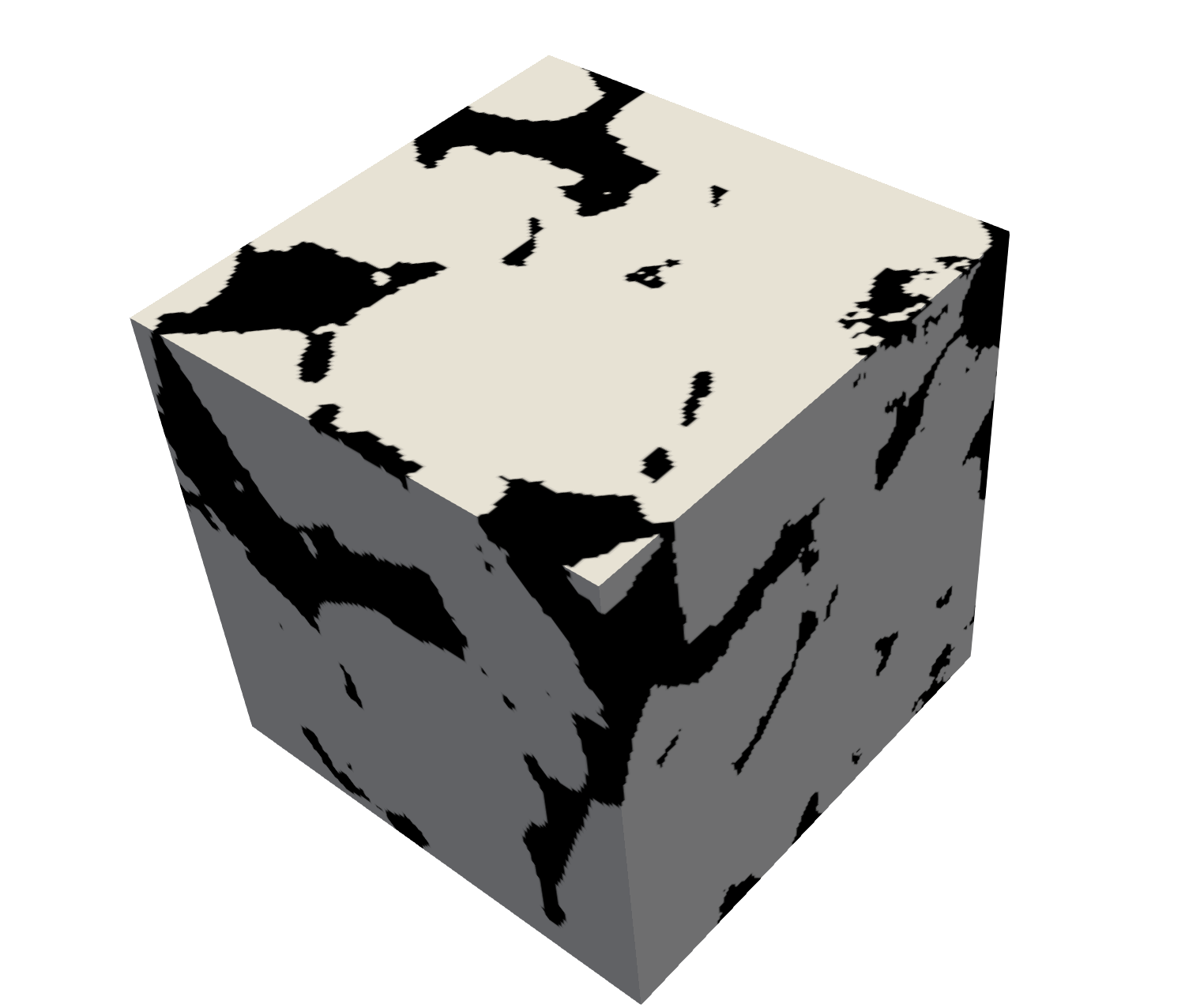}
        \caption{Real}
        \label{fig:berea_real}
    \end{subfigure}
    ~ 
    \begin{subfigure}[b]{0.09\textwidth}
        \includegraphics[width=\textwidth]{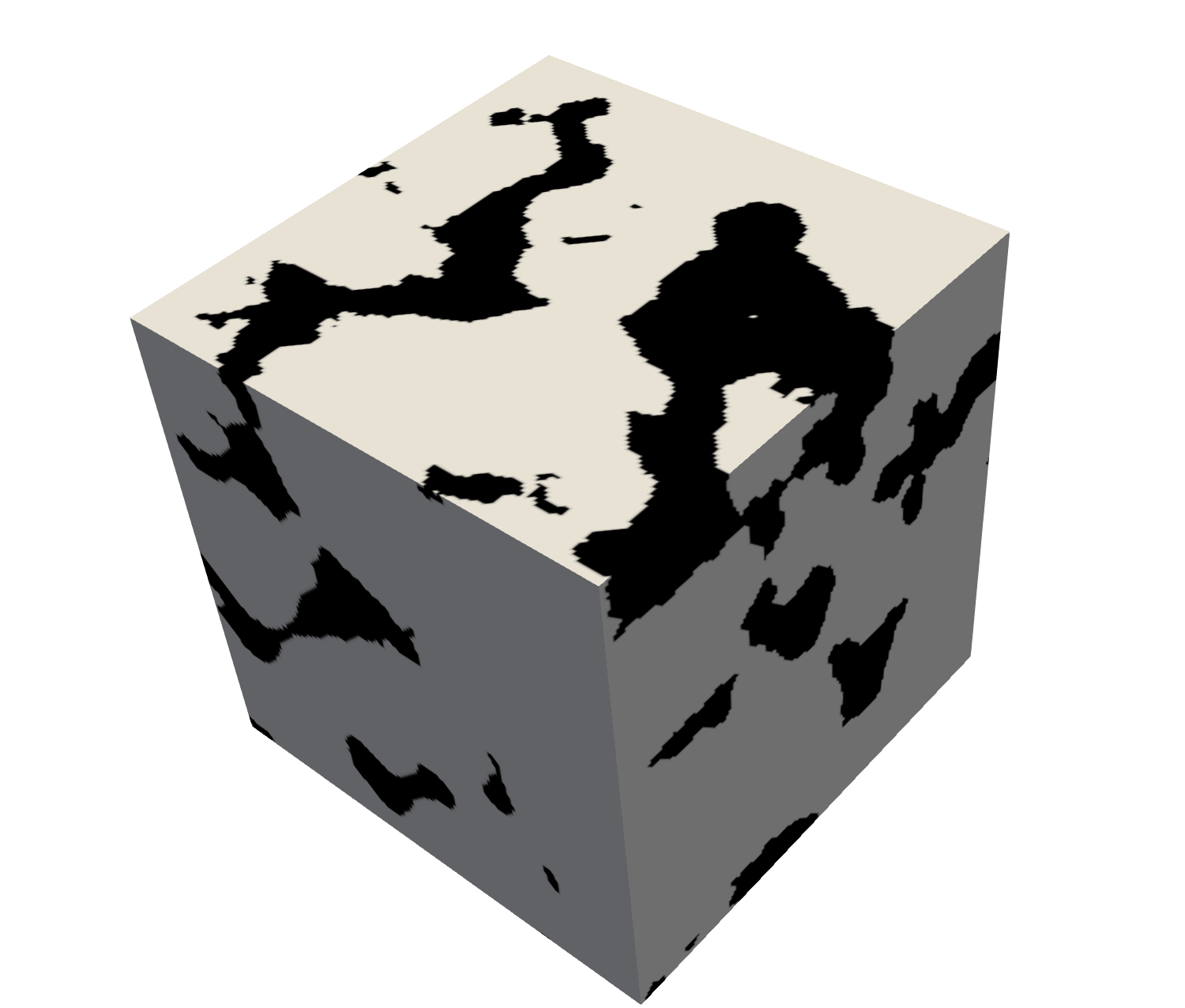}
        \caption{Ours}
        \label{fig:berea_ours}
    \end{subfigure}
    ~
    \begin{subfigure}[b]{0.09\textwidth}
        \includegraphics[width=\textwidth]{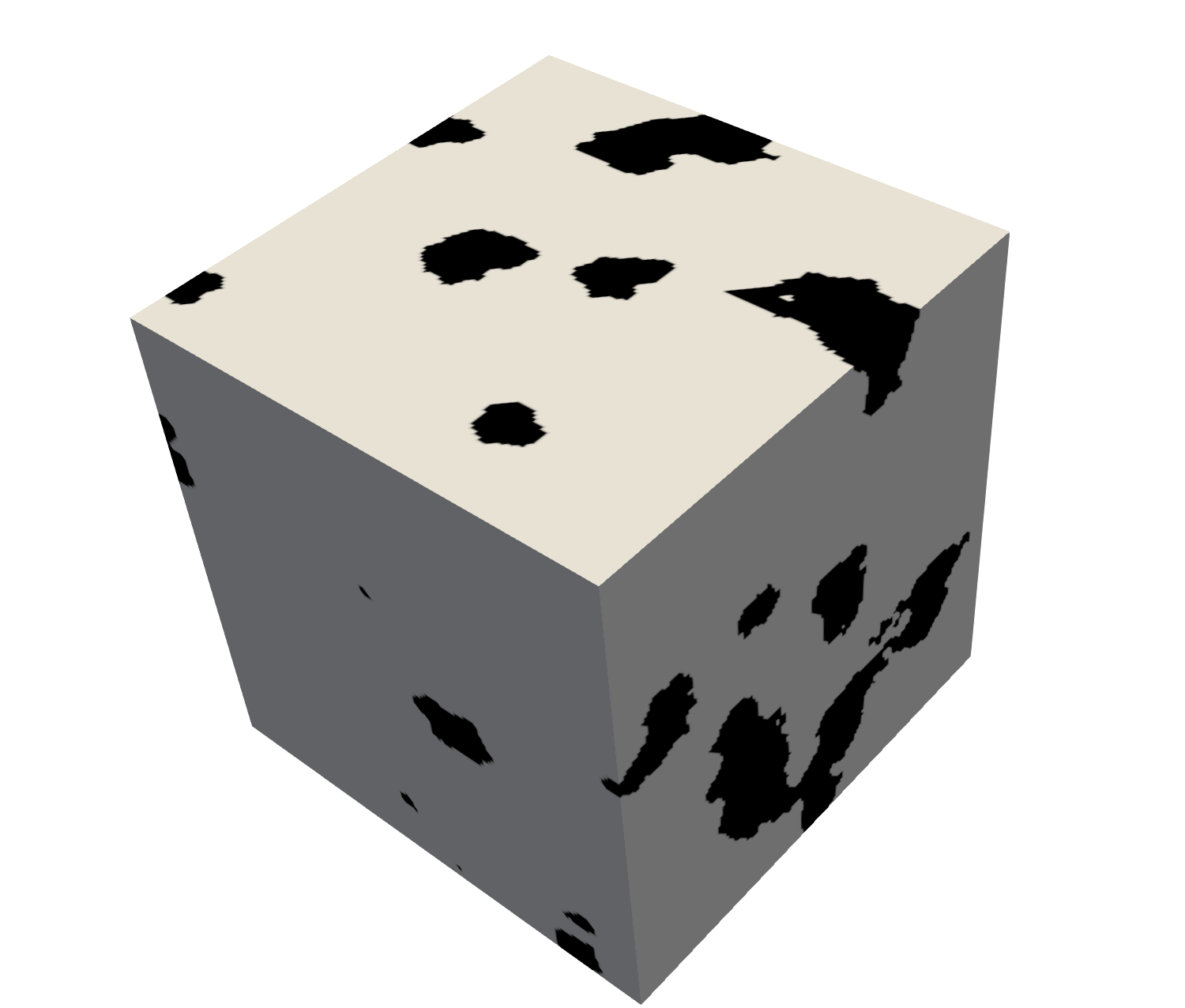}
        \caption{Baseline}
        \label{fig:berea_baseline}
    \end{subfigure}
    
    \caption{Real, synthetic (our model) and synthetic (baseline model) Berea samples of size $150^3$}\label{fig:berea_3d}
    \vspace{-0.6cm}
\end{figure}

In this section, we demonstrate the applicability of our model to the Digital Rock Physics. We trained our model on 3D Porous Media structures\footnote{All samples were taken from \href{http://www.imperial.ac.uk/earth-science/research/research-groups/perm/research/pore-scale-modelling/micro-ct-images-and-networks/}{this site}} (i.e. see Fig. \ref{fig:berea_real}) of five different types: Ketton, Berea, Doddington, Estaillades and Bentheimer. Each type of rock has an initial size $1000^3$ binary voxels. As the baseline, we considered Porous Media GANs \cite{mosser2017reconstruction}, which is deep convolutional GANs with 3D convolutional layers.


\begin{figure}[!h]
    \centering
        \includegraphics[width=0.2568\textwidth]{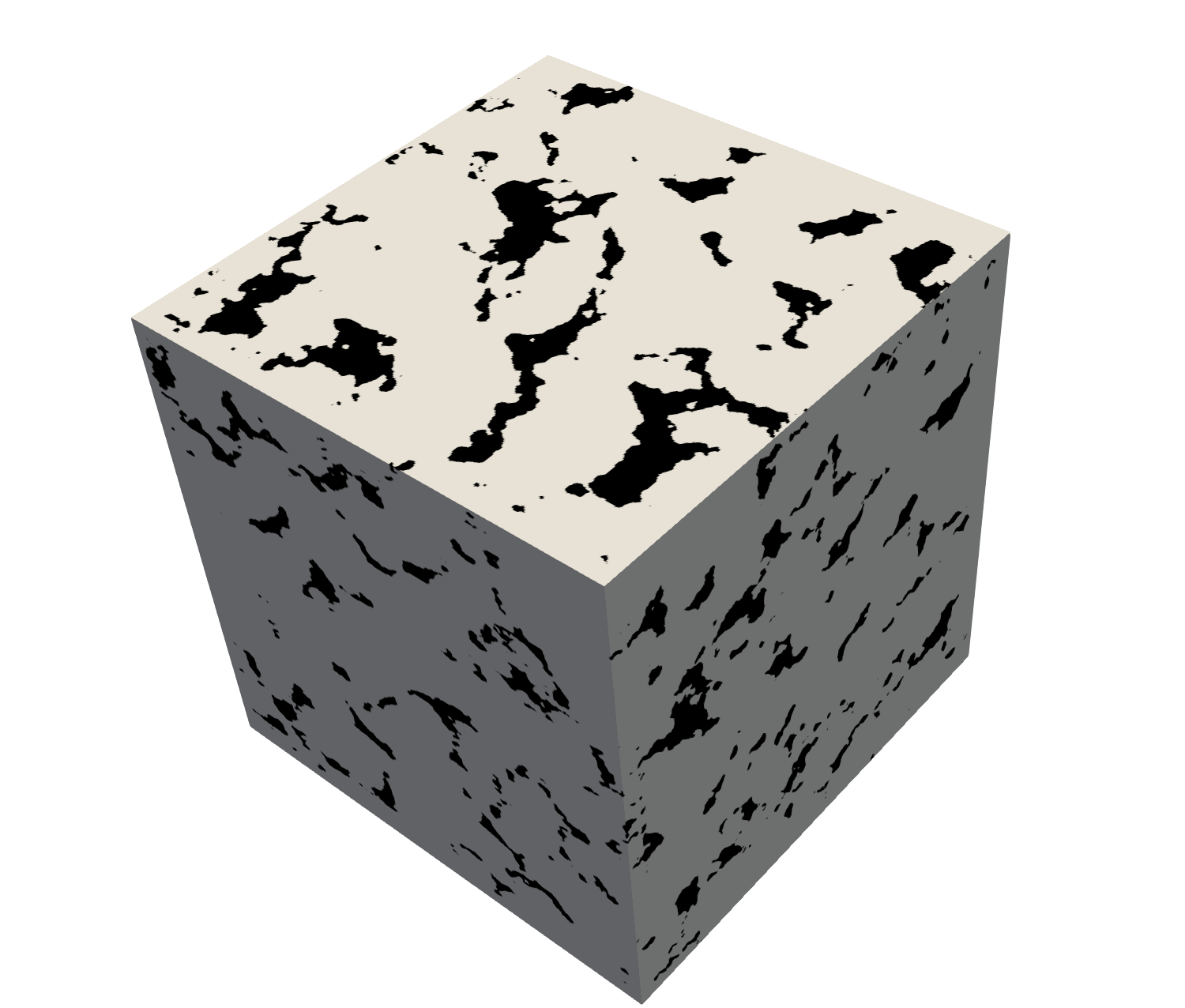}
        \caption{Synthetic Berea sample of size $428^3$ generated with our model}
        \label{fig:berea_ours_big}
        \vspace{-0.4cm}
\end{figure}

For the comparison of our model with real samples and the baseline samples, we use permeability statistics and two so-called Minkowski functionals \cite{minkowski}. The permeability is a measure of the ability of a porous material to allow fluids to pass through it. Minkowski functionals describe the morphology and topology of 3D binary structures. In our experiments, we used two functionals: Surface area and Euler characteristic. If the considered measures on synthetic samples are close to that on real ones, it will guarantee that the synthetic samples are valid for Digital Rock Physics applications.

We used the following experimental setup. We trained our model on random crops of size $160^3$ on all types of porous structures. We also trained five baseline models on each type separately. Then we generated $500$ synthetic samples of size $160^3$ of each type using our model and the baseline model. We also cropped $500$ samples of size $160^3$ from the real data. As a result, for each type of structure, we obtained three sets of objects: real, synthetic and baseline.

The visual result of the synthesis is presented in Fig. \ref{fig:berea_3d} for Berea. In the figure, there are three samples: real (i.e., cropped from the original big sample), ours and a sample of the baseline model. Other types of porous materials along with architecture details are presented in Appendix \ref{sec:appendix_3d}.
Because our model is fully convolutional, we can increase the generated sample size by expanding the spatial dimensions of the latent embedding $z$. We demonstrate the synthesized 3D porous media of size $428^3$ in Figure \ref{fig:berea_ours_big}.
Then, 
\begin{enumerate}
    \vspace{-0.22cm}
    \item For each real, synthetic and baseline objects we calculated three statistics: permeability, Surface Area and Euler characteristics.
    \vspace{-0.22cm}
    \item To measure the distance between distributions of statistics for real, our and baseline samples we approximated these distributions by discrete ones obtained using the histogram method with $50$ bins.
    \vspace{-0.22cm}
    
    \item Then for each statistic, we calculated KL divergence between the distributions of the statistic of a) real and our generated samples; b) real and baseline generated samples.
    \vspace{-0.22cm}
\end{enumerate}

The comparison of the KL divergences is presented at Tab. \ref{tab:permeability_KL_160} for the permeability and for Minkowski functionals. As we can see, our model performs better accordingly for most types of porous structures.

In this section, we showed the application of our model to Digital Rock Physics. Our model outperforms the baseline in most of the cases what proves its usefulness in solving real-world problems. Moreover, its critical advantage is the ability to generate multiple textures with the same model.

\section{Conclusion}

In this paper, we proposed a novel model for multi-texture synthesis. We showed it ensures full dataset coverage and can detect textures on images in the unsupervised setting. We provided a way to learn a manifold of training textures even from a collection of raw high-resolution photos. We also demonstrated that the proposed model applies to the real-world 3D texture synthesis problem: porous media generation. Our model outperforms the baseline by better reproducing physical properties of real data. In future work, we want to study the texture detection ability of our model and seek for its new applications. 




\paragraph{Acknowledgements.} Aibek Alanov, Max Kochurov, Dmitry Vetrov were supported by Samsung Research, Samsung Electronics. The work of E. Burnaev and D. Volkhonskiy was supported by The Ministry of Education and Science of Russian Federation, grant No.14.615.21.0004, grant code: RFMEFI61518X0004. The authors E. Burnaev and D. Volkhonskiy acknowledge the usage of the Skoltech CDISE HPC cluster Zhores for obtaining some results presented in this paper.

{\small
\bibliographystyle{ieee}
\bibliography{main}

\begin{thebibliography}{10}\itemsep=-1pt

\bibitem{arjovsky17a}
M.~Arjovsky, S.~Chintala, and L.~Bottou.
\newblock {W}asserstein generative adversarial networks.
\newblock In D.~Precup and Y.~W. Teh, editors, {\em Proceedings of the 34th
  International Conference on Machine Learning}, volume~70 of {\em Proceedings
  of Machine Learning Research}, pages 214--223, International Convention
  Centre, Sydney, Australia, 06--11 Aug 2017. PMLR.

\bibitem{bergmann2017learning}
U.~Bergmann, N.~Jetchev, and R.~Vollgraf.
\newblock Learning texture manifolds with the periodic spatial gan.
\newblock {\em ICML}, 2017.

\bibitem{brock2016neural}
A.~Brock, T.~Lim, J.~M. Ritchie, and N.~Weston.
\newblock Neural photo editing with introspective adversarial networks.
\newblock {\em ICLR}, 2017.

\bibitem{cimpoi2014dtw}
M.~Cimpoi, S.~Maji, I.~Kokkinos, S.~Mohamed, and A.~Vedaldi.
\newblock Describing textures in the wild.
\newblock In {\em Proceedings of the 2014 IEEE Conference on Computer Vision
  and Pattern Recognition}, CVPR '14, pages 3606--3613, Washington, DC, USA,
  2014. IEEE Computer Society.

\bibitem{donahue2016adversarial}
J.~Donahue, P.~Kr{\"a}henb{\"u}hl, and T.~Darrell.
\newblock Adversarial feature learning.
\newblock {\em ICLR}, 2017.

\bibitem{dumoulin2017learned}
V.~Dumoulin, J.~Shlens, and M.~Kudlur.
\newblock A learned representation for artistic style.
\newblock {\em Proc. of ICLR}, 2017.

\bibitem{efros2001image}
A.~A. Efros and W.~T. Freeman.
\newblock Image quilting for texture synthesis and transfer.
\newblock In {\em Proceedings of the 28th annual conference on Computer
  graphics and interactive techniques}, pages 341--346. ACM, 2001.

\bibitem{efros1999texture}
A.~A. Efros and T.~K. Leung.
\newblock Texture synthesis by non-parametric sampling.
\newblock In {\em iccv}, page 1033. IEEE, 1999.

\bibitem{frigo2016split}
O.~Frigo, N.~Sabater, J.~Delon, and P.~Hellier.
\newblock Split and match: Example-based adaptive patch sampling for
  unsupervised style transfer.
\newblock In {\em Proceedings of the IEEE Conference on Computer Vision and
  Pattern Recognition}, pages 553--561, 2016.

\bibitem{gatys2015texture}
L.~Gatys, A.~S. Ecker, and M.~Bethge.
\newblock Texture synthesis using convolutional neural networks.
\newblock In {\em Advances in Neural Information Processing Systems}, pages
  262--270, 2015.

\bibitem{gatys2016preserving}
L.~A. Gatys, M.~Bethge, A.~Hertzmann, and E.~Shechtman.
\newblock Preserving color in neural artistic style transfer.
\newblock {\em arXiv preprint arXiv:1606.05897}, 2016.

\bibitem{gatys2016image}
L.~A. Gatys, A.~S. Ecker, and M.~Bethge.
\newblock Image style transfer using convolutional neural networks.
\newblock In {\em Proceedings of the IEEE Conference on Computer Vision and
  Pattern Recognition}, pages 2414--2423, 2016.

\bibitem{goodfellow2014generative}
I.~Goodfellow, J.~Pouget-Abadie, M.~Mirza, B.~Xu, D.~Warde-Farley, S.~Ozair,
  A.~Courville, and Y.~Bengio.
\newblock Generative adversarial nets.
\newblock In {\em Advances in neural information processing systems}, pages
  2672--2680, 2014.

\bibitem{heeger1995pyramid}
D.~J. Heeger and J.~R. Bergen.
\newblock Pyramid-based texture analysis/synthesis.
\newblock In {\em Proceedings of the 22nd annual conference on Computer
  graphics and interactive techniques}, pages 229--238. ACM, 1995.

\bibitem{jetchev2016texture}
N.~Jetchev, U.~Bergmann, and R.~Vollgraf.
\newblock Texture synthesis with spatial generative adversarial networks.
\newblock {\em arXiv preprint arXiv:1611.08207}, 2016.

\bibitem{johnson2016perceptual}
J.~Johnson, A.~Alahi, and L.~Fei-Fei.
\newblock Perceptual losses for real-time style transfer and super-resolution.
\newblock In {\em European Conference on Computer Vision}, pages 694--711.
  Springer, 2016.

\bibitem{kingma2015adam}
D.~P. Kingma and J.~Ba.
\newblock Adam: A method for stochastic optimization.
\newblock In {\em International Conference on Learning Representations (ICLR)},
  2015.

\bibitem{kingma2013auto}
D.~P. Kingma and M.~Welling.
\newblock Auto-encoding variational bayes.
\newblock {\em ICLR}, 2014.

\bibitem{kwatra2003graphcut}
V.~Kwatra, A.~Sch{\"o}dl, I.~Essa, G.~Turk, and A.~Bobick.
\newblock Graphcut textures: image and video synthesis using graph cuts.
\newblock {\em ACM Transactions on Graphics (ToG)}, 22(3):277--286, 2003.

\bibitem{minkowski}
D.~Legland, K.~Ki{\^e}u, and M.-F. Devaux.
\newblock Computation of minkowski measures on 2d and 3d binary images.
\newblock {\em Image Analysis \& Stereology}, 26(2):83--92, 2011.

\bibitem{li2017alice}
C.~Li, H.~Liu, C.~Chen, Y.~Pu, L.~Chen, R.~Henao, and L.~Carin.
\newblock Alice: Towards understanding adversarial learning for joint
  distribution matching.
\newblock In {\em Advances in Neural Information Processing Systems}, pages
  5495--5503, 2017.

\bibitem{li2016combining}
C.~Li and M.~Wand.
\newblock Combining markov random fields and convolutional neural networks for
  image synthesis.
\newblock In {\em Proceedings of the IEEE Conference on Computer Vision and
  Pattern Recognition}, pages 2479--2486, 2016.

\bibitem{li2016precomputed}
C.~Li and M.~Wand.
\newblock Precomputed real-time texture synthesis with markovian generative
  adversarial networks.
\newblock In {\em European Conference on Computer Vision}, pages 702--716.
  Springer, 2016.

\bibitem{li2017diversified}
Y.~Li, C.~Fang, J.~Yang, Z.~Wang, X.~Lu, and M.-H. Yang.
\newblock Diversified texture synthesis with feed-forward networks.
\newblock In {\em Proc. CVPR}, 2017.

\bibitem{makhzani2015adversarial}
A.~Makhzani, J.~Shlens, N.~Jaitly, I.~Goodfellow, and B.~Frey.
\newblock Adversarial autoencoders.
\newblock {\em ICLR}, 2016.

\bibitem{miyato2018spectral}
T.~Miyato, T.~Kataoka, M.~Koyama, and Y.~Yoshida.
\newblock Spectral normalization for generative adversarial networks.
\newblock {\em ICLR}, 2018.

\bibitem{mosser2017reconstruction}
L.~Mosser, O.~Dubrule, and M.~J. Blunt.
\newblock Reconstruction of three-dimensional porous media using generative
  adversarial neural networks.
\newblock {\em Physical Review E}, 96(4):043309, 2017.

\bibitem{portilla2000parametric}
J.~Portilla and E.~P. Simoncelli.
\newblock A parametric texture model based on joint statistics of complex
  wavelet coefficients.
\newblock {\em International journal of computer vision}, 40(1):49--70, 2000.

\bibitem{radford2015unsupervised}
A.~Radford, L.~Metz, and S.~Chintala.
\newblock Unsupervised representation learning with deep convolutional
  generative adversarial networks.
\newblock {\em arXiv preprint arXiv:1511.06434}, 2015.

\bibitem{rezende2014stochastic}
D.~J. Rezende, S.~Mohamed, and D.~Wierstra.
\newblock Stochastic backpropagation and approximate inference in deep
  generative models.
\newblock {\em ICML}, 2014.

\bibitem{rosca2017variational}
M.~Rosca, B.~Lakshminarayanan, D.~Warde-Farley, and S.~Mohamed.
\newblock Variational approaches for auto-encoding generative adversarial
  networks.
\newblock {\em arXiv preprint arXiv:1706.04987}, 2017.

\bibitem{szegedy2016inception}
C.~Szegedy, V.~Vanhoucke, S.~Ioffe, J.~Shlens, and Z.~Wojna.
\newblock Rethinking the inception architecture for computer vision.
\newblock In {\em 2016 {IEEE} Conference on Computer Vision and Pattern
  Recognition, {CVPR} 2016, Las Vegas, NV, USA, June 27-30, 2016}, pages
  2818--2826, 2016.

\bibitem{thanh2019improving}
H.~Thanh-Tung, T.~Tran, and S.~Venkatesh.
\newblock Improving generalization and stability of generative adversarial
  networks.
\newblock {\em arXiv preprint arXiv:1902.03984}, 2019.

\bibitem{titsias2014doubly}
M.~Titsias and M.~L{\'a}zaro-Gredilla.
\newblock Doubly stochastic variational bayes for non-conjugate inference.
\newblock In {\em International Conference on Machine Learning}, pages
  1971--1979, 2014.

\bibitem{ulyanov2016texture}
D.~Ulyanov, V.~Lebedev, A.~Vedaldi, and V.~S. Lempitsky.
\newblock Texture networks: Feed-forward synthesis of textures and stylized
  images.
\newblock In {\em ICML}, pages 1349--1357, 2016.

\bibitem{ulyanovinstance}
D.~Ulyanov, A.~Vedaldi, and V.~Lempitsky.
\newblock Instance normalization: the missing ingredient for fast stylization.
  corr abs/1607.0 (2016).

\bibitem{ulyanov2017improved}
D.~Ulyanov, A.~Vedaldi, and V.~Lempitsky.
\newblock Improved texture networks: Maximizing quality and diversity in
  feed-forward stylization and texture synthesis.
\newblock In {\em Proceedings of the IEEE Conference on Computer Vision and
  Pattern Recognition}, pages 6924--6932, 2017.

\bibitem{UlyanovVL18}
D.~Ulyanov, A.~Vedaldi, and V.~S. Lempitsky.
\newblock It takes (only) two: Adversarial generator-encoder networks.
\newblock In {\em {AAAI}}. {AAAI} Press, 2018.

\bibitem{volkhonskiy2019reconstruction}
D.~Volkhonskiy, E.~Muravleva, O.~Sudakov, D.~Orlov, B.~Belozerov, E.~Burnaev,
  and D.~Koroteev.
\newblock Reconstruction of 3d porous media from 2d slices.
\newblock {\em arXiv preprint arXiv:1901.10233}, 2019.

\bibitem{wei2000fast}
L.-Y. Wei and M.~Levoy.
\newblock Fast texture synthesis using tree-structured vector quantization.
\newblock In {\em Proceedings of the 27th annual conference on Computer
  graphics and interactive techniques}, pages 479--488. ACM
  Press/Addison-Wesley Publishing Co., 2000.

\bibitem{xian2018texturegan}
W.~Xian, P.~Sangkloy, V.~Agrawal, A.~Raj, J.~Lu, C.~Fang, F.~Yu, and J.~Hays.
\newblock Texturegan: Controlling deep image synthesis with texture patches.
\newblock In {\em Proceedings of the IEEE Conference on Computer Vision and
  Pattern Recognition}, pages 8456--8465, 2018.

\bibitem{zhou2018non}
Y.~Zhou, Z.~Zhu, X.~Bai, D.~Lischinski, D.~Cohen-Or, and H.~Huang.
\newblock Non-stationary texture synthesis by adversarial expansion.
\newblock {\em arXiv preprint arXiv:1805.04487}, 2018.

\bibitem{CycleGAN2017}
J.-Y. Zhu, T.~Park, P.~Isola, and A.~A. Efros.
\newblock Unpaired image-to-image translation using cycle-consistent
  adversarial networks.
\newblock In {\em Computer Vision (ICCV), 2017 IEEE International Conference
  on}, 2017.

\end{thebibliography}
}

\newpage\phantom{blabla}

 

\onecolumn
\appendix
\section*{\huge Appendix}
\section{Additional Experiments}
\subsection{Results on Datasets: Scaly, Braided, Honeycomb, Striped\label{app:attachments}} 
In this appendix section, we provide results for all datasets mentioned in the main text as \textit{attached files}. The below listing of files is a reference to these files.
\begin{enumerate}
    \item Scaly Dataset
    \begin{itemize}
        \item \texttt{scaly\_ours\_2D\_samples.jpg} -- $16\times16$ grid of samples images for our model with $d^g=2$
        \item \texttt{scaly\_ours\_2D\_recon.jpg} -- reconstructions for every texture present in the training dataset for our model with $d^g=2$
        \item \texttt{scaly\_ours\_40D\_samples.jpg} -- $16\times16$ grid of samples images for our model with $d^g=40$
        \item \texttt{scaly\_ours\_40D\_recon.jpg} -- reconstructions for every texture present in the training dataset for our model with $d^g=40$
        \item \texttt{movie-2d-plane-ours.gif} -- a visualization for the training process of the 2D latent space for our model
        \item \texttt{movie-2d-plane-psgan.gif} -- a visualization for the training process of the 2D latent space for PSGAN model
    \end{itemize}
    \item Braided Dataset
    \begin{itemize}
        \item \texttt{braided\_ours\_2D\_samples.jpg} -- $16\times16$ grid of samples images for our model with $d^g=2$
        \item \texttt{braided\_ours\_2D\_recon.jpg} -- reconstructions for every texture present in the training dataset for our model with $d^g=2$
        \item \texttt{braided\_ours\_40D\_samples.jpg} -- $16\times16$ grid of samples images for our model with $d^g=40$
        \item \texttt{braided\_ours\_40D\_recon.jpg} -- reconstructions for every texture present in the training dataset for our model with $d^g=40$
    \end{itemize}
    \item Honeycomb Dataset
    \begin{itemize}
        \item \texttt{honeycomb\_ours\_2D\_samples.jpg} -- $16\times16$ grid of samples images for our model with $d^g=2$
        \item \texttt{honeycomb\_ours\_2D\_recon.jpg} -- reconstructions for every texture present in the training dataset for our model with $d^g=2$
        \item \texttt{honeycomb\_ours\_40D\_samples.jpg} -- $16\times16$ grid of samples images for our model with $d^g=40$
        \item \texttt{honeycomb\_ours\_40D\_recon.jpg} -- reconstructions for every texture present in the training dataset for our model with $d^g=40$
    \end{itemize}
    \item Striped Dataset
    \begin{itemize}
        \item \texttt{striped\_ours\_2D\_samples.jpg} -- $16\times16$ grid of samples images for our model with $d^g=2$
        \item \texttt{striped\_ours\_2D\_recon.jpg} -- reconstructions for every texture present in the training dataset for our model with $d^g=2$
    \end{itemize}
\end{enumerate}
\subsection{Results on a Collection of Raw Images}
\begin{figure}[!h]
    \centering
    \begin{subfigure}{0.45\linewidth}
    \includegraphics[width=\linewidth]{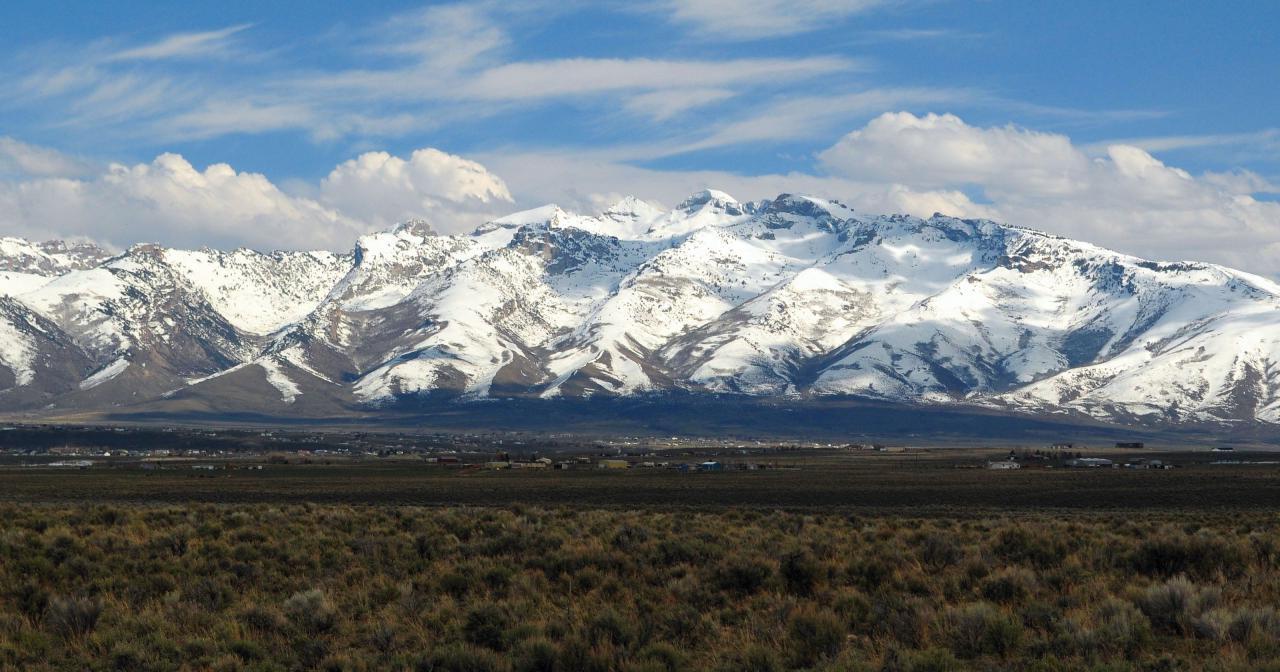}
    \caption{Size $1680\times3200$}
    \end{subfigure}~%
    \begin{subfigure}{0.45\linewidth}
    \includegraphics[width=\linewidth]{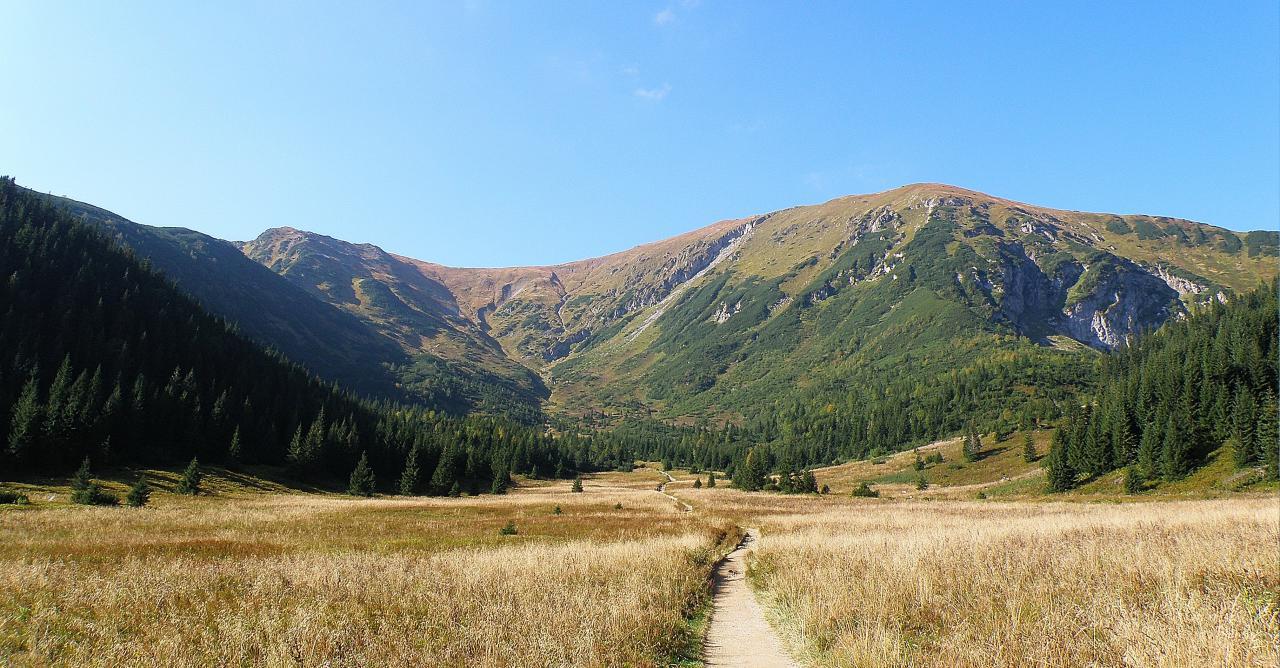}
    \caption{Size $1904\times3648$}
    \end{subfigure}
    \begin{subfigure}{0.45\linewidth}
    \includegraphics[width=\linewidth]{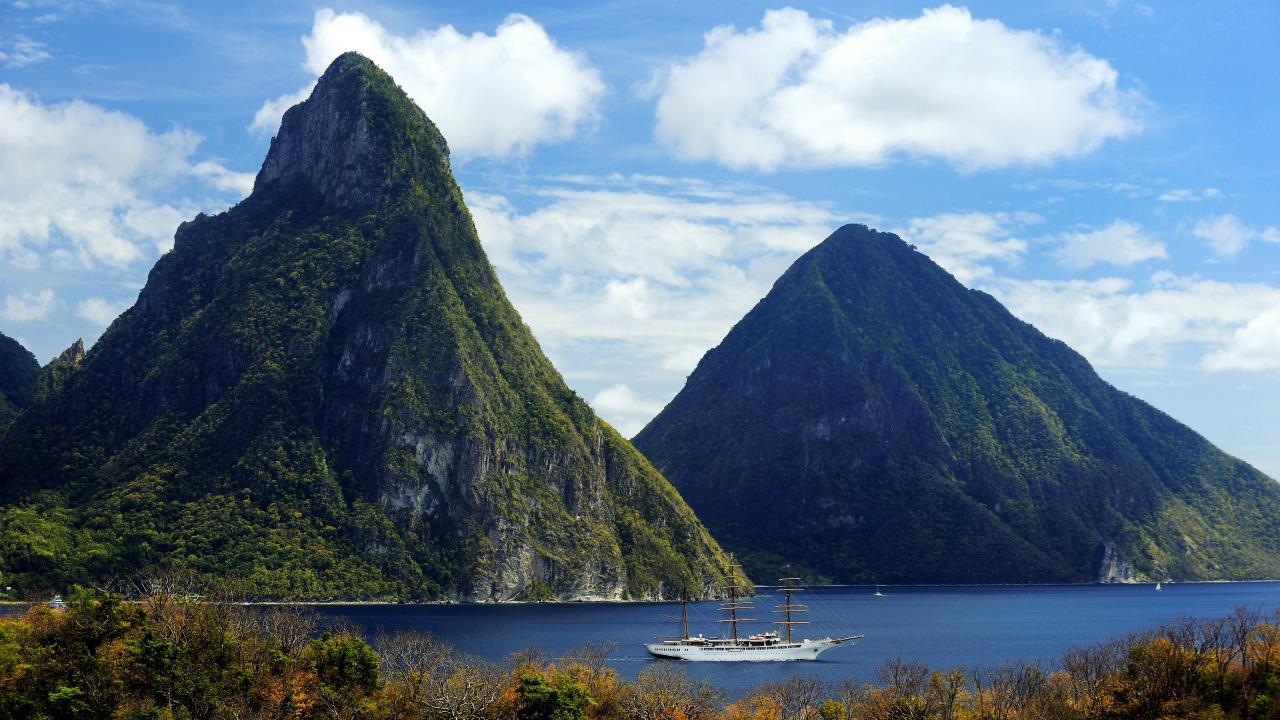}
    \caption{Size $2394\times 4256$}
    \end{subfigure}~%
    \begin{subfigure}{0.45\linewidth}
    \includegraphics[width=\linewidth]{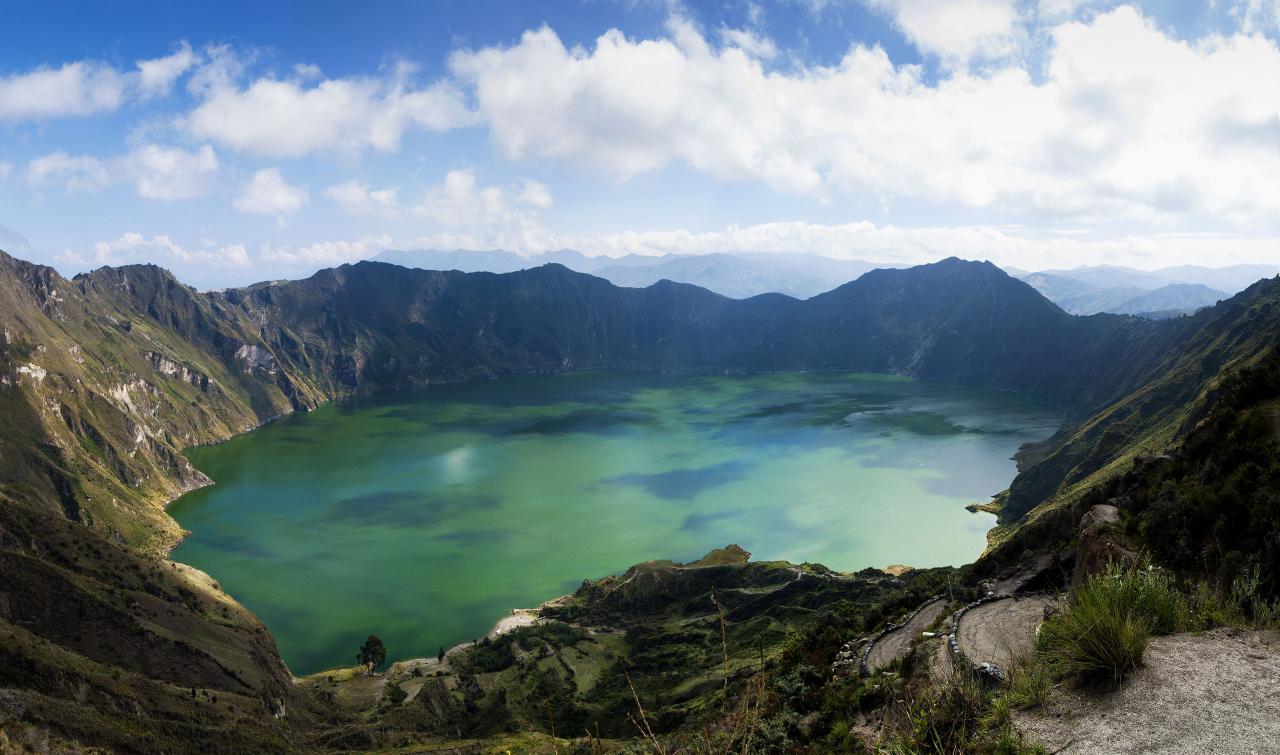}
    \caption{Size $2424\times 4107$}
    \end{subfigure}
    \begin{subfigure}{0.45\linewidth}
    \includegraphics[width=\linewidth]{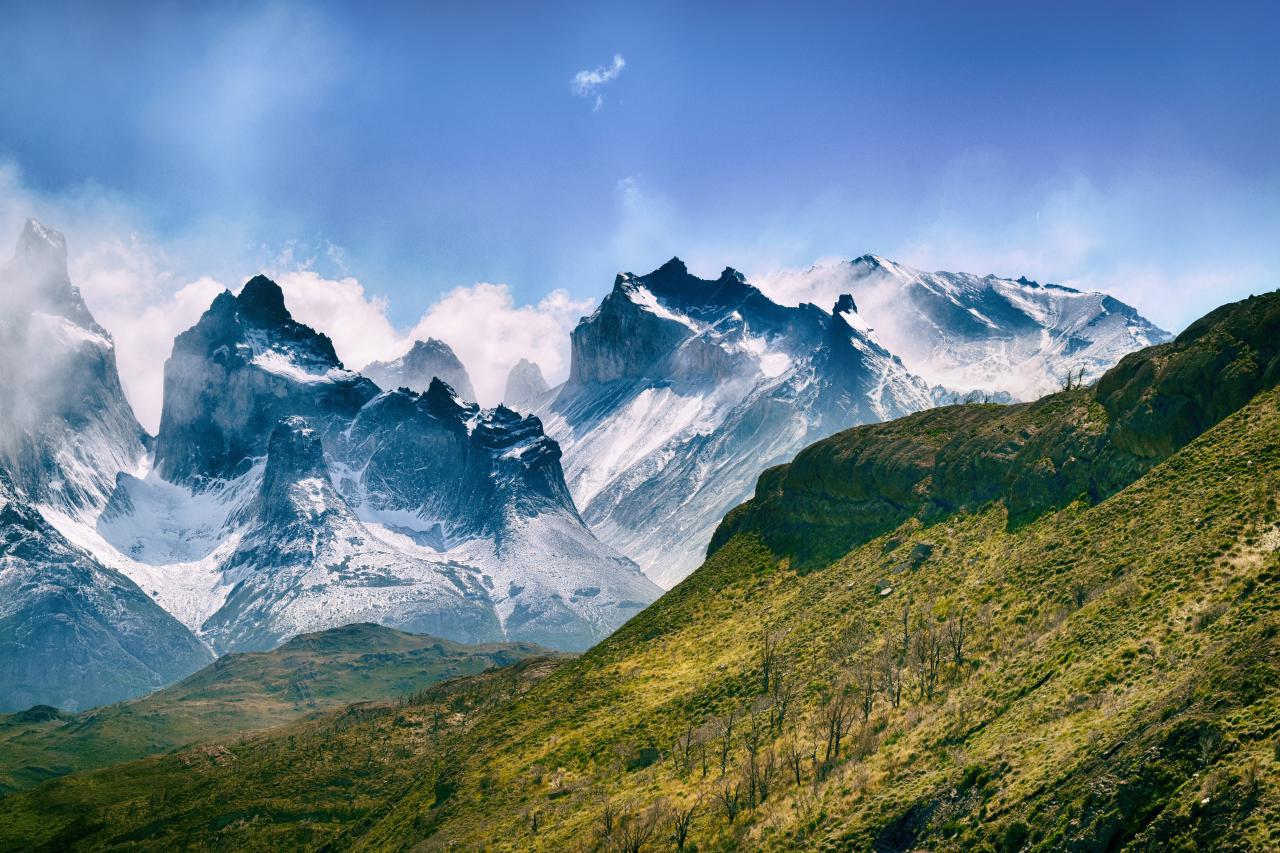}
    \caption{Size $2667\times 4000$}
    \end{subfigure}
    \caption{Training images}
    \label{fig:raw_data}
    \vspace{-0.7cm}
\end{figure}

\begin{figure}
    \centering
    \includegraphics[width=\linewidth]{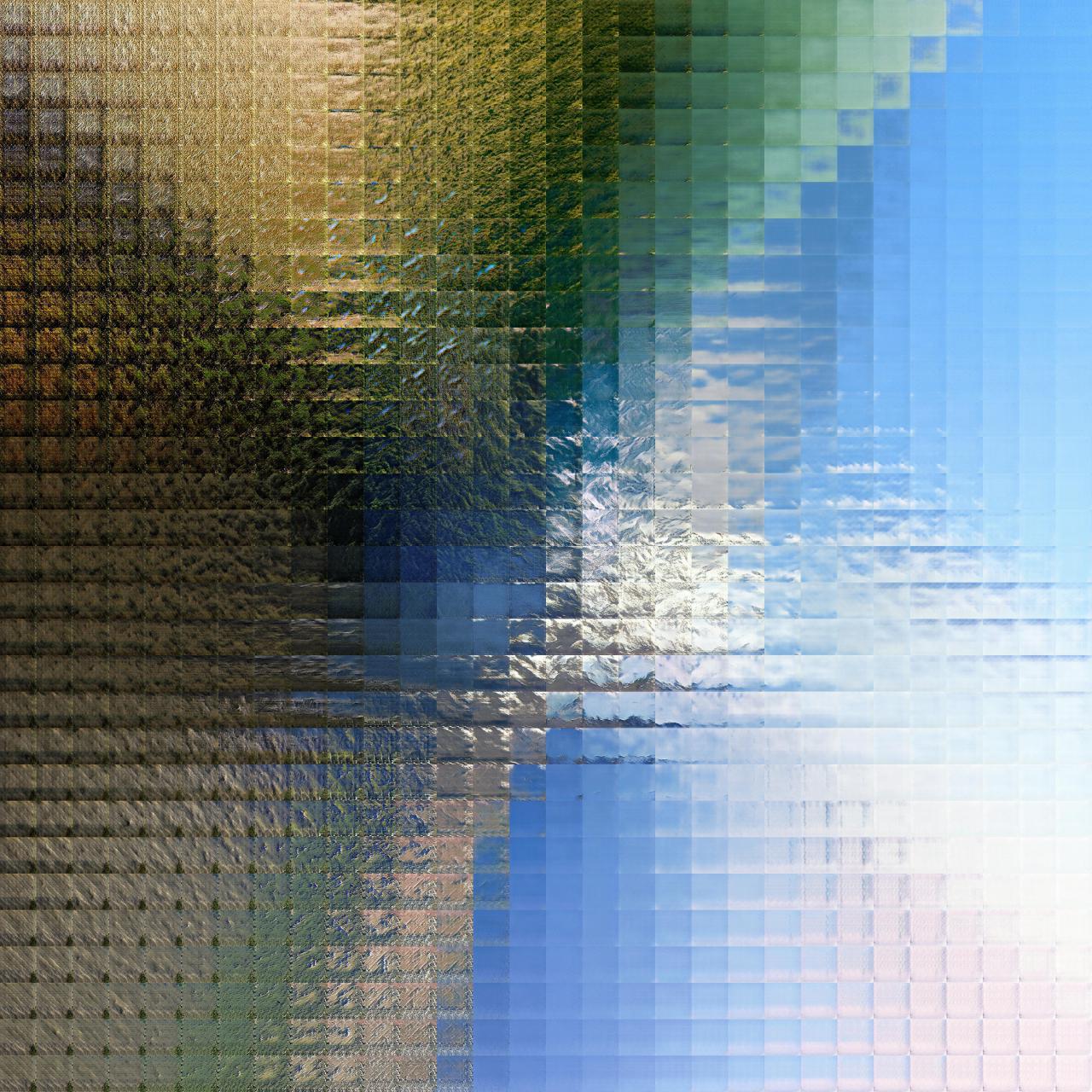}
    \caption{Learned manifold from training images}
    \label{fig:raw_manifold}
\end{figure}

\begin{figure}[!h]
    \centering
    \begin{subfigure}{0.45\linewidth}
    \includegraphics[width=\linewidth]{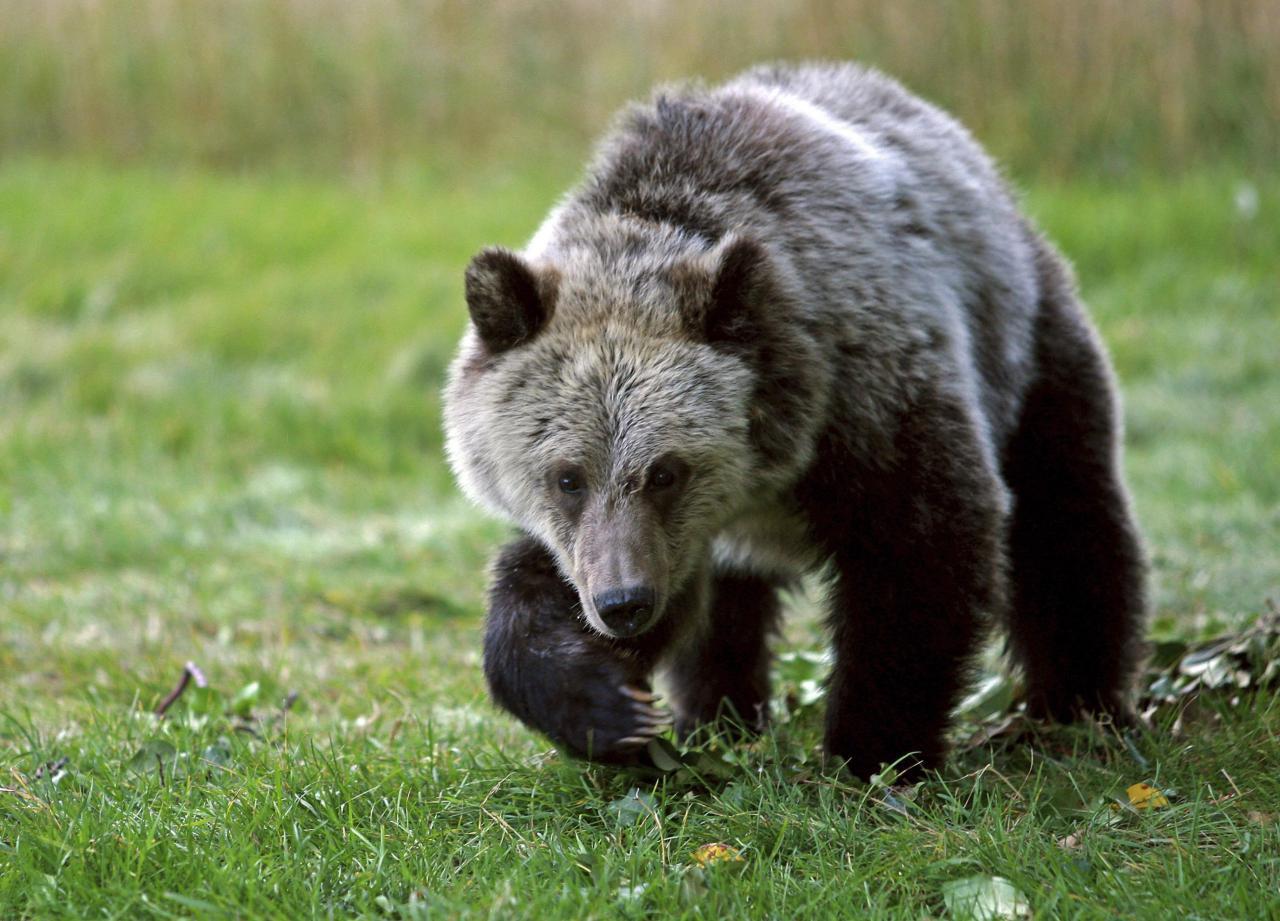}
    \caption{Initial image ($1912\times2657$)}
    \end{subfigure}~%
    \begin{subfigure}{0.45\linewidth}
    \includegraphics[width=\linewidth]{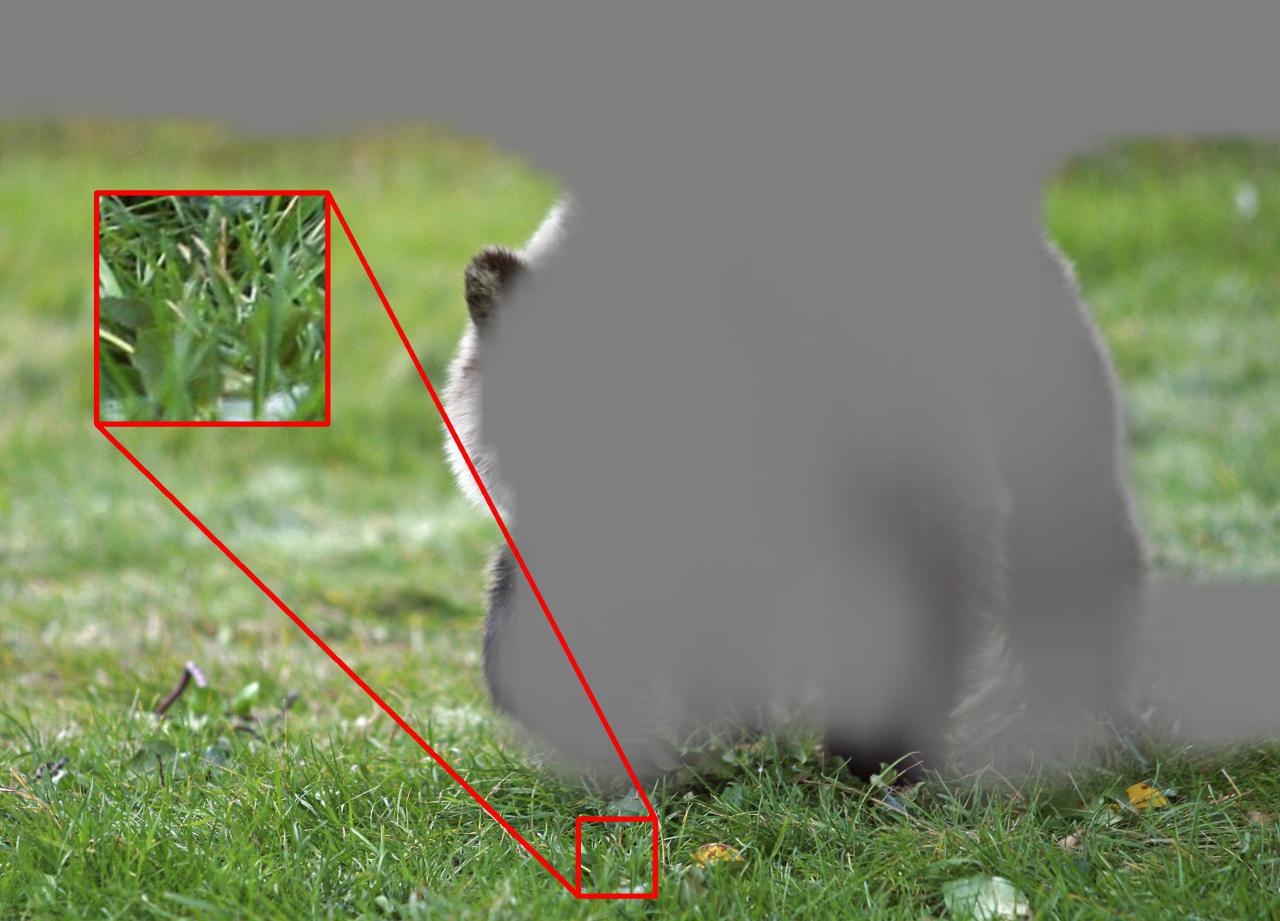}
    \caption{Grass texture is detected}
    \end{subfigure}
    \begin{subfigure}{0.45\linewidth}
    \includegraphics[width=\linewidth]{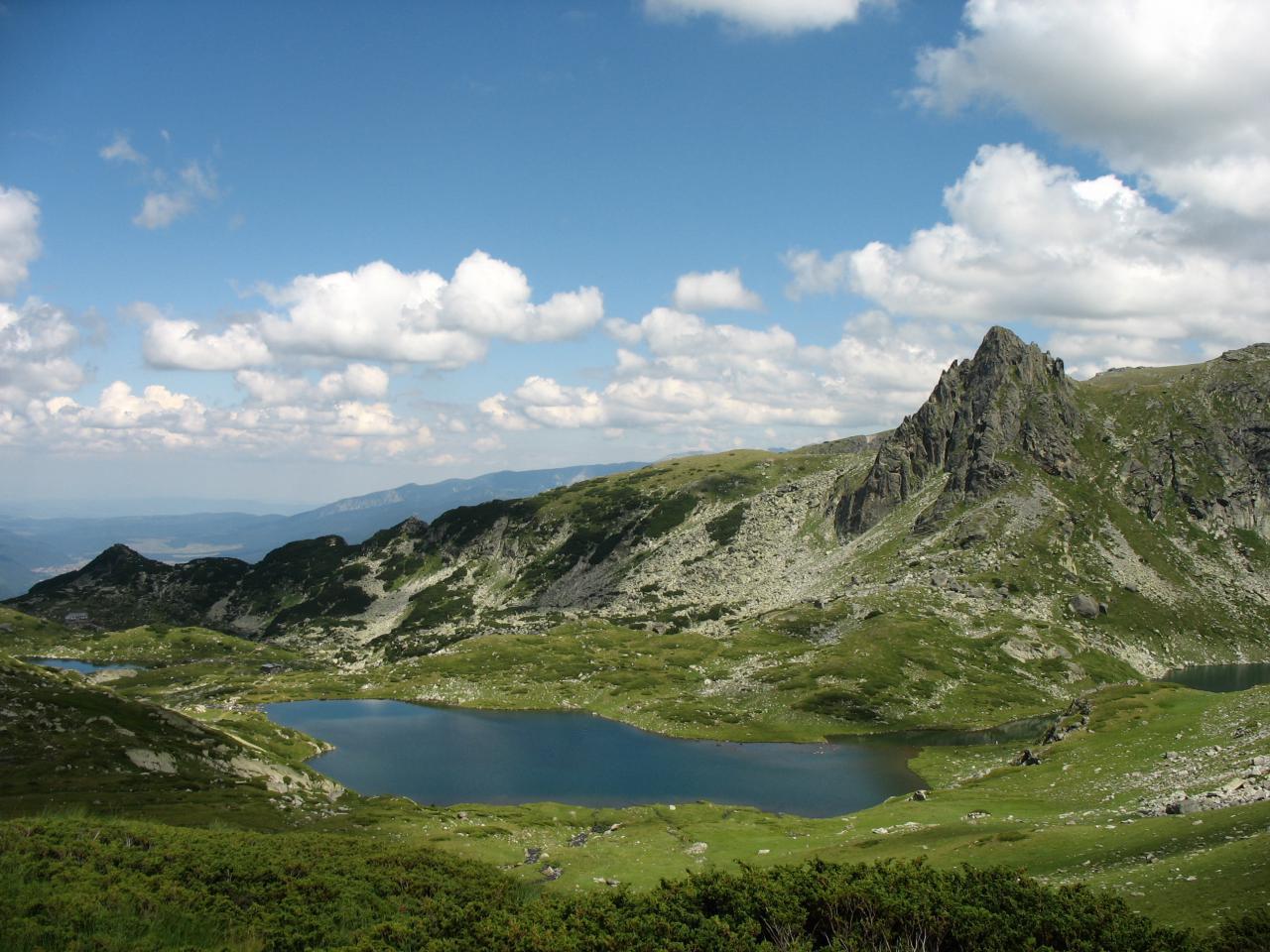}
    \caption{Initial image ($1944\times 2592$)}
    \end{subfigure}~%
    \begin{subfigure}{0.45\linewidth}
    \includegraphics[width=\linewidth]{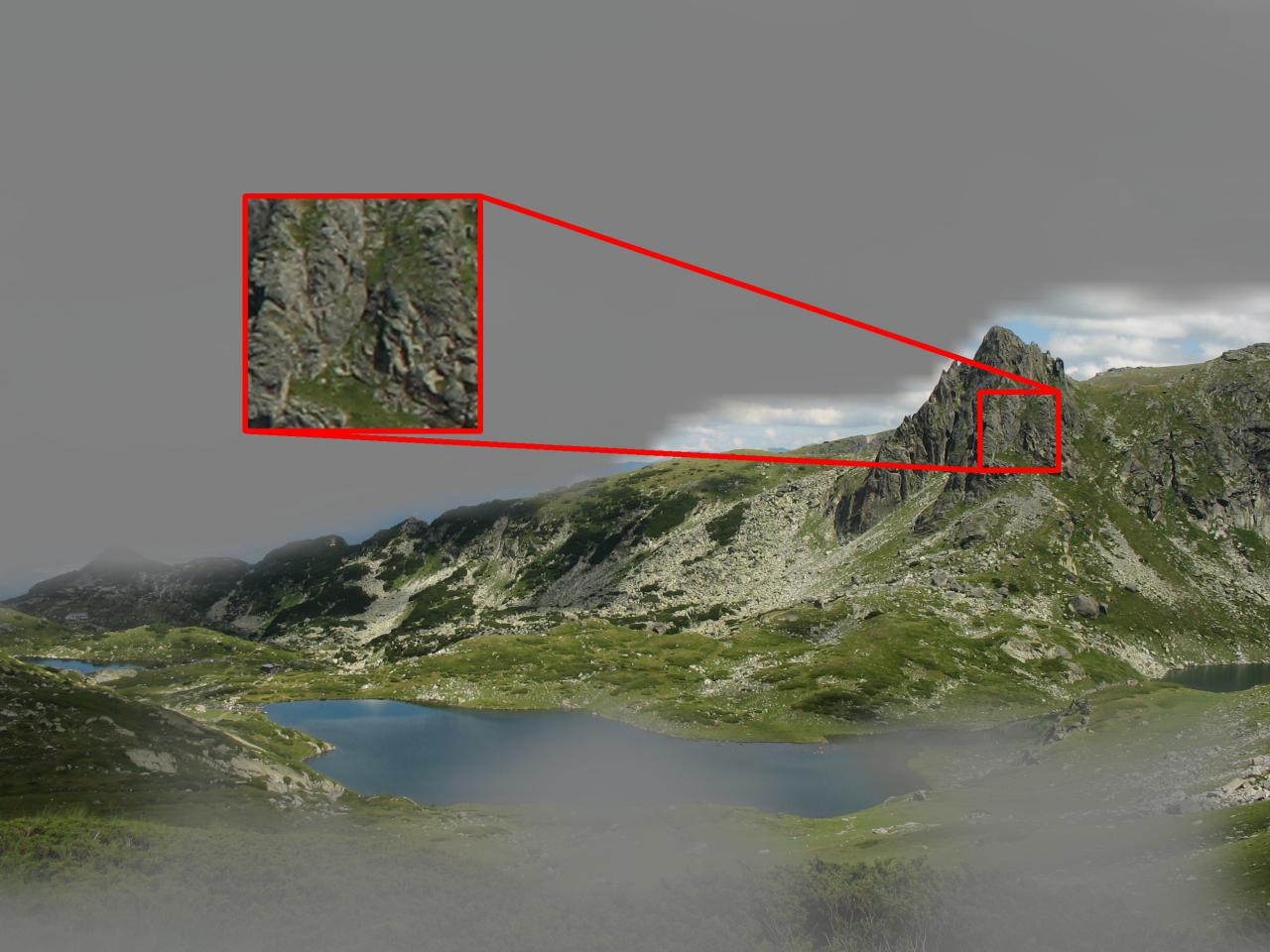}
    \caption{Mountain texture is detected}
    \end{subfigure}
    \begin{subfigure}{0.45\linewidth}
    \includegraphics[width=\linewidth]{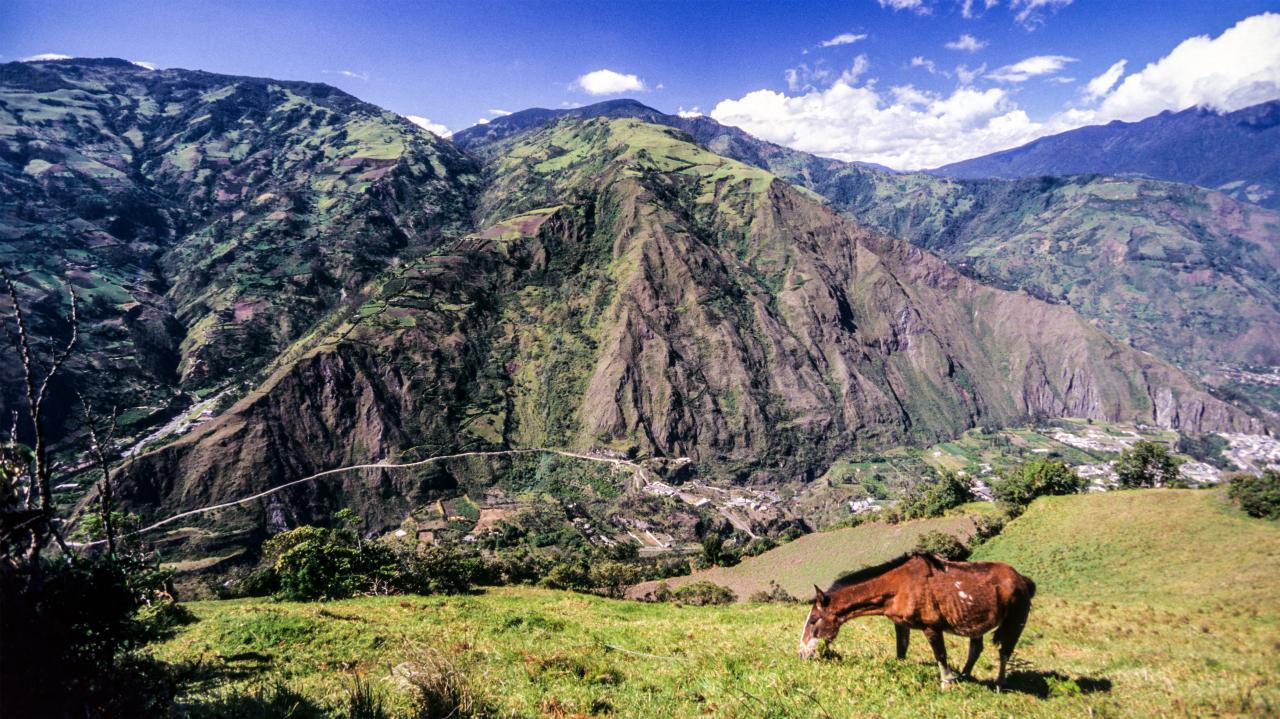}
    \caption{Initial image ($2362\times 4200$)}
    \end{subfigure}~%
    \begin{subfigure}{0.45\linewidth}
    \includegraphics[width=\linewidth]{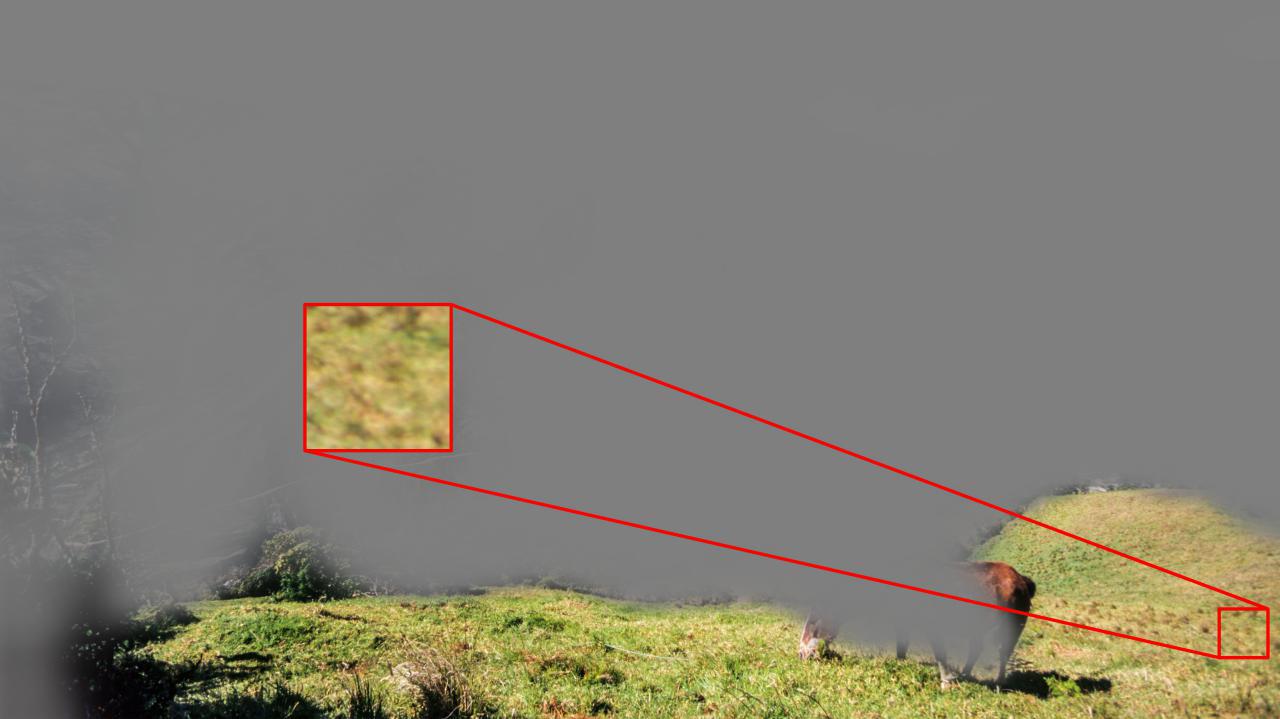}
    \caption{Field texture is detected}
    \end{subfigure}
    \caption{Examples of texture detection by our model for unseen images.}
    \label{fig:raw_test}
    \vspace{-0.7cm}
\end{figure}

In this part, we train our model with $d^g = 2$ for 5 high-resolution images (see Figure~\ref{fig:raw_data}) in a fully unsupervised way. We show that our method learns a descriptive manifold of textures from these images (see Figure~\ref{fig:raw_manifold}) which we can use for texture detection. We demonstrate that we can apply this technique for unseen images (see Figure~\ref{fig:raw_test}). 


\clearpage
\section{Algorithm Description} 
\label{sec:alg_desc}
The main algorithm for the training pipeline is presented in Algorithm~\ref{alg:pagan} and visualised in Figure~\ref{fig:pipeline}. Additionally, it is possible to add $KL$ regularization $KL(q_{\varphi}(\hat{z}^g_i|x_i)\|\mathcal{N}(0, I))$  for embeddings along with $\mathcal{L}_e$ that will increase variance of $q_{\varphi}(\hat{z}^g_i|x_i)$ distribution. This term does not worsen the results added with weight $0.1$ for $d^g=2$, or with $1$ for $d^g>5$ but slightly improves training stability. The improvement is marginal and optional and for this reason we omitted this in the main text. We also did not use this loss term to train models presented in the main text.
\begin{algorithm}[!h]
\begin{algorithmic}
    \State $\theta, \varphi, \psi, \zeta, \tau \gets \text{initialize network parameters}$
    \Repeat
		\State $\bm{x}^{(1)}, \ldots, \bm{x}^{(N)} \sim p^*(\bm{x})$
            \Comment{Draw $N$ textures from the dataset}
        \State $\bar{\bm{x}}^{(1)}, \ldots, \bar{\bm{x}}^{(N)}$
            \Comment{For each $\bm{x}^{(i)}$ draw another example $\bar{\bm{x}}^{(i)}$ of the same texture} \\
		\State $\bm{z}^{(1)}, \ldots, \bm{z}^{(N)} \sim p(\bm{z})$
		    \Comment{Draw $N$ noise samples from prior where $\bm{z}^{(i)} = [z_i^g, z_i^l, z_i^p]$} \\
		\State $q(\hat{z}^g_i|x_i) = E_{\varphi}(\bm{x}^{(i)}), \quad i = 1, \ldots, N$
		\State $\hat{z}^g_i \sim q(\hat{z}^g_i|x_i), \quad i = 1, \ldots, N$
		\State $\hat{\bm{z}}^{(i)} = [\hat{z}^g_i, \hat{z}_i^l, \hat{z}_i^p], \quad i = 1, \ldots, N$
            \Comment{Sample embeddings for each $\bm{x}^{(i)}$ from the encoder distribution} \\
		\State $\bm{x}_{pr}^{(j)}  = G_{\theta}(\bm{z}^{(j)}),
				\quad j = 1, \ldots, N$
			\Comment{Sample textures from prior}
		\State $\bm{x}_{rec}^{(i)} = G_{\theta}(\hat{\bm{z}}^{(i)}),
				\quad j = 1, \ldots, N$
			\Comment{Sample reproductions for each real texture $\bm{x}^{(i)}$} \\
        \State $\mathcal{L}_d^x \gets
            -\frac{1}{N} \sum\limits_{i=1}^N\frac{1}{st}\sum\limits_{k,l}^{s,t} \log D^{kl}_{\psi}(\bm{x}^{(i)})
            -\frac{1}{N} \sum\limits_{j=1}^N\frac{1}{st}\sum\limits_{k,l}^{s,t}\log\left(1 - D_{\psi}^{kl}(\bm{x}_{pr}^{(j)})\right)
            -\frac{1}{N} \sum\limits_{i=1}^N\frac{1}{st}\sum\limits_{k,l}^{s,t}\log\left(1 - D_{\psi}^{kl}(\bm{x}_{rec}^{(i)})\right)$ \\
            \Comment{Compute the loss of the discriminator $D_{\psi}(x)$}
        \State $\mathcal{L}_d^z \gets
            -\frac{1}{N} \sum\limits_{i=1}^N \log D_{\zeta}(z_i^g)
            -\frac{1}{N} \sum\limits_{j=1}^N \log\left(1 - D_{\zeta}(\hat{z}^g_j)\right)$ \\
            \Comment{Compute the loss of the discriminator $D_{\zeta}(z^g)$} \\
        \State $\mathcal{L}_d^{xx} \gets
            -\frac{1}{N} \sum\limits_{i=1}^N\frac{1}{pq}\sum\limits_{k,l}^{p,q} \log D_{\tau}^{kl}(\bm{x}^{(i)}, \bar{\bm{x}}^{(i)})
            -\frac{1}{N} \sum\limits_{j=1}^N\frac{1}{pq}\sum\limits_{k,l}^{p,q}\log\left(1 - D_{\tau}^{kl}(\bm{x}^{(j)}, \bm{x}^{(j)}_{rec})\right)$ \\
            \Comment{Compute the loss of the discriminator $D_{\tau}(x, y)$} \\
        \State $\mathcal{L}_g \gets
            -\frac{1}{N} \sum\limits_{i=1}^N\frac{1}{st}\sum\limits_{k,l}^{s,t} \log D_{\psi}^{kl}(\bm{x}^{(i)}_{pr})
            -\frac{1}{N} \sum\limits_{i=1}^N\frac{1}{st}\sum\limits_{k,l}^{s,t} \log D_{\psi}^{kl}(\bm{x}^{(i)}_{rec})
            -\frac{1}{N} \sum\limits_{j=1}^N\frac{1}{pq}\sum\limits_{k,l}^{p,q} \log D_{\tau}^{kl}(\bm{x}^{(j)}, \bm{x}^{(j)}_{rec})$ \\
            \Comment{Compute the loss of the generator $G_{\theta}(z)$}
        \State $\mathcal{L}_e \gets
            -\frac{1}{N} \sum\limits_{i=1}^N \log D_{\zeta}(\hat{z}^g_i)
            -\frac{1}{N} \sum\limits_{j=1}^N\frac{1}{pq}\sum\limits_{k,l}^{p,q} \log D_{\tau}^{kl}(\bm{x}^{(j)}, \bm{x}^{(j)}_{rec})$ \\
            \Comment{Compute the loss of the encoder $E_{\varphi}(x)$} \\
        \State $\psi \gets \psi - \nabla_{\psi} \mathcal{L}_d^x, \; \zeta \gets \zeta - \nabla_{\zeta} \mathcal{L}_d^z$
            \Comment{Gradient update on discriminator networks}
        \State $\tau \gets \tau - \nabla_{\tau} \mathcal{L}_d^{xx}$
        \State $\theta \gets \theta - \nabla_{\theta} \mathcal{L}_g, \; \varphi \gets \varphi - \nabla_{\varphi} \mathcal{L}_e$
            \Comment{Gradient update on generator-encoder networks}
    \Until{convergence}
\end{algorithmic}
\caption{\label{alg:pagan} The training algorithm of the proposed method.}
\end{algorithm}
\clearpage

\section{Network Architectures \label{app:network-architectures}} 
\subsection{Encoder Network Architecture}
The architecture of the encoder $E_\varphi(x)$ is similar to the discriminator $D_\psi(x)$ architecture. The first difference is that batch norm layers are added. The second one is that convolutional layers are followed by global average pooling to obtain a single embedding for an input texture. See Table~\ref{tab:encoder_arch}.

\begin{table}[!h]
    \centering
    \caption{Architecture description for the encoder $E_\varphi(x)$}
    \label{tab:encoder_arch}
    \ra{1.2}
    \begin{tabular}{@{}ccc@{}}\toprule
        \textbf{Layer} & \textbf{Output size} & \textbf{Parameters}\\
        \midrule
        Input& $3\times160\times160$&\\
        Conv2d & $64\times80\times80$ &kernel=5, stride=2, pad=2\\
        LeakyReLU & $64\times80\times80$ &slope=0.2\\
        \hline
        Conv2d & $128\times40\times40$ &kernel=5, stride=2, pad=2, bias=False\\
        BatchNorm2d & $128\times40\times40$  &eps=1e-05, momentum=1.0, affine=True\\
        LeakyReLU & $128\times40\times40$  &slope=0.2\\
        \hline
        Conv2d & $256\times20\times20$  &kernel=5, stride=2, pad=2, bias=False\\
        BatchNorm2d & $256\times20\times20$ &eps=1e-05, momentum=1.0, affine=True\\
        LeakyReLU & $256\times20\times20$ &slope=0.2\\
        \hline
        Conv2d & $512\times10\times10$ &kernel=5, stride=2, pad=2, bias=False\\
        BatchNorm2d & $512\times10\times10$ &eps=1e-05, momentum=1.0, affine=True\\
        LeakyReLU & $512\times10\times10$ &slope=0.2\\
        \hline
        Conv2d & $2 d^g\times5\times5$ &kernel=5, stride=2, pad=2\\
        AdaptiveAvgPool2d& $2 d^g\times1\times1$ &output\_size=1 \\
        \bottomrule
    \end{tabular}
\end{table}
\FloatBarrier

\subsection{Generator Network Architecture}
The architecture for the generator $G_{\theta}(z)$ is taken from the PSGAN model without any changes. See Table~\ref{tab:generator_arch}.

\subsubsection*{Details on $\widetilde G_{\theta}(z)$} 
As mentioned in Section~\ref{sec:spatial-embeddings-and-texture-detection}, we need to modify $G_{\theta}(z)$ to obtain "reconstructed" picture in Figure~\ref{fig:teaser}. This is done modifying only \textbf{Compute Period Coefs} part in the generator by replacing Linear layer with Conv1x1 with the same weight matrix. Previously, we had shared period coefficients along spatial dimensions and they were dependent only on one global code $z^g$. Once we apply $\widetilde E_{\varphi}(x)$ (replacing global pooling with spatial pooling in $E_{\varphi}(x)$) to an image, we obtain varying "global" codes  $z^g$ along spatial dimension. Conv1x1 allows to efficiently compute periodic coefficients for every spatial position to obtain $z^p$. $z^p_{kij} = \sin(a_k(z^g)_{ij}\cdot i + b_k(z^g)_{ij}\cdot j)$, note, that random offset is manually set to zero. Then global tensor $z^g$ is stacked with $z^p$ and $z^l$ to get $[z^g, z^l, z^p]$ that is passed to the \textbf{Upsampling part} in the generator. As the generator is fully convolutional, we are free in an input spatial size.

\begin{table}[!h]
    \centering
    \caption{Architecture description for the generator $G_\theta(x)$}
    \label{tab:generator_arch}
    \ra{1.2}
    \begin{tabular}{@{}ccc@{}}\toprule
        \textbf{Layer} & \textbf{Output size} & \textbf{Parameters}\\
        \midrule
        \textbf{Upsampling part}& & \\
        \hline
        Input & $(d^g+d^l+d^p)\times5\times5$ &\\
        ConvTranspose2d & $512\times10\times10$ &kernel=5, stride=2, pad=2, output\_pad=1, bias=False\\
        BatchNorm2d & $512\times10\times10$ &eps=1e-05, momentum=1.0, affine=True\\
        LeakyReLU & $512\times10\times10$ &slope=0.2\\
        \hline
        ConvTranspose2d & $256\times20\times20$ &kernel=5, stride=2, pad=2, output\_pad=1, bias=False\\
        BatchNorm2d & $256\times20\times20$ &eps=1e-05, momentum=1.0, affine=True\\
        LeakyReLU & $256\times20\times20$ &slope=0.2\\
        \hline
        ConvTranspose2d & $128\times40\times40$ &kernel=5, stride=2, pad=2, output\_pad=1, bias=False\\
        BatchNorm2d & $128\times40\times40$ &eps=1e-05, momentum=1.0, affine=True\\
        LeakyReLU & $128\times40\times40$ &slope=0.2\\
        \hline
        ConvTranspose2d & $64\times80\times80$ &kernel=5, stride=2, pad=2, output\_pad=1, bias=False\\
        BatchNorm2d & $64\times80\times80$ &eps=1e-05, momentum=1.0, affine=True\\
        LeakyReLU & $64\times80\times80$ &slope=0.2\\
        \hline
        ConvTranspose2d & $3\times160\times160$ &kernel=5, stride=2, pad=2, output\_pad=1, bias=False\\
        Tanh & $3\times160\times160$ &\\
        \hline
        \textbf{Compute Period Coefs}& & \\
        \hline
        Input&$d^g$&\\
        Linear& $40$& \\
        ReLU & $40$ &\\
        Linear& $2d^p$& \\
        \bottomrule
    \end{tabular}
\end{table}
\FloatBarrier

\subsection{Architectures of Discriminator Networks}
\subsubsection{Discriminator $D_{\psi}(x)$}
The architecture for discriminator $D_\psi(x)$ is taken from PSGAN model with added spectral norm in it. Spectral norm improves the stability of training. See Table~\ref{tab:disc_x_arch}.
\begin{table}[!h]
    \centering
    \caption{Architecture description for the texture discriminator $D_\psi(x)$}
    \label{tab:disc_x_arch}
    \ra{1.3}
    \begin{tabular}{@{}ccc@{}}\toprule
        \textbf{Layer} & \textbf{Output size} & \textbf{Parameters}\\
        \midrule
        Input& $3\times160\times160$&\\

        Conv2d & $64\times80\times80$ &kernel=5, stride=2, pad=2 + spectral\_norm\\
        LeakyReLU & $64\times80\times80$ &slope=0.2\\
        \hline
        Conv2d & $128\times40\times40$ &kernel=5, stride=2, pad=2 + spectral\_norm\\
        LeakyReLU & $128\times40\times40$ &slope=0.2\\
        \hline
        Conv2d & $256\times20\times20$ &kernel=5, stride=2, pad=2\\
        LeakyReLU & $256\times20\times20$ &slope=0.2\\
        \hline
        Conv2d & $512\times10\times10$ &kernel=5, stride=2, pad=2 + spectral\_norm\\
        LeakyReLU &  $512\times10\times10$ &slope=0.2\\
        \hline
        Conv2d & $1\times5\times5$ &kernel=5, stride=2, pad=2 + spectral\_norm \\
        \bottomrule
    \end{tabular}
\end{table}
\FloatBarrier

\subsubsection{Discriminator $D_{\tau}(x, y)$}
The proposed architecture for the discriminator on pairs $D_{\tau}(x, y)$ the \textbf{Convolutional part} is same as for $D_\psi(x)$ except the last number of channels. The output for two images constructs a matrix of size $c\times h_1\cdot w_1\times h_2\cdot w_2$ and \textbf{Conv 1x1 part} is applied to this matrix to obtain spatial predictions for each image. This architecture is symmetric with respect to input order and can work with different sized images pairs (we did not require this feature in our algorithm). See Table~\ref{tab:disc_xx_arch}.

\begin{table}[!h]
    \centering
    \caption{Architecture description for the pair discriminator $D_\tau(x, y)$}
    \label{tab:disc_xx_arch}
    \ra{1.3}
    \begin{tabular}{@{}ccc@{}}\toprule
        \textbf{Layer} & \textbf{Output size} & \textbf{Parameters}\\
        \midrule
        \textbf{Convolutional part}& &\\
        \hline
        Input& $3\times160\times160$&\\
        Conv2d & $64\times80\times80$ &kernel=5, stride=2, pad=2 + spectral\_norm\\
        LeakyReLU & $64\times80\times80$ &slope=0.2\\
        \hline
        Conv2d & $128\times40\times40$ &kernel=5, stride=2, pad=2 + spectral\_norm\\
        LeakyReLU & $128\times40\times40$ &slope=0.2\\
        \hline
        Conv2d & $256\times20\times20$ &kernel=5, stride=2, pad=2\\
        LeakyReLU & $256\times20\times20$ &slope=0.2\\
        \hline
        Conv2d & $512\times10\times10$ &kernel=5, stride=2, pad=2 + spectral\_norm\\
        LeakyReLU & $512\times10\times10$ &slope=0.2\\
        \hline
        Conv2d & $96\times5\times5$ &kernel=5, stride=2, pad=2 + spectral\_norm\\
        \hline
        \textbf{Conv 1x1 part}& &\\
        \hline
        Input& $96\times25\times25$& \\
        Conv2d & $48\times25\times25$ &kernel=1, stride=1 + spectral\_norm\\
        LeakyReLU & $48\times25\times25$ &slope=0.2\\
        Conv2d & $1\times25\times25$ &kernel=1, stride=1 + spectral\_norm\\
        \bottomrule
    \end{tabular}
\end{table}
\FloatBarrier

\subsubsection{Discriminator $D_{\zeta}(z)$}
Following recent works \cite{makhzani2015adversarial} motivated to use adversarial trainign scheme for latent representations. The other benefit from using an additional discriminator is to make loss terms to be at the same scale. See Table~\ref{tab:disc_z_arch}.

\begin{table}[!h]
    \centering
    \caption{Architecture description for the latent discriminator $D_\zeta(z)$}
    \label{tab:disc_z_arch}
    \ra{1.3}
    \begin{tabular}{@{}ccc@{}}\toprule
        \textbf{Layer} & \textbf{Output size} & \textbf{Parameters}\\
        \midrule
        Input&$d^g$&\\
        Linear& $512$& \\
        LeakyReLU & $512$ &slope=0.2\\
        Linear& $256$& \\
        LeakyReLU & $256$ &slope=0.2\\
        Linear& $1$& \\
        \bottomrule
    \end{tabular}
\end{table}
\FloatBarrier

\newpage

\section{Hyperparameters} 
We used the set of hyperparameters to train the model on 116 textures from scaly provided in Table~\ref{tab:gen_hyper-2d} and Table~\ref{tab:network_hyper_2d}. 

\begin{table}[!h]
    \centering
    \caption{General hyperparameters of the model}
    \label{tab:gen_hyper-2d}
    \ra{1.2}
    \begin{tabular}{@{}>{\raggedright}p{5cm}c@{}}\toprule
        \textbf{Hyperparameter} & \textbf{Value} \\
        \midrule
        crop size from image &  (160, 160) \\
        batch size & 64 \\
        spectral normalization for discriminators & True \\
        number of steps for discriminator per 1 step of generator &  1\\
        iterations & 100000\\
        latent prior &  $\mathcal{N}(0, I)$\\
        $d^g$ & 2 \\
        $d^l$ &  20\\
        $d^p$ &  4\\
        \bottomrule
    \end{tabular}
\end{table}

\begin{table}[!h]
    \centering
    \caption{Network specific hyperparameters for $G_{\theta}(z)$, $E_{\varphi}(x)$, $D_{\psi}(x)$, $D_{\tau}(x, y)$, $D_{\zeta}(z)$}
    \label{tab:network_hyper_2d}
    \ra{1.2}
    \begin{tabular}{@{}>{\raggedright}p{5cm}c@{}}\toprule
        \textbf{Hyperparameter} & \textbf{Value} \\
        \midrule
        initialization for weights &  $\mathcal{N}(0, 0.02)$ \\
        optimizer & adam\\
        adam betas &  0.5, 0.999\\
        learning rate &  0.0002\\
        weight decay & 0.0001\\
        \bottomrule
    \end{tabular}
\end{table}


\newpage

\section{3D Porous Media Synthesis} 
\label{sec:appendix_3d}

In this section, we describe network architectures, hyperparameters and experiments for the 3D porous media generation.

\subsection{Network Architectures}
Architectures for 3D porous media synthesis have almost the same structure as for 2D textures. The main differences are the following:
\begin{enumerate}
    \item instead of Conv2D (TransposedConv2D) layers we used Conv3D (TransposedConv3D) layers;
    \item we do not use periodical latent component since there is no need in periodicity in porous structures.
\end{enumerate}

In order to honestly compare our model with the baseline, we used the same generator and discriminator networks in both our model and the baseline.

\subsubsection{3D Encoder Network Architecture}
The architecture of the 3D encoder $E_{\varphi}(x)$ is presented in Table~\ref{tab:3d_encoder_arch}.

\begin{table}[!h]
    \centering
    \caption{Architecture description for the 3D encoder $E_\varphi(x)$}
    \label{tab:3d_encoder_arch}
    \ra{1.2}
    \begin{tabular}{@{}ccc@{}}\toprule
        \textbf{Layer} & \textbf{Output size} & \textbf{Parameters}\\
        \midrule
        Input& $1\times160\times160\times160$&\\
        Conv3d & $16\times80\times80\times80$ &kernel=4, stride=1, pad=0\\
        LeakyReLU & $16\times80\times80\times80$ &slope=0.01\\
        \hline
        Conv3d & $32\times40\times40\times40$ &kernel=4, stride=2, pad=1, bias=False\\
        BatchNorm3d & $32\times40\times40\times40$  &eps=1e-05, momentum=1.0, affine=True\\
        LeakyReLU & $32\times40\times40\times40$  &slope=0.01\\
        \hline
        Conv3d & $64\times20\times20\times20$  &kernel=4, stride=2, pad=1, bias=False\\
        BatchNorm3d & $64\times20\times20\times20$ &eps=1e-05, momentum=1.0, affine=True\\
        LeakyReLU & $64\times20\times20\times20$ &slope=0.01\\
        \hline
        Conv3d & $72\times10\times10\times10$ &kernel=4, stride=2, pad=1, bias=False\\
        BatchNorm3d & $72\times10\times10\times10$ &eps=1e-05, momentum=1.0, affine=True\\
        LeakyReLU & $72\times10\times10\times10$ &slope=0.01\\
        \hline
        Conv3d & $128\times5\times5\times5$ &kernel=4, stride=2, pad=1, bias=False\\
        BatchNorm3d & $128\times5\times5\times5$ &eps=1e-05, momentum=1.0, affine=True\\
        LeakyReLU & $128\times5\times5\times5$ &slope=0.01\\
        \hline
        Conv3d & $16\times7\times7\times7$ &kernel=1, stride=1, pad=0\\
        AdaptiveAvgPool3d& $16\times1\times1\times1$ &output\_size=1 \\
        \bottomrule
    \end{tabular}
\end{table}

\subsubsection{3D Generator Network Architecture}
The architecture of the 3D generator $G_{\theta}(z)$ is presented in Table~\ref{tab:3d_generator_arch}. The same generator architecture was used in the baseline.

\begin{table}[!h]
    \centering
    \caption{Architecture description for the 3D generator $G_\theta(x)$}
    \label{tab:3d_generator_arch}
    \ra{1.2}
    \begin{tabular}{@{}ccc@{}}\toprule
        \textbf{Layer} & \textbf{Output size} & \textbf{Parameters}\\
        \midrule
        \textbf{Upsampling part}& & \\
        \hline
        Input & $(d^g+d^l)\times7\times7\times7$ &\\
        ConvTranspose3d & $512\times10\times10\times10$ &kernel=4, stride=1, pad=0, output\_pad=1, bias=False\\
        BatchNorm3d & $512\times10\times10\times10$ &eps=1e-05, momentum=1.0, affine=True\\
        LeakyReLU & $512\times10\times10\times10$ &slope=0.01\\
        \hline
        ConvTranspose3d & $256\times20\times20\times20$ &kernel=4, stride=2, pad=1, output\_pad=1, bias=False\\
        BatchNorm3d & $256\times20\times20\times20$ &eps=1e-05, momentum=1.0, affine=True\\
        LeakyReLU & $256\times20\times20\times20$ &slope=0.01\\
        \hline
        ConvTranspose3d & $128\times40\times40\times40$ &kernel=4, stride=2, pad=1, output\_pad=1, bias=False\\
        BatchNorm3d & $128\times40\times40\times40$ &eps=1e-05, momentum=1.0, affine=True\\
        LeakyReLU & $128\times40\times40\times40$ &slope=0.01\\
        \hline
        ConvTranspose3d & $64\times80\times80\times80$ &kernel=4, stride=2, pad=1, output\_pad=1, bias=False\\
        BatchNorm3d & $64\times80\times80\times80$ &eps=1e-05, momentum=1.0, affine=True\\
        LeakyReLU & $64\times80\times80\times80$ &slope=0.01\\
        \hline
        ConvTranspose3d & $1\times160\times160\times160$ &kernel=4, stride=2, pad=1, output\_pad=1, bias=False\\
        Tanh & $1\times160\times160\times160$ &\\
        \bottomrule
    \end{tabular}
\end{table}
\FloatBarrier

\subsubsection{Architectures of the 3D Discriminator Network $D_{\psi}(x)$}
The architecture of the 3D texture discriminator $D_{\psi}(x)$ is presented in Table~\ref{tab:3d_disc_x_arch}.  The same discriminator architecture was used in the baseline.

\begin{table}[!h]
    \centering
    \caption{Architecture description for the 3D texture discriminator $D_\psi(x)$}
    \label{tab:3d_disc_x_arch}
    \ra{1.3}
    \begin{tabular}{@{}ccc@{}}\toprule
        \textbf{Layer} & \textbf{Output size} & \textbf{Parameters}\\
        \midrule
        Input& $1\times160\times160\times160$&\\

        Conv3d & $64\times80\times80\times80$ &kernel=4, stride=2, pad=1 + spectral\_norm\\
        LeakyReLU & $64\times80\times80\times80$ &slope=0.01\\
        \hline
        Conv3d & $128\times40\times40\times40$ &kernel=4, stride=2, pad=1 + spectral\_norm\\
        LeakyReLU & $128\times40\times40\times40$ &slope=0.01\\
        \hline
        Conv3d & $256\times20\times20\times20$ &kernel=4, stride=2, pad=1\\
        LeakyReLU & $256\times20\times20\times20$ &slope=0.01\\
        \hline
        Conv3d & $512\times10\times10\times10$ &kernel=4, stride=2, pad=1 + spectral\_norm\\
        LeakyReLU &  $512\times10\times10\times10$ &slope=0.01\\
        \hline
        Conv3d & $1\times7\times7\times7$ &kernel=4, stride=1, pad=0 + spectral\_norm \\
        \bottomrule
    \end{tabular}
\end{table}
\FloatBarrier

\subsubsection{Architectures of the 3D Discriminator Network $D_{\tau}(x, y)$}
The architecture of the 3D pair discriminator $D_{\tau}(x, y)$ is presented in Table~\ref{tab:3d_disc_xx_arch}.

\begin{table}[!h]
    \centering
    \caption{Architecture description for the 3D pair discriminator $D_\tau(x, y)$}
    \label{tab:3d_disc_xx_arch}
    \ra{1.3}
    \begin{tabular}{@{}ccc@{}}\toprule
        \textbf{Layer} & \textbf{Output size} & \textbf{Parameters}\\
        \midrule
        \textbf{Convolutional part}& &\\
        \hline
        Input& $1\times160\times160\times160$&\\

        Conv3d & $16\times80\times80\times80$ &kernel=4, stride=2, pad=1 + spectral\_norm\\
        LeakyReLU & $16\times80\times80\times80$ &slope=0.01\\
        \hline
        Conv3d & $32\times40\times40\times40$ &kernel=4, stride=2, pad=1 + spectral\_norm\\
        LeakyReLU & $32\times40\times40\times40$ &slope=0.01\\
        \hline
        Conv3d & $64\times20\times20\times20$ &kernel=4, stride=2, pad=1\\
        LeakyReLU & $64\times20\times20\times20$ &slope=0.01\\
        \hline
        Conv3d & $73\times10\times10\times10$ &kernel=4, stride=2, pad=1 + spectral\_norm\\
        LeakyReLU &  $73\times10\times10\times10$ &slope=0.01\\
        \hline
        Conv3d & $128 \times7\times7\times7$ &kernel=4, stride=1, pad=0 + spectral\_norm \\
        \hline
        \textbf{Conv 1x1 part}& &\\
        \hline
        Input& $128\times49\times49\times49$& \\
        Conv3d & $128\times49\times49\times49$ &kernel=1, stride=1 + spectral\_norm\\
        LeakyReLU & $128\times49\times49\times49$ &slope=0.01\\
        Conv3d &  $64\times49\times49\times49$ &kernel=1, stride=1 + spectral\_norm\\
        LeakyReLU & $64\times49\times49\times49$ &slope=0.01\\
        Conv3d & $1\times49\times49\times49$ &kernel=1, stride=1 + spectral\_norm\\
        \bottomrule
    \end{tabular}
\end{table}
\FloatBarrier

\subsubsection{Architectures of the 3D Discriminator Network $D_{\zeta}(z)$}
The architecture of the 3D latent discriminator $D_\zeta(z)$ is presented in Table~\ref{tab:3d_disc_z_arch}.

\begin{table}[!h]
    \centering
    \caption{Architecture description for the 3D latent discriminator $D_\zeta(z)$}
    \label{tab:3d_disc_z_arch}
    \ra{1.3}
    \begin{tabular}{@{}ccc@{}}\toprule
        \textbf{Layer} & \textbf{Output size} & \textbf{Parameters}\\
        \midrule
        Input&$d^g$&\\
        Linear& $16$& \\
        LeakyReLU & $512$ &slope=0.01\\
        Linear& $256$& \\
        LeakyReLU & $256$ &slope=0.01\\
        Linear& $1$& \\
        \bottomrule
    \end{tabular}
\end{table}
\FloatBarrier

\newpage

\newpage

\subsection{Hyperparameters of 3D model and experimental details}

We used the set of hyperparameters to train the model on 5 types of porous media provided in Table~\ref{tab:gen_hyper-3d} and Table~\ref{tab:network_hyper_3d}. For the baseline, we used the same parameters.

\begin{table}[!h]
    \centering
    \caption{General hyperparameters used for training on 5 types of porous media}
    \label{tab:gen_hyper-3d}
    \ra{1.2}
    \begin{tabular}{@{}>{\raggedright}p{5cm}c@{}}\toprule
        \textbf{Hyperparameter} & \textbf{Value} \\
        \midrule
        crop size from porous media &  (160, 160, 160) \\
        batch size & 8 \\
        spectral normalization for discriminators & True \\
        number of steps for discriminator per 1 step of generator &  1\\
        iterations & 25000\\
        latent prior &  $\mathcal{N}(0, I)$\\
        $d^g$ & 16 \\
        $d^l$ &  16\\
        $d^p$ &  0\\
        \bottomrule
    \end{tabular}
\end{table}

\begin{table}[!h]
    \centering
    \caption{Network specific hyperparameters for 3D $G_{\theta}(z)$, $E_{\varphi}(x)$, $D_{\psi}(x)$, $D_{\tau}(x, y)$, $D_{\zeta}(z)$}
    \label{tab:network_hyper_3d}
    \ra{1.2}
    \begin{tabular}{@{}>{\raggedright}p{5cm}c@{}}\toprule
        \textbf{Hyperparameter} & \textbf{Value} \\
        \midrule
        initialization for weights &  $\mathcal{N}(0, 0.02)$ \\
        optimizer & adam\\
        adam betas &  0.5, 0.999\\
        learning rate &  0.0001\\
        weight decay & 0.0001\\
        \bottomrule
    \end{tabular}
\end{table}


During training of both our and the baseline model we used spatial latent size $7$. In other words, our latent tensor had the dimension $[\text{batch\_size}, d^g+d^l, 7, 7, 7].$

\newpage

\subsection{3D samples}
\begin{figure}[!h]
    \centering
    \begin{subfigure}[b]{0.15\textwidth}
        \includegraphics[width=\textwidth]{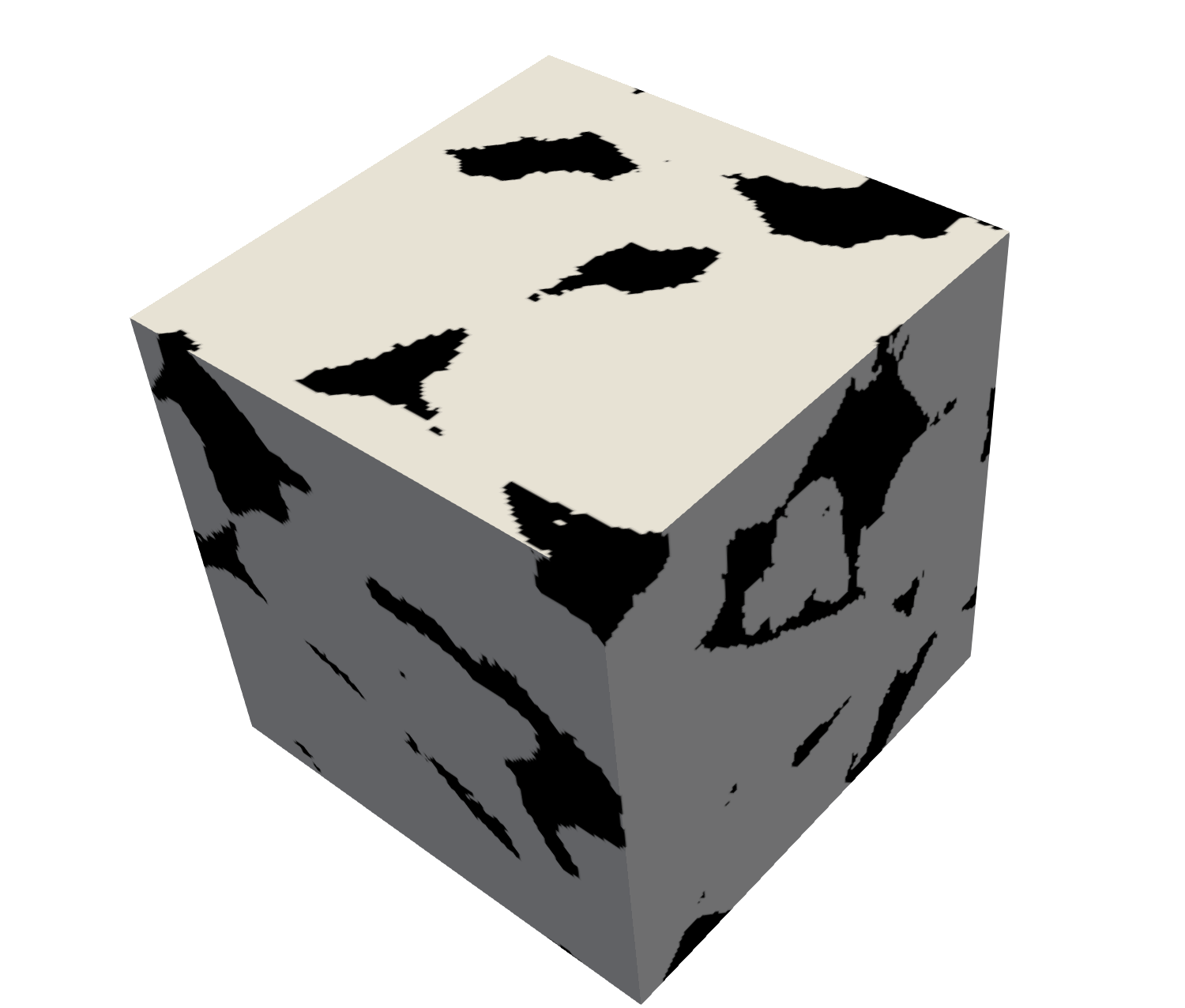}
        \caption{Real $150^3$}
        \label{fig:tiger}
    \end{subfigure}
    ~
    \begin{subfigure}[b]{0.15\textwidth}
        \includegraphics[width=\textwidth]{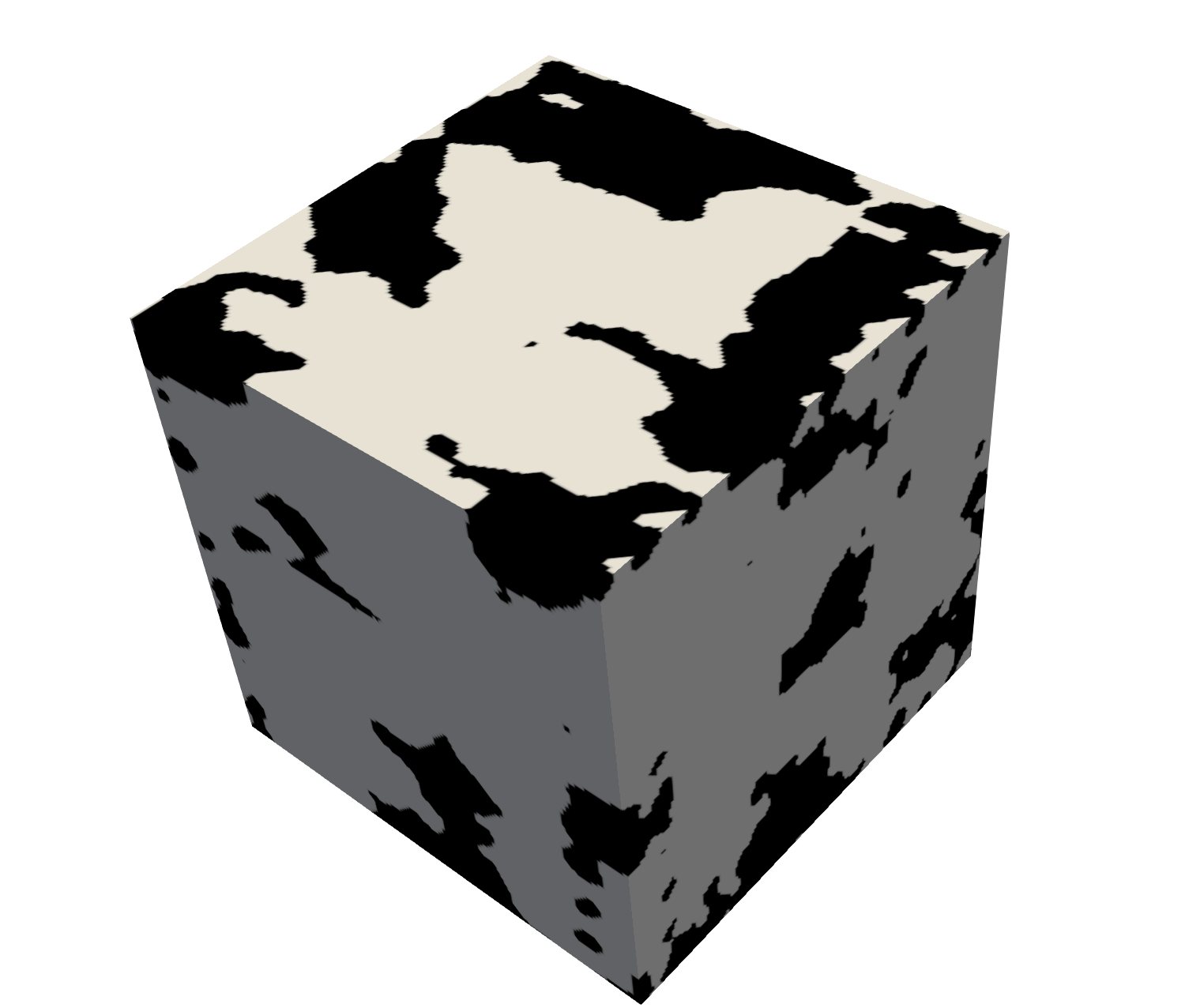}
        \caption{Our $150^3$}
        \label{fig:gull}
    \end{subfigure}
    ~
    \begin{subfigure}[b]{0.15\textwidth}
        \includegraphics[width=\textwidth]{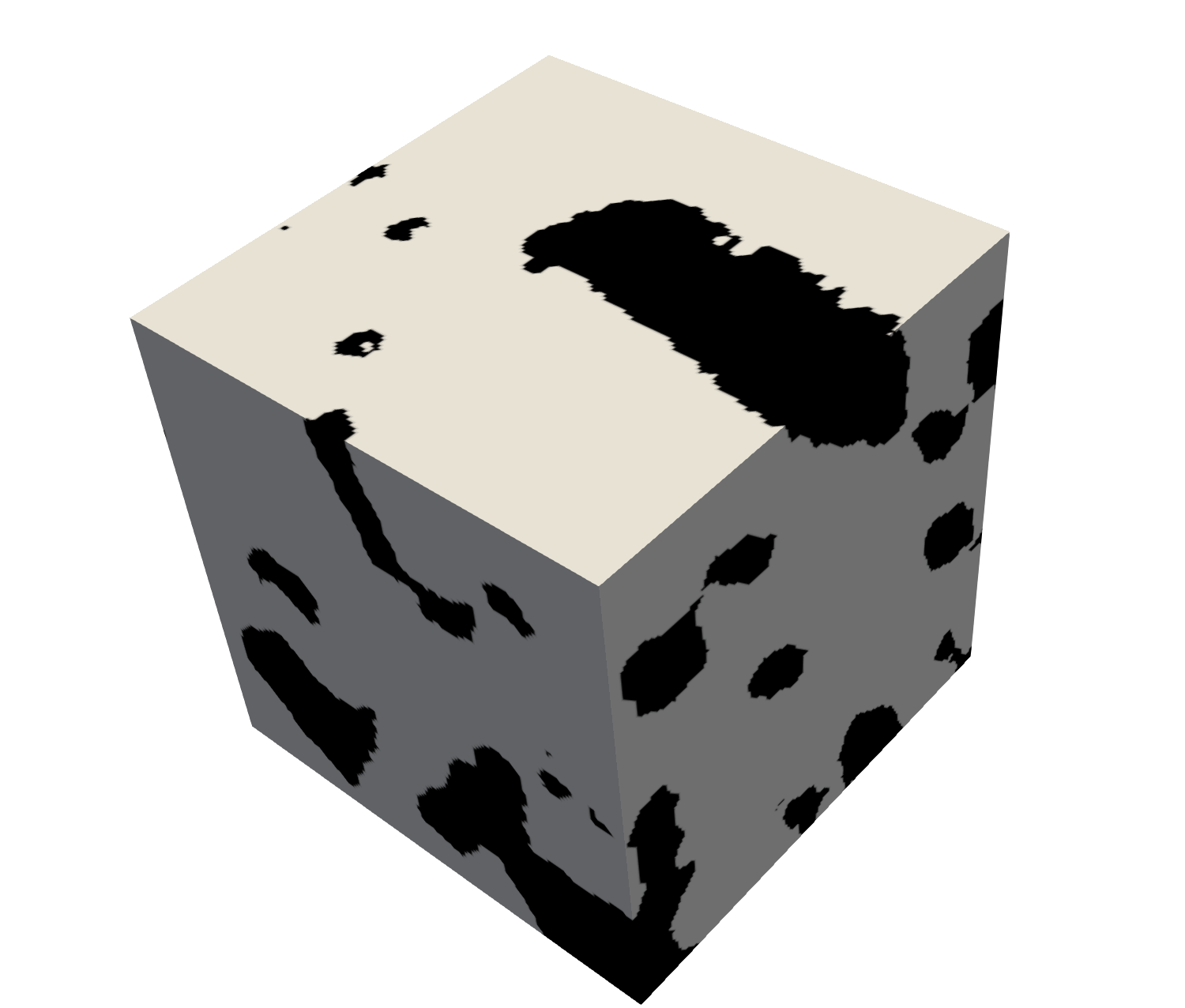}
        \caption{Baseline $150^3$}
        \label{fig:mouse}
    \end{subfigure}
    ~ 
    \begin{subfigure}[b]{0.428\textwidth}
        \includegraphics[width=\textwidth]{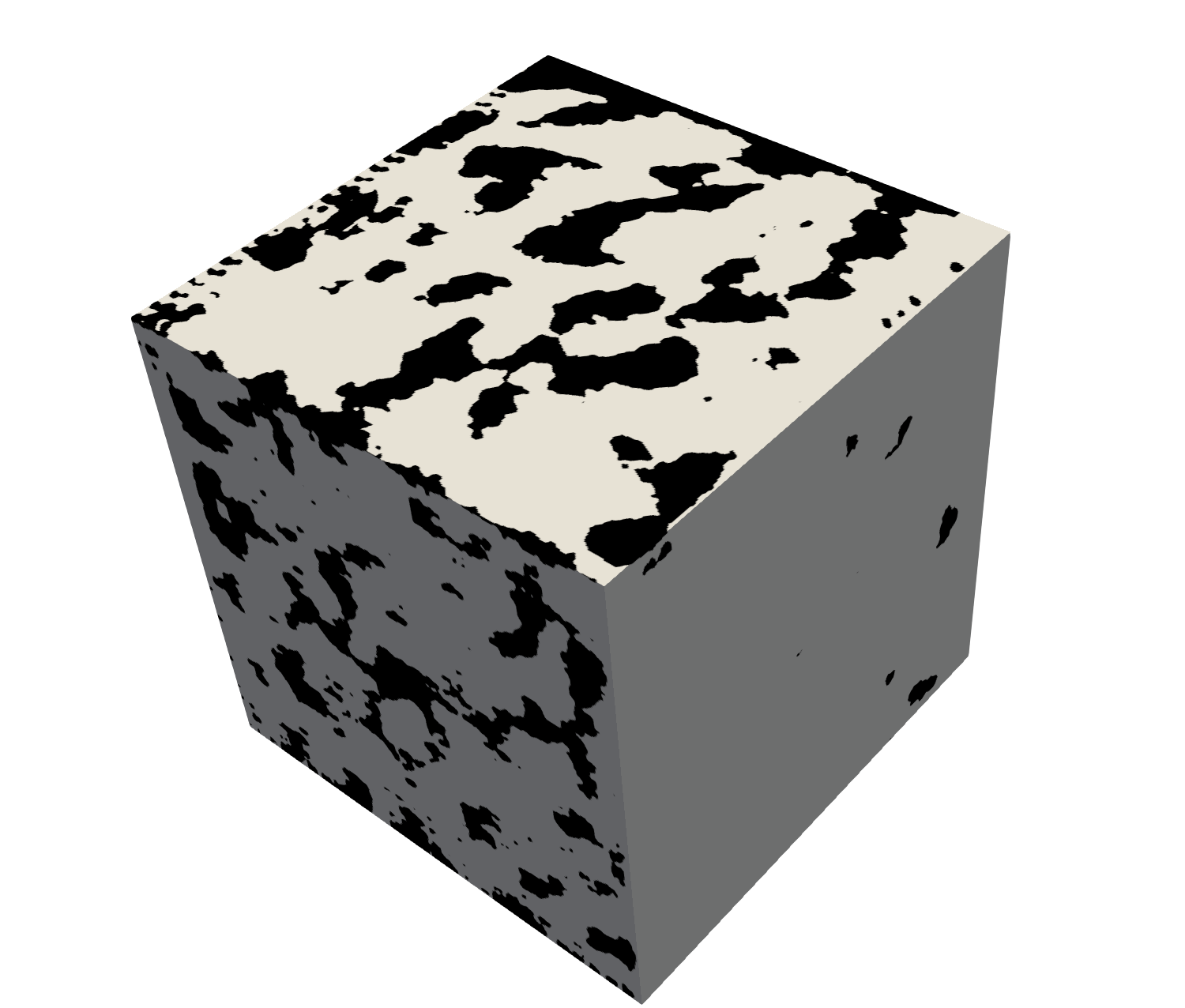}
        \caption{Our $428^3$}
        \label{fig:mouse}
    \end{subfigure}
    \caption{Bentheimer sample. Real (size~$150^3$), sampled with our model (size~$150^3$), samples with the baseline model (size~$150^3$), sampled with our model (size~$428^3$)}
    \label{fig:3d_bentheimer}
\end{figure}

\begin{figure}[!h]
    \centering
    \begin{subfigure}[b]{0.15\textwidth}
        \includegraphics[width=\textwidth]{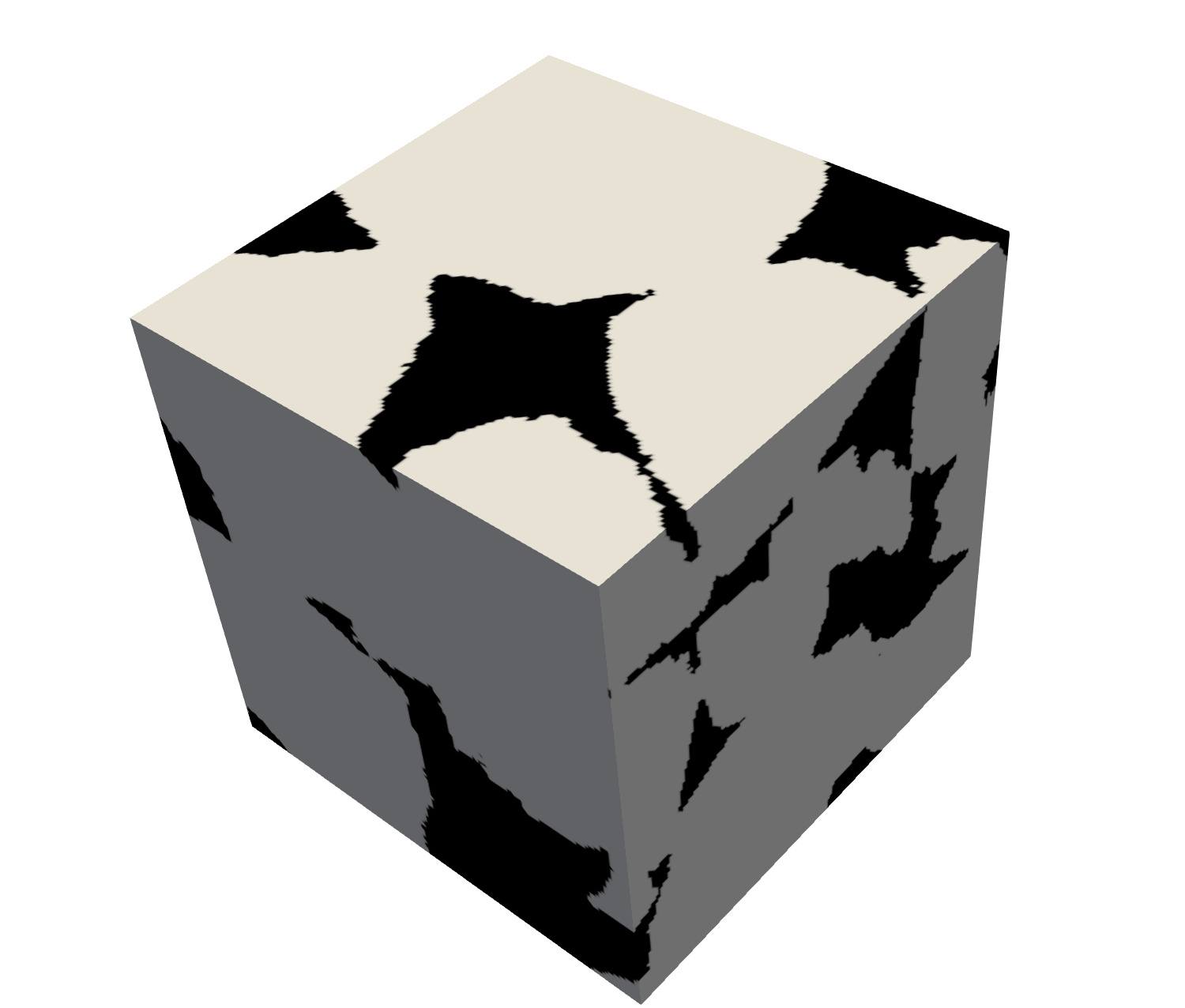}
        \caption{Real $150^3$}
        \label{fig:tiger}
    \end{subfigure}
    ~ 
    \begin{subfigure}[b]{0.15\textwidth}
        \includegraphics[width=\textwidth]{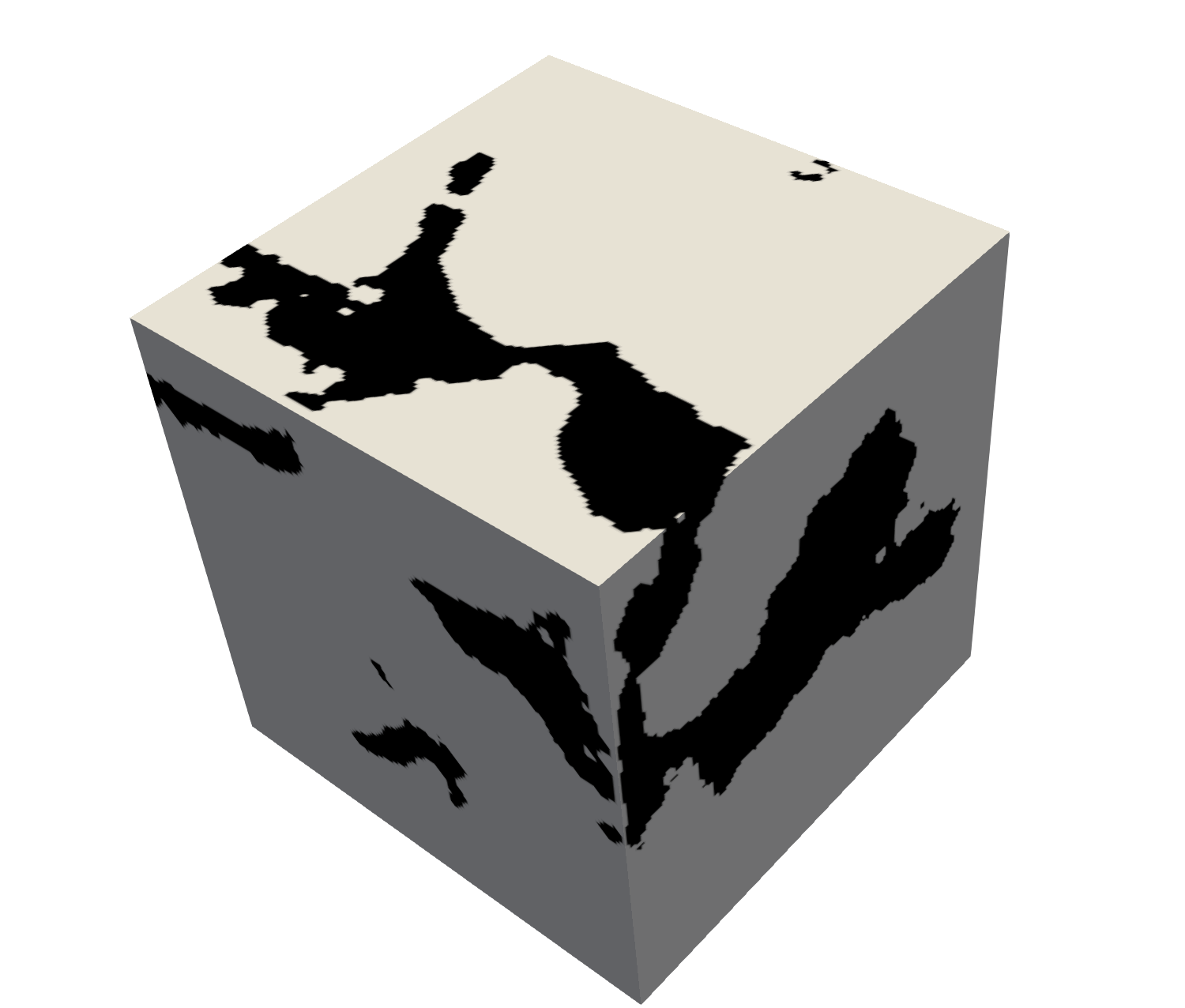}
        \caption{Our $150^3$}
        \label{fig:gull}
    \end{subfigure}
    ~ 
    \begin{subfigure}[b]{0.15\textwidth}
        \includegraphics[width=\textwidth]{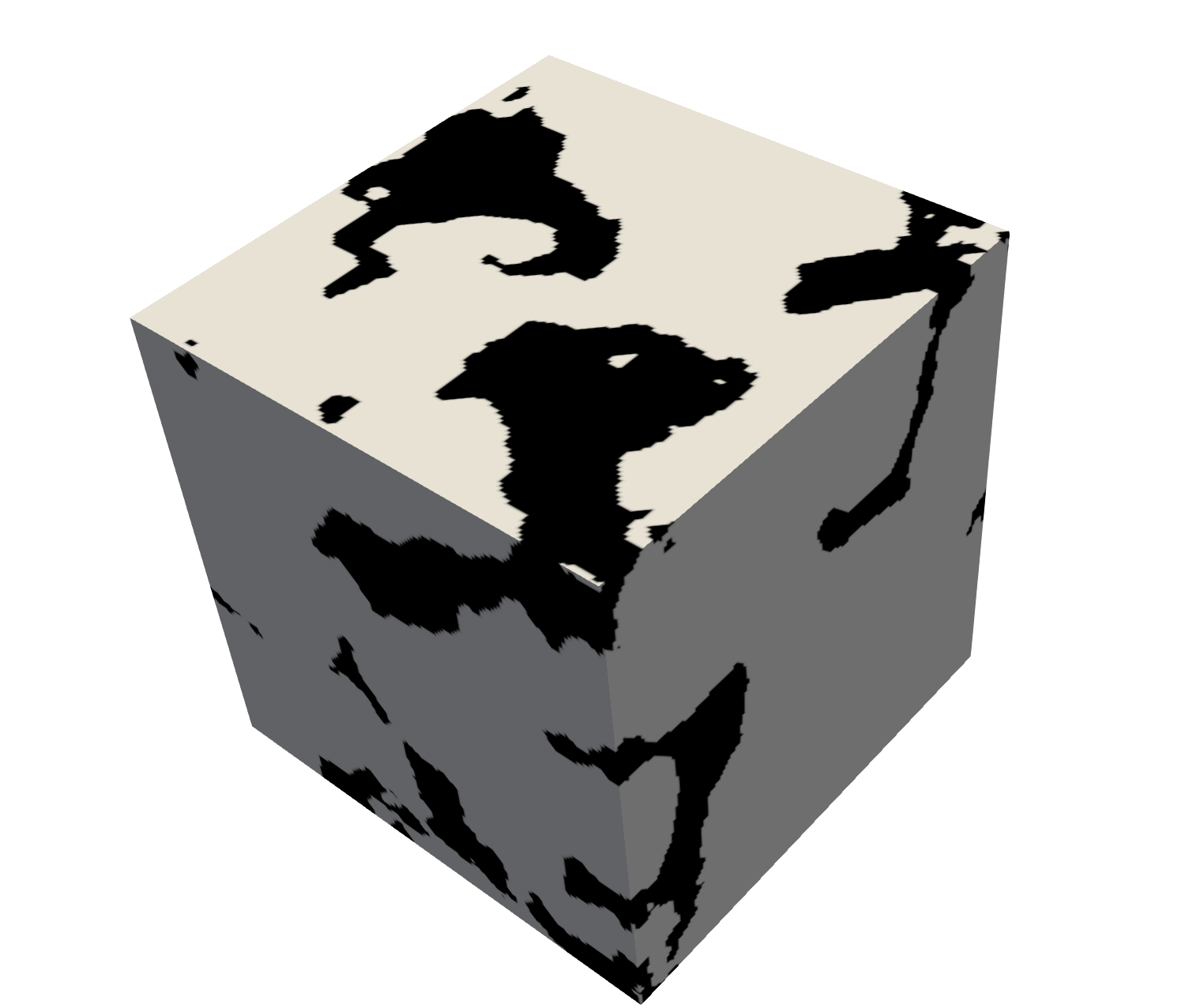}
        \caption{Baseline $150^3$}
        \label{fig:mouse}
    \end{subfigure}
     ~ 
    \begin{subfigure}[b]{0.428\textwidth}
        \includegraphics[width=\textwidth]{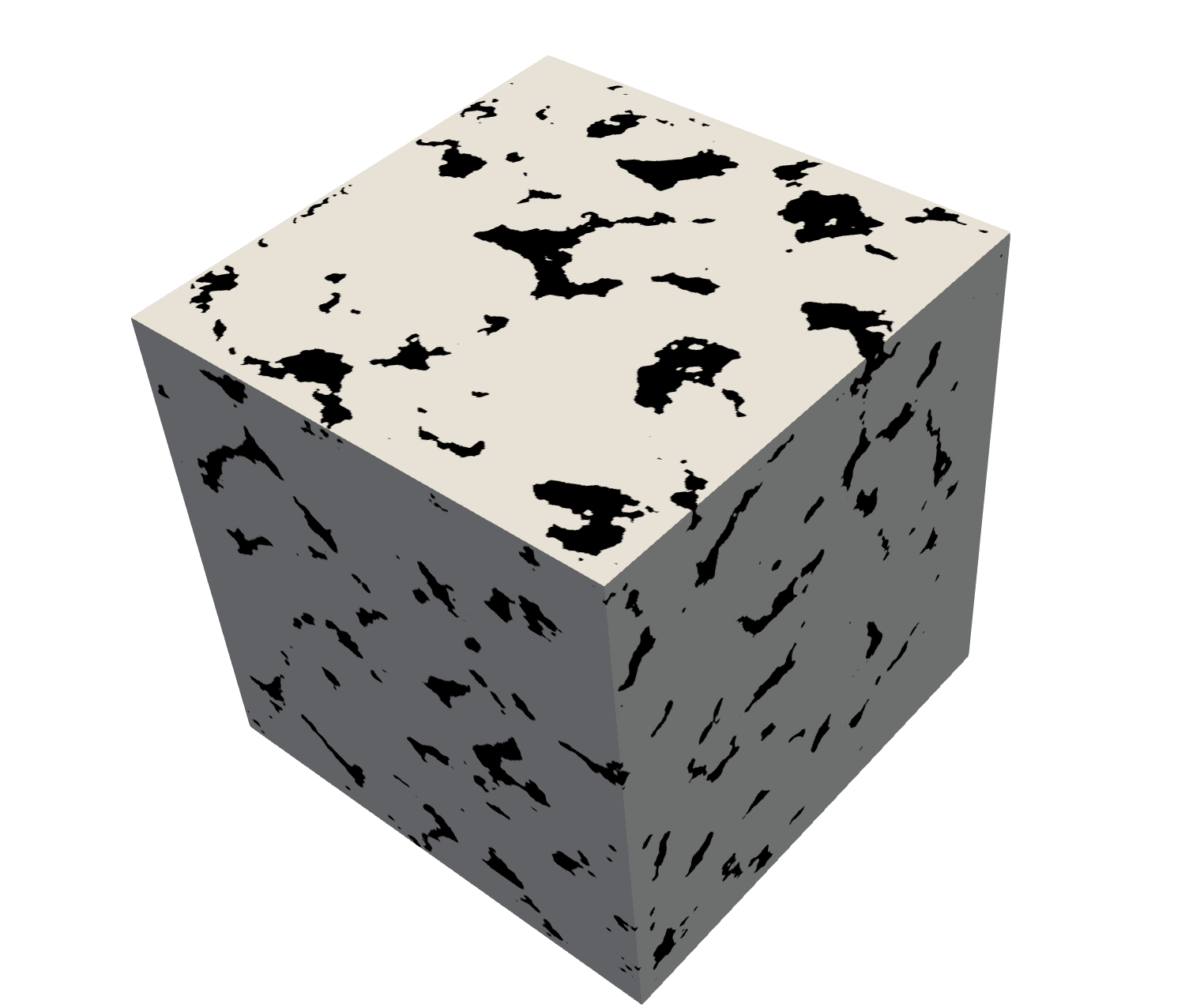}
        \caption{Our $428^3$}
        \label{fig:mouse}
    \end{subfigure}
    \caption{Doddington sample. Real (size~$150^3$), sampled with our model (size~$150^3$), samples with the baseline model (size~$150^3$), sampled with our model (size~$428^3$)}\label{fig:3d_doddington}
\end{figure}

\begin{figure}[!h]
    \centering
    \begin{subfigure}[b]{0.15\textwidth}
        \includegraphics[width=\textwidth]{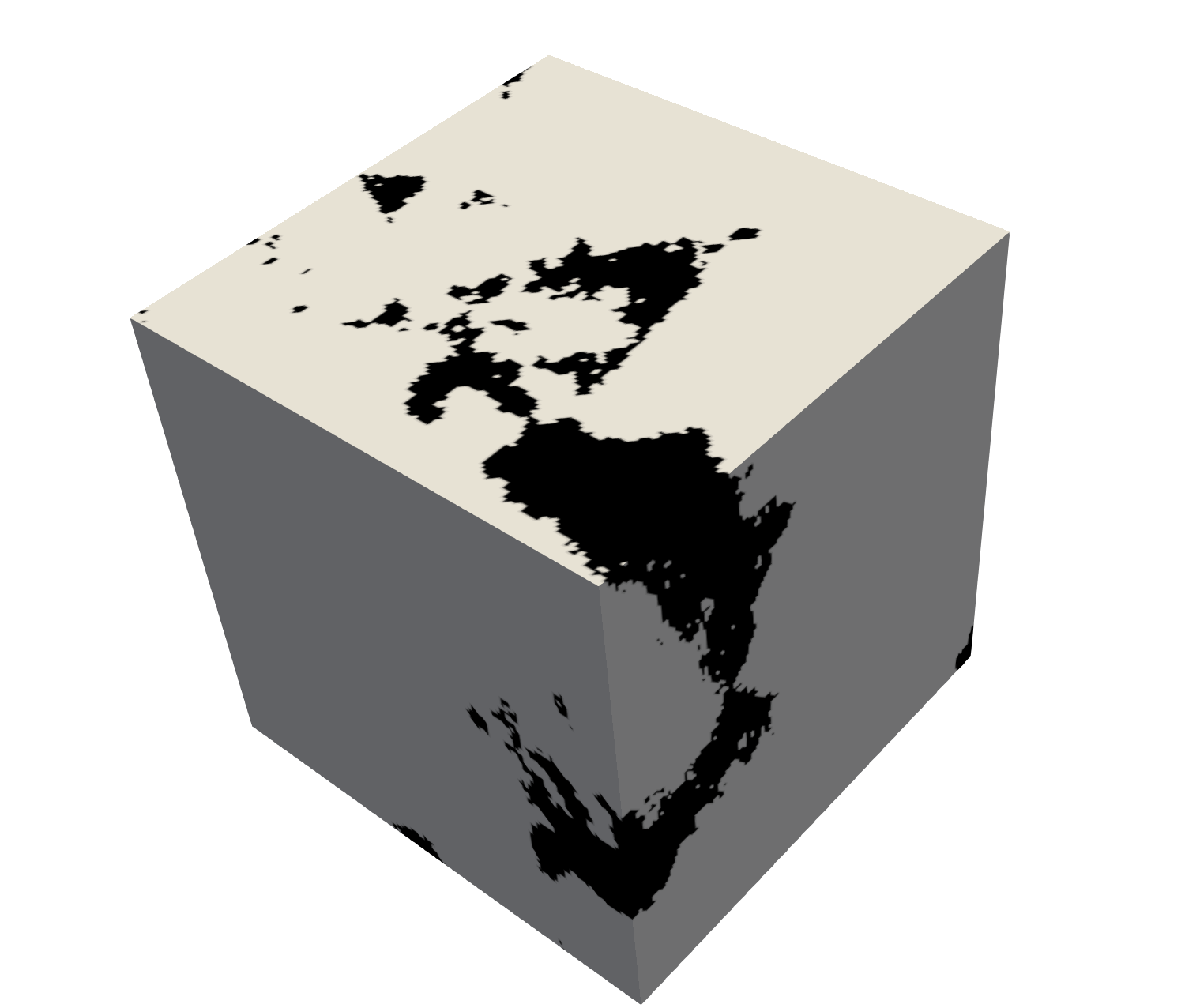}
        \caption{Real $150^3$}
        \label{fig:tiger}
    \end{subfigure}
    ~ 
    \begin{subfigure}[b]{0.15\textwidth}
        \includegraphics[width=\textwidth]{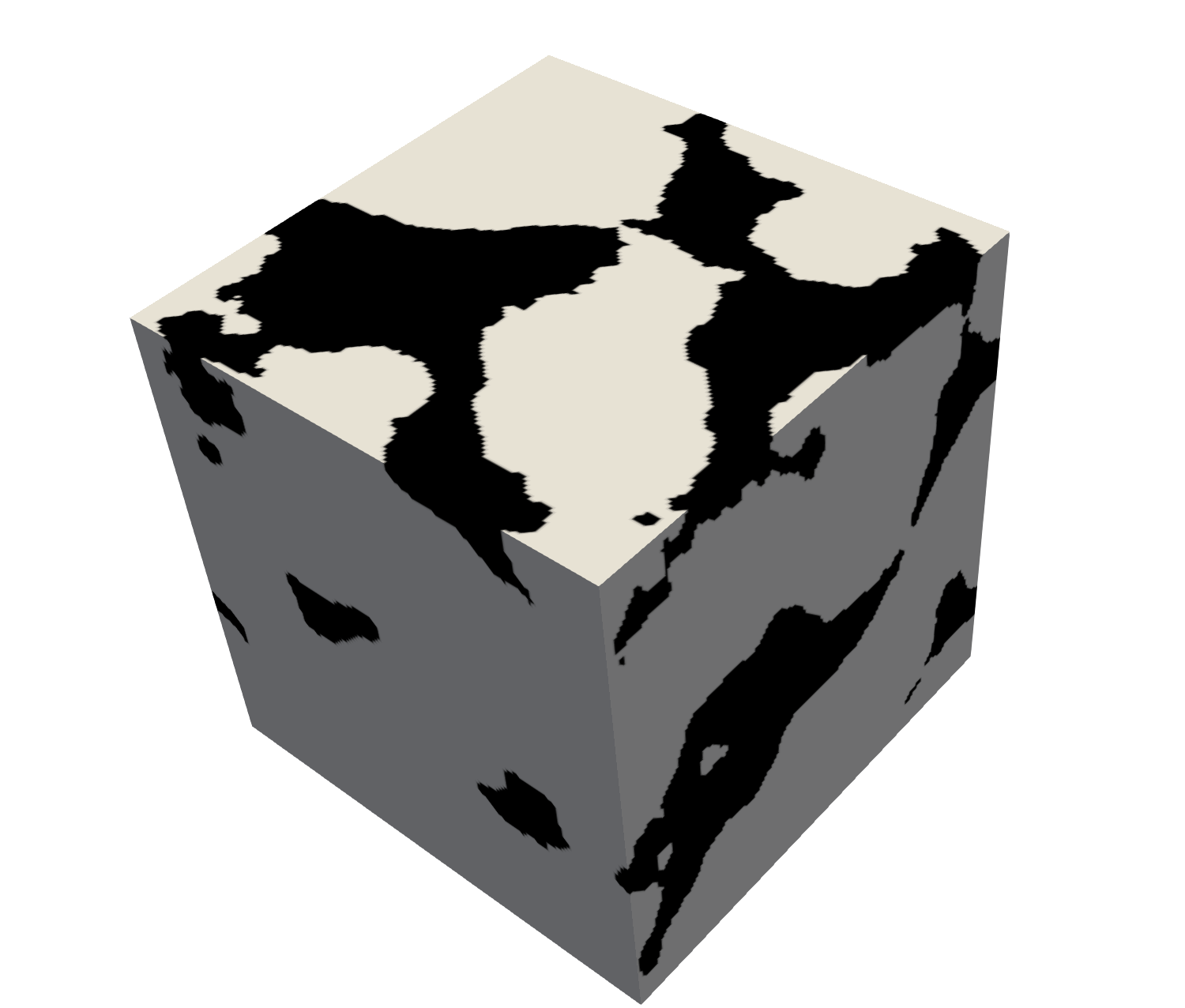}
        \caption{Our $150^3$}
        \label{fig:gull}
    \end{subfigure}
    ~ 
    \begin{subfigure}[b]{0.15\textwidth}
        \includegraphics[width=\textwidth]{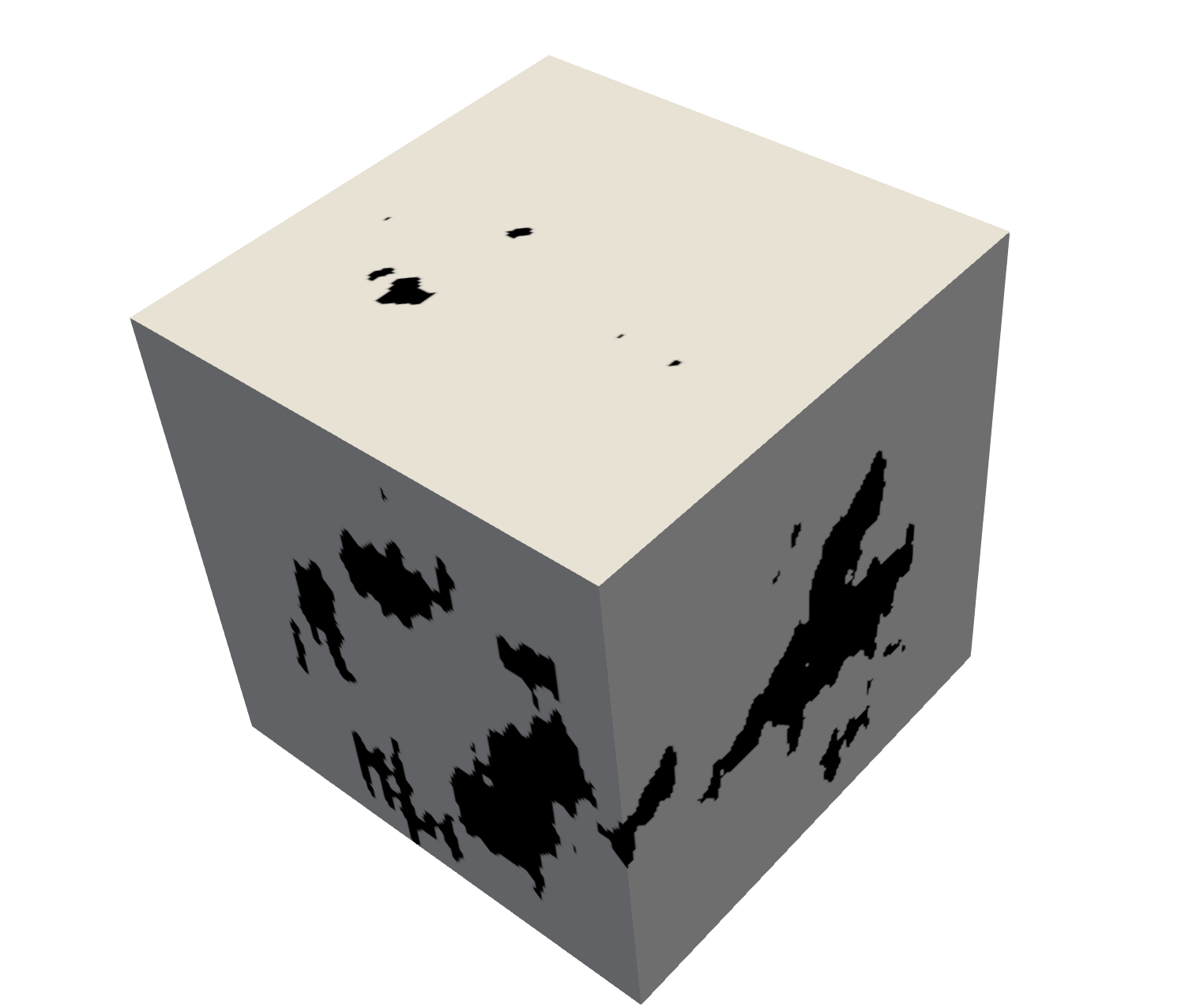}
        \caption{Baseline $150^3$}
        \label{fig:mouse}
    \end{subfigure}
     ~ 
    \begin{subfigure}[b]{0.428\textwidth}
        \includegraphics[width=\textwidth]{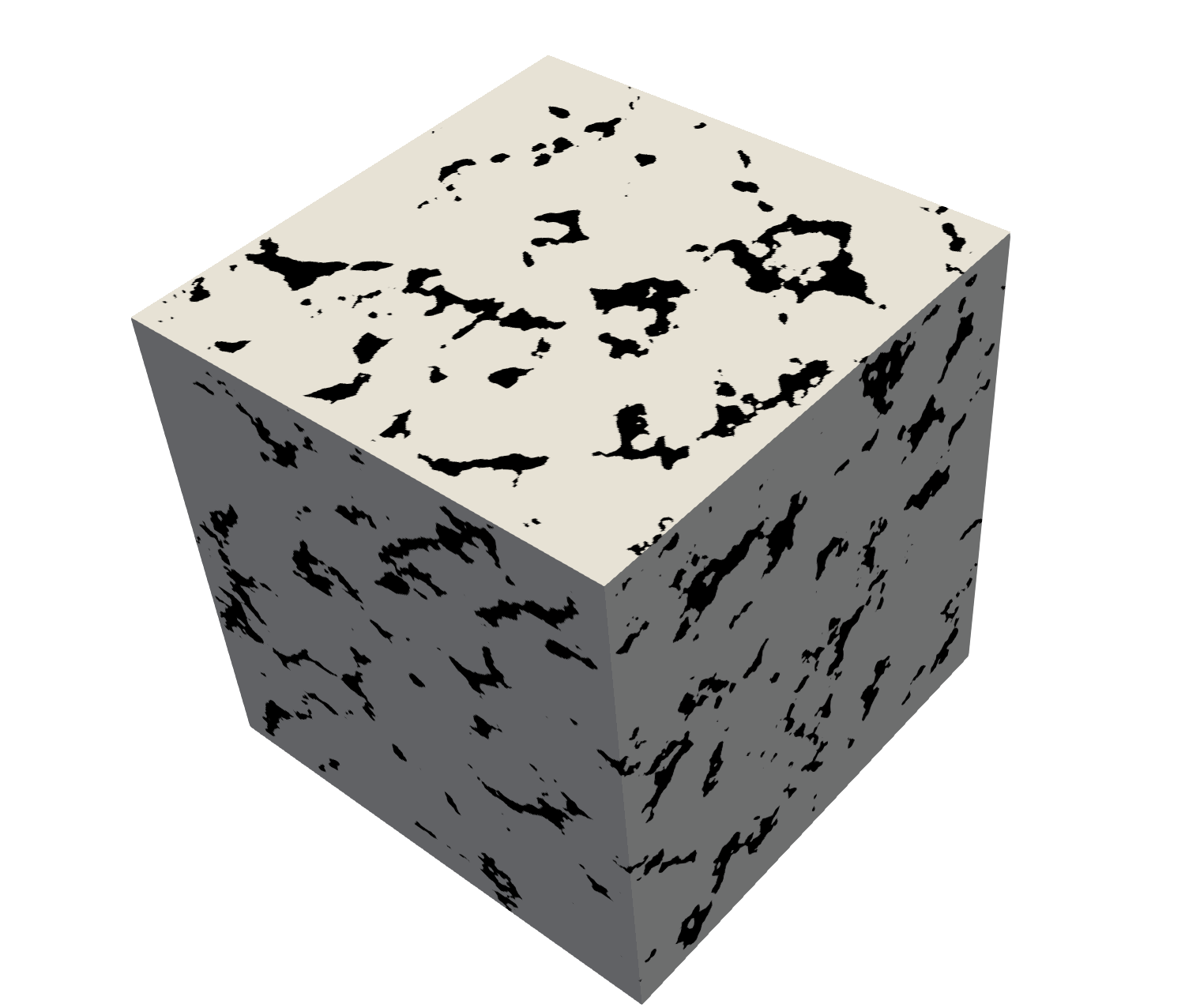}
        \caption{Our $428^3$}
        \label{fig:mouse}
    \end{subfigure}
    \caption{Estaillades sample. Real (size~$150^3$), sampled with our model (size~$150^3$), samples with the baseline model (size~$150^3$), sampled with our model (size~$428^3$)}\label{fig:3d_estaillades}
\end{figure}

\begin{figure}[!h]
    \centering
    \begin{subfigure}[b]{0.15\textwidth}
        \includegraphics[width=\textwidth]{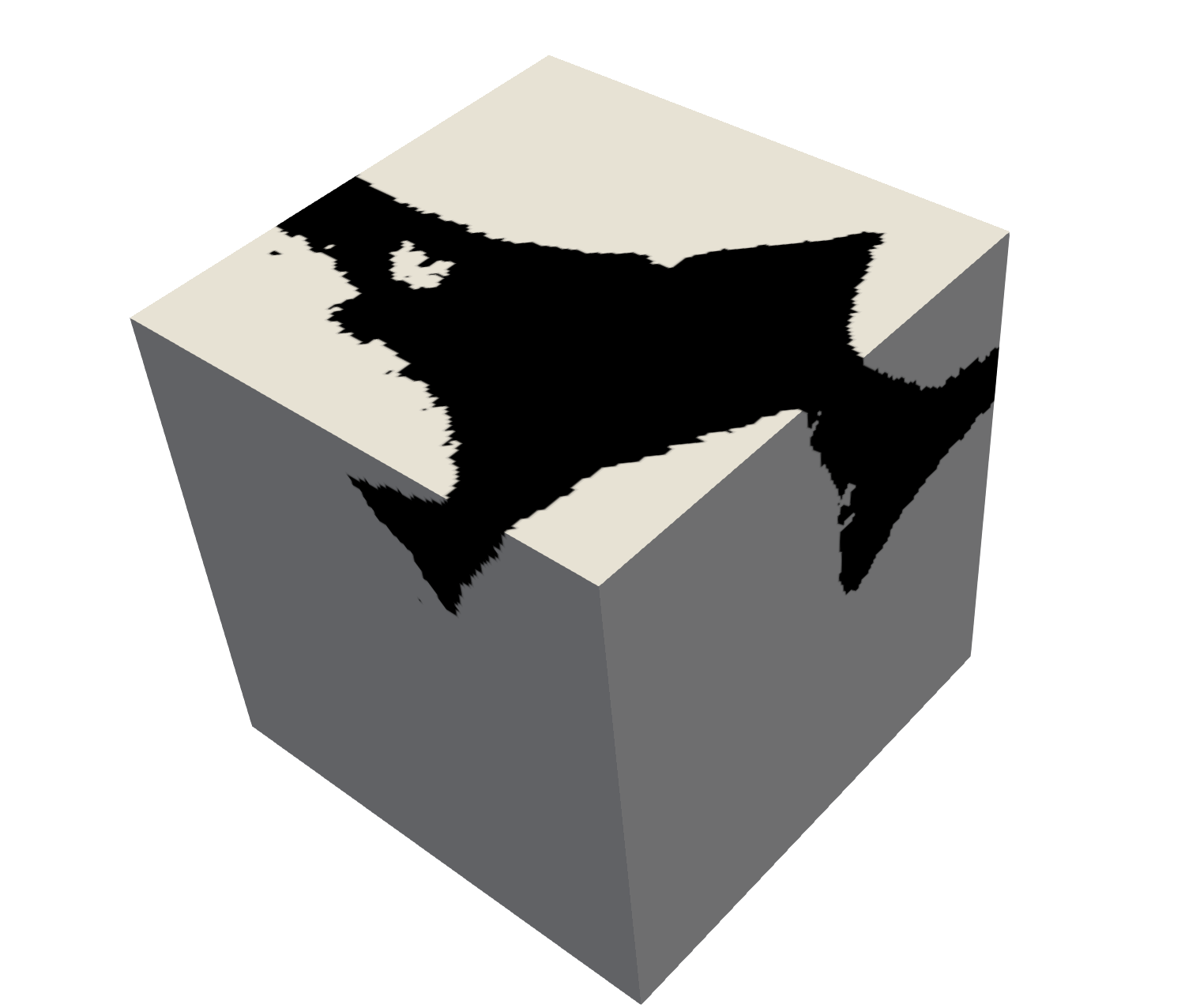}
        \label{fig:tiger}
    \end{subfigure}
    ~ 
    \begin{subfigure}[b]{0.15\textwidth}
        \includegraphics[width=\textwidth]{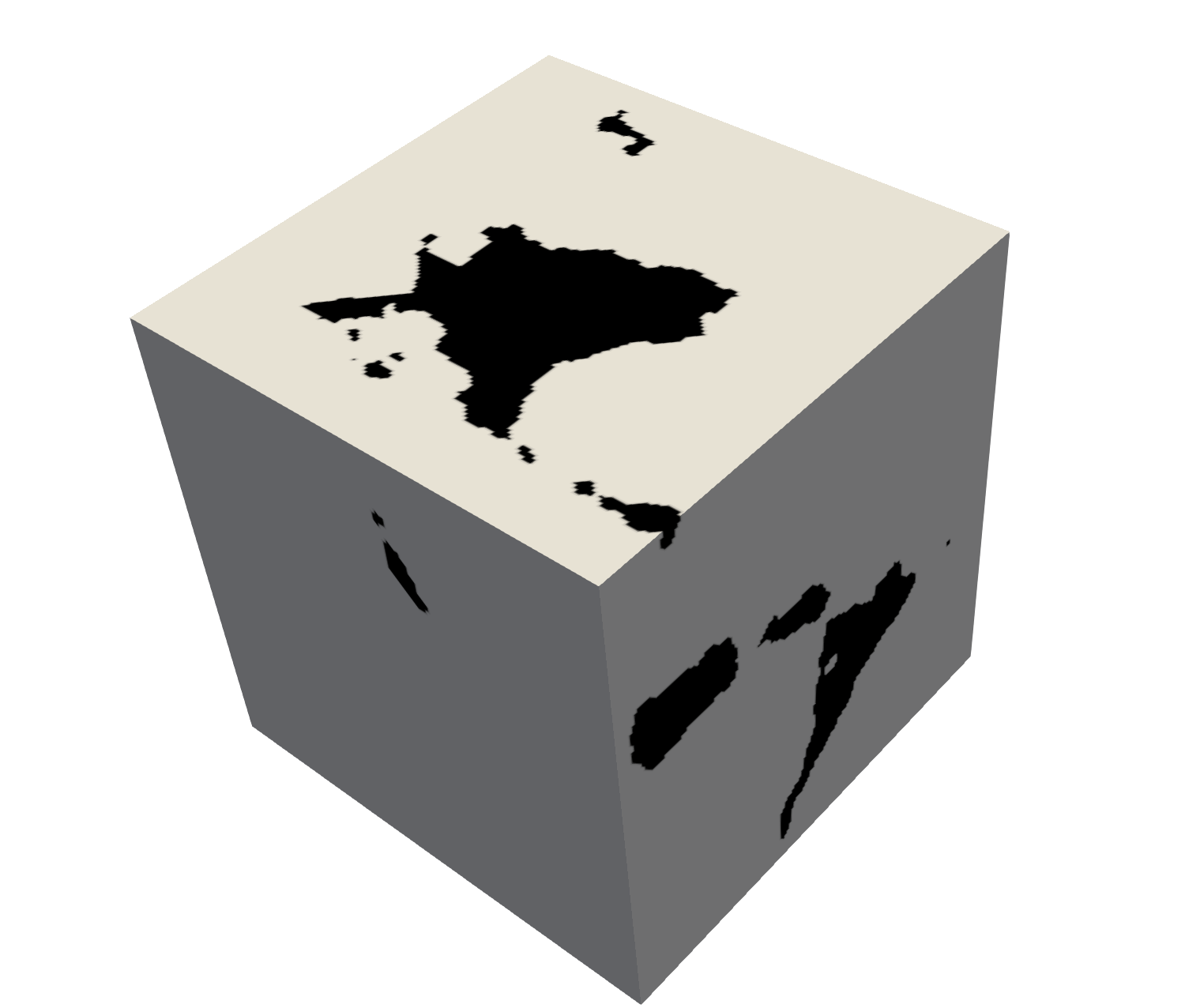}
        \caption{Our $150^3$}
        \label{fig:gull}
    \end{subfigure}
    ~ 
    \begin{subfigure}[b]{0.15\textwidth}
        \includegraphics[width=\textwidth]{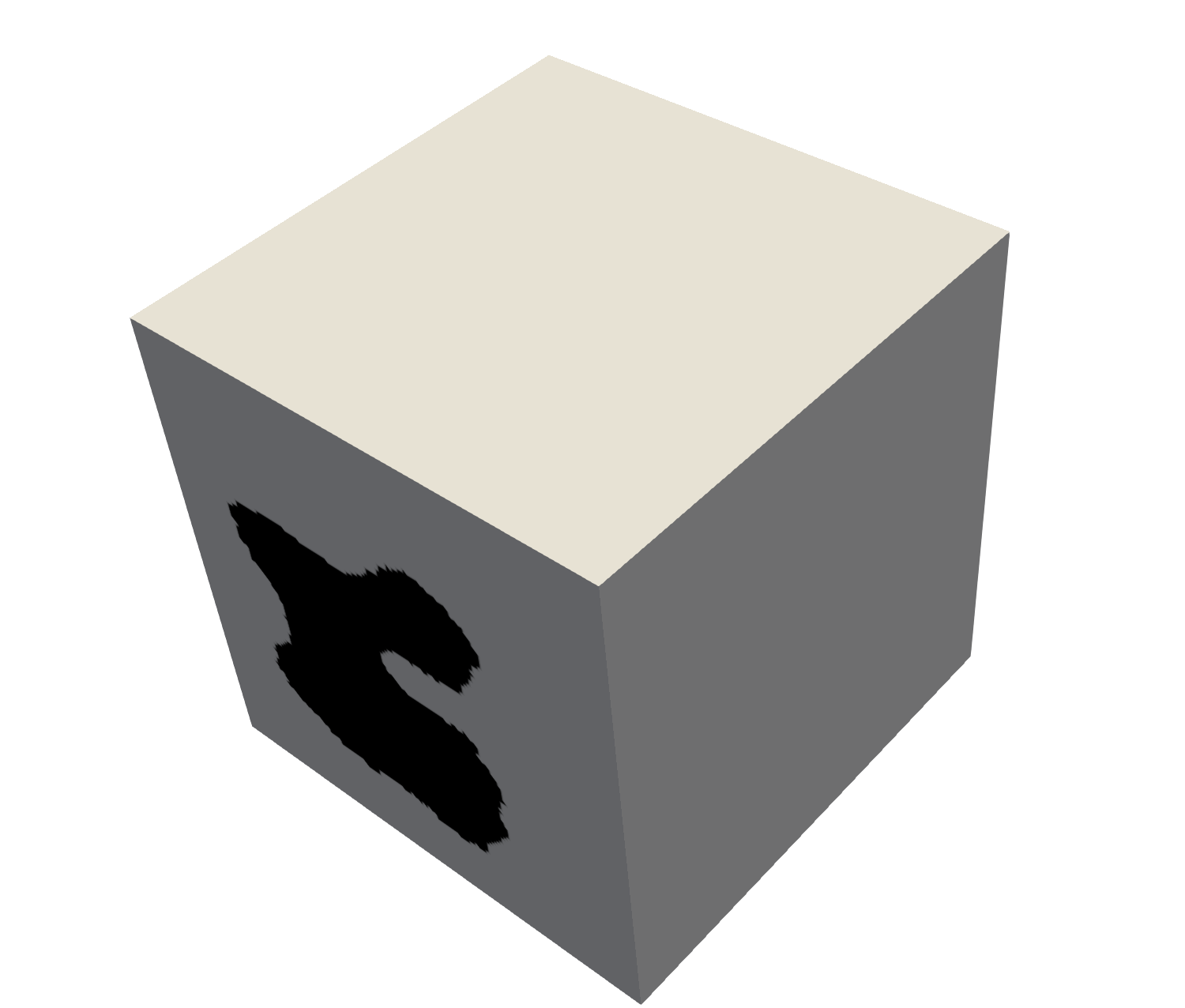}
        \caption{Baseline $150^3$}
        \label{fig:mouse}
    \end{subfigure}
    ~ 
    \begin{subfigure}[b]{0.428\textwidth}
        \includegraphics[width=\textwidth]{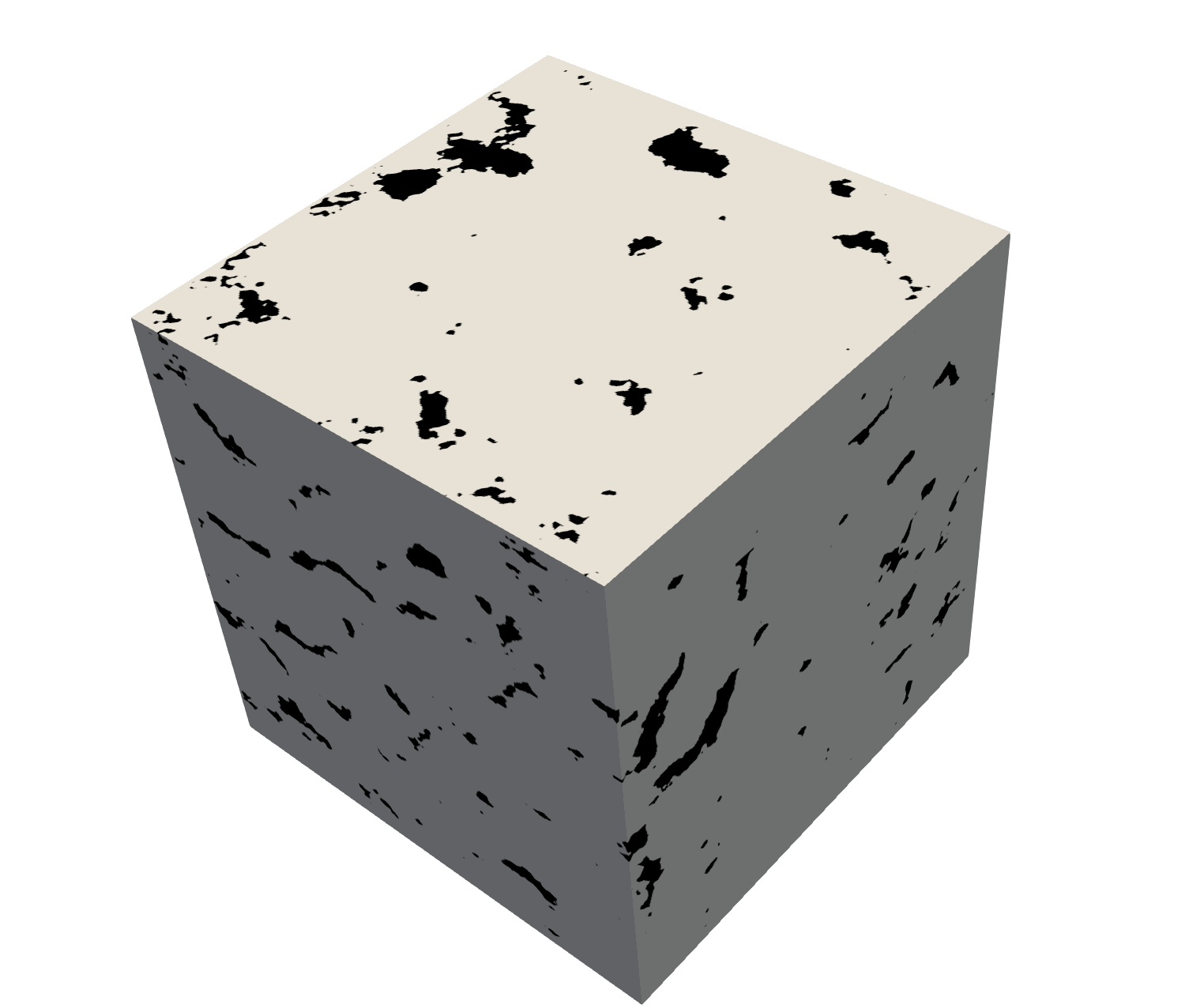}
        \caption{Our $428^3$}
        \label{fig:mouse}
    \end{subfigure}
    \caption{Ketton sample. Real (size~$150^3$), sampled with our model (size~$150^3$), samples with the baseline model (size~$150^3$), sampled with our model (size $428^3$)}\label{fig:3d_ketton}
\end{figure}

We provide a 3D visualization for other types of porous media. For Bentheimer see Fig. \ref{fig:3d_bentheimer}, for Doddington see Fig. \ref{fig:3d_doddington}, for Estaillades see Fig. \ref{fig:3d_estaillades} and for Ketton see Fig. \ref{fig:3d_ketton}. In each figure, there are four samples: real, our, baseline and our big sample. 

In order to increase the synthetic sample size we should increase the spatial size of the latent space. In our case, we trained the model with the spatial latent size $7$, what corresponds to the output samples of size $160^3$. 

For visual comparison we cropped the central cube of size $150^3$ from the synthetic one both for our model and for the baseline. This is caused by side effects because of non-zero padding. Considering big samples, we used spatial latent size $25$, which resulted in samples of size $448^3$. However, due to side effects we cropped the synthetic samples to the size $428^3$ for visual illustration.

\clearpage

\subsection{Metrics comparison}
As in section \ref{sec:3d}, we provide the comparison of the KL divergences between other Minkowski functionals of real and our samples and between real and baseline samples. The results for the functionals Volume and Mean Breadth are provided in Table~\ref{tab:mink_3d_appendix}. As we can see, in most of cases our model is better then the baseline.

\begin{table}[h]
\centering
\caption{KL divergence between real, our and the baseline distributions of statistics (Volume and Mean Breadth) for size $160^3$. The standard deviation was computed using the bootstrap method with $1000$ resamples}
\ra{1.3}
\begin{tabular}{|c|c|c|c|c|}
\hline
\multirow{2}{*}{}                       & \multicolumn{2}{c|}{Volume}               & \multicolumn{2}{c|}{Mean breadth}                       \\ \cline{2-5} 
                  & $KL(p_{real}, p_{ours})$ & $KL(p_{real}, p_{baseline})$ & $KL(p_{real}, p_{ours})$ & $KL(p_{real}, p_{baseline})$ \\ \hline
 Ketton   & \textbf{3.60 $\pm$ 0.44} & 7.28 $\pm$ 0.13 & 4.94 $\pm$ 1.14 & \textbf{1.37 $\pm$ 0.66} \\ \hline 
 Berea   & 4.44 $\pm$ 1.07 & 3.80 $\pm$ 0.62 & 2.63 $\pm$ 0.80 & 2.80 $\pm$ 0.47 \\ \hline 
 Doddington   & 6.34 $\pm$ 2.10 & \textbf{3.59 $\pm$ 0.42} & 8.24 $\pm$ 2.21 & 8.85 $\pm$ 1.64 \\ \hline 
 Estaillades   & \textbf{0.71 $\pm$ 0.13} & 3.11 $\pm$ 0.58 & 5.53 $\pm$ 0.70 & 4.59 $\pm$ 0.89 \\ \hline 
 Bentheimer   & \textbf{0.70 $\pm$ 0.28} & 6.20 $\pm$ 0.51 & \textbf{0.42 $\pm$ 0.16} & 2.62 $\pm$ 1.10 \\ \hline
\end{tabular}
\label{tab:mink_3d_appendix}
\vspace{-0.5cm}
\end{table}

\end{document}